\documentclass[twocolumn]{article}
\usepackage{arxiv}
\usepackage[utf8]{inputenc} %
\usepackage[T1]{fontenc}    %
\usepackage{hyperref}       %
\usepackage{url}            %
\usepackage{booktabs}       %
\usepackage{amsfonts}       %
\usepackage{nicefrac}       %
\usepackage{microtype}      %
\usepackage{graphicx}
\usepackage[numbers,sort&compress]{natbib}
\usepackage{doi}
\usepackage{authblk}
\usepackage{xurl}
\usepackage{changepage}
\usepackage{array}
\usepackage{siunitx}
\usepackage{threeparttable}
\usepackage{amssymb}
\usepackage{caption}
\usepackage{makecell}

\usepackage{amsmath}        %
\usepackage{cleveref}

\title{Using profiles of cognitive capability to assess AI suitability for workplace tasks}
\date{} 	
\author{
\textbf{Jonathan Prunty}$^{1}$\quad
\textbf{Marko Tešić}$^{2}$ \quad
\textbf{Patrick Quinn}$^{1}$\quad
\textbf{José Hernández-Orallo}$^{1,3}$ \quad
\textbf{Lucy Cheke}$^{1}$ 
\vspace{12pt} \\
$^{1}$University of Cambridge \\
$^{2}$Department for Science, Innovation and Technology \\
$^{3}$Universitat Politècnica de València
}

\renewcommand{\undertitle}{Technical Report}
\renewcommand{\headeright}{Technical Report}
\renewcommand{\shorttitle}{A Cognitive Framework for Assessing Task Suitability}

\begin{document}

\twocolumn[
  \maketitle
]
\begin{adjustwidth}{0.0001cm}{0.0001cm}
\begin{abstract}
Organisations deploying AI face a scoping problem: which tasks can be automated, which should remain with humans, and which are best shared between the two. Aggregate benchmark scores provide little insight into where systems will succeed or fail in practice, while human judgements of model capabilities quickly become outdated. We introduce a pipeline that profiles agents and tasks using a shared set of core cognitive capabilities. \textit{Cognitive capability profiling} infers an agent’s capabilities from performance on a benchmark battery annotated for the cognitive demands of each item. \textit{Task requirements weighting} elicits from domain experts the relative importance of these same capabilities for their work. As both use a common set of cognitive dimensions, they can be updated independently as models and roles change, and combined to estimate AI suitability at the level of a domain, organisation, role, or individual duty. We validate capability recovery on synthetic agents, profile six AI systems, and elicit task requirements from 410 employees across six occupational domains. AI systems differed more across cognitive dimensions than across model families, while workplace activities converged on a shared cognitive core. The resulting scores provide a comparative scoping tool for identifying promising candidates for piloting and areas where current systems are unlikely to be well suited. We discuss extending the framework to profile human workers alongside AI systems, moving from AI suitability towards human–machine task allocation.
\end{abstract}
\end{adjustwidth}


\section{Introduction}\label{sec:intro}
Pre-deployment evaluations of AI systems often fail to predict real-world performance. Whether relying on aggregate benchmark scores, relative positions on a leader-board, or piecemeal trial-and-error feedback gathered via ad-hoc testing or red-teaming, the outcome is familiar. Once deployed, AI systems frequently exhibit brittleness, with performance shifting or collapsing unpredictably under the complex and evolving demands of real-world deployment \citep{aisi2025trajectories, rabanser2026towards, pan2025measuring, gu2025illusion, fodor2025line, eriksson2025can, mcintosh2025inadequacies}.

This gap between pre-deployment evaluation and real-world performance is reflected in industry concerns. Business leaders consistently cite reliability as a major barrier to AI adoption, often ranking it above cost or capability limitations, and viewed as a primary risk to be mitigated amongst high adopters \citep{pan2025measuring, dsit2026aiadoption, hassan2024barriers, mckinsey2025stateofai, poon2025adoption}. These concerns extend beyond organisational decision-makers: Anthropic's survey of 81,000 users similarly found reliability to be respondents' top concern \citep{huang2026interviewer}. Consequently, the promise of transformative productivity gains -- the headline justification for the substantial capital flowing into AI infrastructure \citep{maslej2025artificial} -- remains largely unrealised \citep{narayanan2025ainormaltechnology, yotzov2026firm}, in part because organisations cannot confidently scope which tasks are appropriate candidates for automation, which are better retained by human workers, and which might be most effectively handled through some combination of the two.

At first glance, this may seem like an intractable problem. Organisations have limited time and resources for pre-deployment testing, meaning evaluations can cover only a subset of the tasks a system will ultimately face. Real-world tasks also follow a different distribution of demands than curated evaluation instruments -- typically involving greater complexity and variability \citep{chaytor2003ecological, raji2021ai, aisi2025trajectories}. As a result, it is highly likely that a deployed system will encounter combinations of demands that were never included in testing, leading to unreliable performance.

A closer look, however, suggests that \textit{predictability}, not task reliability \textit{per se}, should be the primary target of concern \citep{zhou2026predictable}. An aggregate benchmark score -- 85\% on MMLU \citep{wang2024mmlu}, for instance -- provides some signal about overall performance on a given task, but it says little about \textit{why} a system failed certain instances, or \textit{how it will perform} on future ones, especially those that differ meaningfully from the evaluation distribution \citep{burnell2023rethink, zhou2026general}. In fact, a less capable system whose failures are nonetheless systematic (that is, reliably tied to identifiable task demands) is substantially more deployable than one with higher benchmark accuracy but unpredictable failure modes \citep{zhou2024larger,zhou2026predictable}. 

When failures are predictable, they are manageable. They permit meaningful human oversight, support the design of appropriate guardrails and handoff protocols, and enable organisations to target deployment to scenarios within the system's capability remit. Unpredictable failure, by contrast, demands either blanket human review, which would negate much of the efficiency gain, or the acceptance of a risk profile that is too opaque for most organisations to responsibly consider.

Part of what makes failure unpredictable is that AI systems tend to have non-human-like -- ``jagged'' -- capability profiles \citep{dell2023navigating}, meaning our intuitions about where they will struggle are unreliable \citep{mineault2026cognitive, zhou2024larger}. One approach is to shift focus from evaluating performance on specific \textit{tasks} to estimating the underlying \textit{capabilities} that determine how performance generalises across them \citep{hernandez2017evaluation, burden2023inferring}. Instead of asking whether a system can perform a particular task at a given moment, capability-based evaluation asks whether a system possesses the capabilities that a task demands, so that performance on novel tasks can be predicted from the match between those capabilities and task requirements.

Consider an analogy: knowing that an athlete has a high-jump capability of two metres (that they will clear any bar at or below that height, and fail reliably above it) is far more useful for predicting future performance than any aggregate score (knowing, for instance, that they have cleared 85\% of their total attempts). By mapping the athlete's capability level (two metres) to the task demand level (a new three metre jump), we can predict their likely performance (failure). While this analogy is simplistic, the same logic applies to the evaluation of agents for the workplace: a sufficiently detailed profile of an agent's \textit{cognitive} capabilities (ability levels associated with functions such as memory, perception and reasoning) when mapped against the demand profiles of workplace tasks, allows the prediction of when performance will hold and when it will break down, for human and AI agents alike \citep{prunty2026reverse}. The challenge, of course, is constructing such profiles in a robust and practically feasible way, and characterising cognitive labour with sufficient precision for meaningful mapping.

\begin{figure}[ht]
    \centering
    \includegraphics[width=0.95\linewidth]{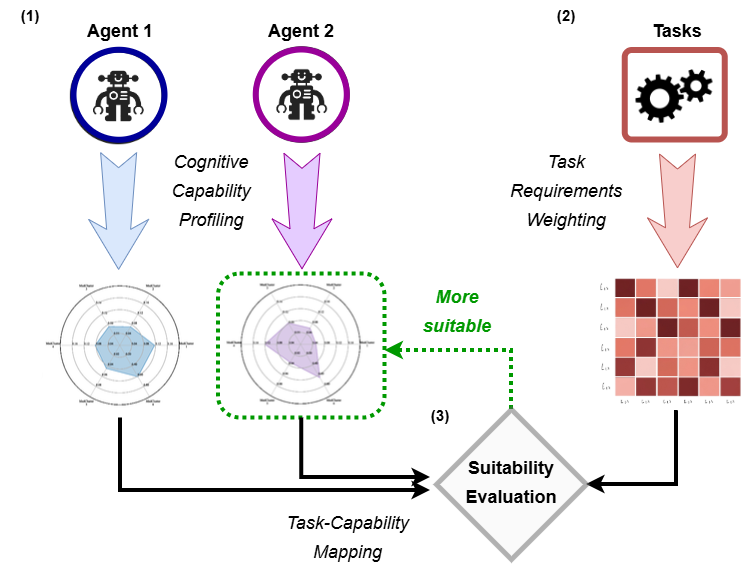}
    \caption{The pipeline for assessing task suitability involves: (1) generating cognitive capability profiles for a selection of agents, (2) weighting the cognitive requirements of target workplace tasks by their relative importance, and (3) mapping capability profiles to relative importance matrices. For simplicity, we have included two candidate AI agents, but commensurable capability profiles can also in principle be generated for humans and for human + AI pairings.}
    \label{fig:pipeline}
\end{figure}

In the present work, we introduce a pipeline for making principled task suitability decisions (Figure~\ref{fig:pipeline}). The pipeline pairs \textit{cognitive capability profiling} with a structured \textit{task requirements weighting} process, such that the cognitive demands most critical to a given workplace task can be matched against the known capability profile of any candidate AI system -- transforming task suitability judgements from intuition or trial-and-error into a systematic, evidence-based assessment process.

\section{Background}\label{sec:background}
\subsection{Assessing the cognitive capabilities of AI and humans}\label{sec:assessing_capabilities}

Traditional approaches to hiring and role assignment focus on attributes and specialist skills that differentiate candidates from one another, on the implicit assumption that all candidates share a common foundation of core cognitive competencies. This assumption holds reasonably well when hiring pools consist entirely of adult humans. Candidates may vary in expertise, personality, or domain knowledge, but they usually share the more fundamental perceptual, reasoning, and communicative capacities that underpin competent performance in the workplace.

Machine intelligence cannot be treated under the same assumptions. The jagged capability profile that characterises modern AI systems is precisely a description of a non-human-like profile of cognitive competencies: certain capabilities, such as the breadth of factual knowledge encoded in large language models, vastly exceed that of any individual human, while others remain surprisingly limited, in ways that can be difficult to anticipate \citep{dell2023navigating, mineault2026cognitive}. Even capabilities that seem elementary by human standards -- reliably perceiving and acting in naturalistic, visually complex environments, for instance -- remain fragile in systems that otherwise appear highly capable \citep{schulze2025visual, tangtartharakul2026visual, prunty2026visuospatial}. Because AI systems do not share the evolutionary history, developmental trajectory, or cognitive architecture of human agents, there is no safe common foundation upon which to build assumptions. Overlooking this risks exposing deployed systems to tasks that demand a ``hidden'' capability the system does not reliably possess, leading to unanticipated failures.

Fortunately, there is a substantial body of work in the cognitive sciences identifying the latent constructs that underpin everyday function in humans \citep{carroll1993human, spelke2007core, lake2017building}. The methodological tools developed to measure these constructs, however, do not transfer straightforwardly to the evaluation of AI. Human cognitive abilities are typically assessed using psychometric instruments -- carefully designed test batteries that infer latent capabilities from patterns of behavioural performance, but that assume the subject is drawn from a population of human adults. AI systems, by contrast, are most often evaluated using large-scale benchmarks that target performance within specific task domains, but without careful attention to the constructs underlying that performance. Neither tradition, in isolation, delivers what capability profiling requires.

Applying human psychometric instruments directly to AI raises the problem of \textit{anthropomorphism} \citep{dowe2012iq,hernandez2017measure,milliere2024anthropocentric, mitchell2026six}. Human evaluation methods are often not well-suited to AI systems. Specific cognitive tests are typically small in scale, and are likely to have been included, or closely approximated, in language model training data, inflating their test performance and making it easy to overestimate their capabilities \citep{xu2024benchmark, li2024task}. For instance, a model with high scores on a Theory of Mind test may be solving the task by pattern-matching to similar tasks in the training data rather than engaging in genuine mental-state inference \citep{hu2025re} -- meaning that their performance may not transfer to novel scenarios \citep{ullman2023large}. In principle, it is possible to design cognitive tests that are valid for both humans and AI \citep[see, for instance,][]{voudouris2025animal, prunty2025intuit, rutar2025general}, but achieving this validity demands careful expert construction, putting such batteries out of reach of the pace and scale that enterprise-level profiling requires.

Existing AI benchmarks offer exactly the scale and ecological breadth that hand-crafted test batteries lack, but carry a validity problem of their own. They are built to measure task performance within broad domains, grouping items under headings such as ``programming'' or ``question answering'', rather than isolating the underlying cognitive constructs that performance draws upon. Crucially, they rarely specify what cognitive demands an individual item places on a system, or how demanding one item is relative to another \citep{burnell2023rethink}. A raw benchmark score therefore reveals little about \textit{which} capabilities a system possesses, or at what level. Returning to the high-jump analogy: knowing that an athlete cleared 85\% of their attempts tells us little if we do not know anything about how the bar heights were distributed.

One approach, however, combines the scale of benchmark testing with the construct validity of psychometric design. The ADeLe framework \citep{zhou2026general} ``unlocks'' existing benchmarks for construct-based evaluation by applying expert-constructed rubrics to annotate individual benchmark items -- with each rubric specifying the demand levels that define difficulty on a targeted cognitive capability. In this method, an LLM judge acts as a scalable proxy for the domain expert, applying the knowledge encoded in the rubric to produce fine-grained, item-level demand annotations across large benchmark catalogues, supplying precisely the per-item information about where each ``bar'' is set that is needed for capability-based evaluation.

One might note a potential circularity here: using an LLM to characterise the cognitive demands of items on which LLMs are themselves evaluated. This concern is partly mitigated by the separation of roles. The annotating model acts as a rubric-applying classifier rather than a subject whose capabilities are under assessment -- a task for which inter-rater reliability with human experts is well established \citep{zheng2023judging, deshpande2025multichallenge} -- while the rubrics themselves encode human expert judgement not model-derived criteria.

The present pipeline applies this methodology primarily to AI capability profiling, where the need is most acute and the methodological challenges most severe. The same framework extends to human capability profiling: because the annotations define a capability space that is not specific to any one type of agent, appropriately sampled subsets of the annotated item catalogue could in principle serve as targeted psychometric instruments for human participants, administered in the conventional way, subject to the practical constraint that any individual can complete only a limited number of items \citep{prunty2026reverse}. This preserves commensurability (human and AI profiles expressed in the same capability space, derived from the same underlying annotations \citep{romero2026human}) without requiring entirely separate assessment instruments. How these profiles, once generated, can be mapped against workplace requirements is taken up in the following subsection.

\subsection{Assessing suitability for workplace tasks}\label{sec:assessing_workplace}

Generating a capability profile addresses only half of the suitability question. Determining whether a candidate agent is suited to a given role also requires a structured account of what that role demands and a principled way of mapping the two together (Figure \ref{fig:pipeline}).

Any occupational role can be decomposed into the tasks and subtasks that comprise it. Jobs have been extensively characterised in terms of their constituent activities, tasks, and roles, most notably through O*NET \citep{onet2026}, a comprehensive taxonomy covering hundreds of occupations.
These range from broad activities common to many roles -- analysing data, communicating with colleagues, scheduling work -- to fine-grained tasks specific to particular occupations. These taxonomies offer a structured way to describe work and have become a standard tool for studying the potential effects of AI on the labour market \citep{felten2019occupational, eloundou2024gpts, acemoglu2025simple}.

A growing body of work uses these task inventories to estimate how exposed occupations are to AI. One strand applies rubrics through which human experts, or LLMs as proxies, judge whether a system could complete a given task \citep{brynjolfsson2018can, eloundou2024gpts}; another infers exposure from observed use, mapping real model interactions onto O*NET tasks \citep{handa2025economic}. For deployment decisions, both are unsuitable. Usage-based estimates are backward-looking, capturing only how today's systems are already deployed; judgements about AI capabilities are forward-looking but unreliable, resting on expert forecasts about opaque, non-human-like systems -- forecasts that may quickly become outdated as capabilities evolve \citep{eloundou2024gpts}. Both are anchored to the current generation of AI, becoming outdated with each new model release.

The alternative is to use expert judgement to characterise the task rather than the technology: to ask not whether current systems can perform an activity, but what cognitive demands it places on any agent, human or artificial. These demands are properties of the work itself, not of any particular model. They shift as organisations and practices change, but far more slowly than AI capabilities, or our impressions of them, making them a more stable basis for analysis. This also places expert judgement where it is most reliable. Domain experts may be poorly placed to predict how an opaque AI system will perform, but they are well placed to say what their own work requires. And a task profile defined in this way can be compared against the capabilities of any agent, present or future.

How, then, should task requirements be characterised so that they map onto a capability profile? Here we can draw on a long tradition in applied psychology concerned with the expertise, aptitudes, and capabilities that successful task performance requires \citep[e.g.][]{onet2026, crandall2006working}. These efforts provide rich accounts of work, but were developed for human workers and as such do not transfer cleanly to commensurable capability profiling. The O*NET skills taxonomy, for instance, foregrounds the professional attributes that differentiate human adults from one another, rather than the core cognitive capacities common across all adults -- precisely the capacities that cannot be assumed in AI systems. Fleishman's ability-requirements taxonomy \citep{fleishman1975toward, fleishman1984taxonomies} comes closest to characterising those core abilities, describing tasks in terms of the human abilities they demand and asking subject-matter experts to rate the importance of each; but its inventory is itself human-centric, with psychomotor and physical abilities prominent among categories that do not apply to AI. 

More recent work has argued that the evaluation of AI for real-world roles should similarly be grounded in underlying capabilities rather than shallow task performance, and has proposed candidate taxonomies of the abilities relevant to work \citep{martinez2020does, hernandez2019ai, tolan2021measuring, OECD2025indicators, burnell2026measuring, HernandezOrallo2021IdentifyingAICapabilities, Cheke2021CommonSense}. Building directly on this tradition, our contribution is to provide a pipeline that makes such accounts usable in practice: profiling tasks and agents on shared capability dimensions and mapping the two to support concrete suitability decisions.

In principle, the most direct solution would be to annotate workplace task instances for their cognitive demands using the very rubrics applied to benchmark items during capability profiling, placing task demands and agent capabilities on a single, common scale. In practice this is rarely feasible. It would require either that employees perform instance-level demand annotation across the full range of their activities, which would be impractical at organisational scale, or that suitable task datasets exist to be annotated using an automated pipeline, which for most real-world roles they do not.

We therefore adopt a pragmatic alternative. Instead of annotating task demands directly, we ask domain experts to weigh the relative \emph{importance} of a common set of core cognitive capabilities for each target activity, the same dimensions along which agent profiles are expressed. This yields a weighted requirement profile that can be mapped against any candidate agent's capabilities within a single space. As the resulting importance weights are relative scores, the mapping reflects how much each capability matters for the task but does not test whether the agent meets a demand threshold -- a distinction we return to in Section \S\ref{sec:suitability_mapping}. The resulting estimates can serve as a scoping tool, identifying where AI deployment is most promising, but do not replace piloting, much as a structured assessment of aptitudes can shortlist job candidates without removing the need for a probationary period \citep{fleishman1975toward, fleishman1984taxonomies}. The following section describes how the pipeline is implemented.

\section{Implementation framework}\label{sec:implementation}

As illustrated in Figure \ref{fig:pipeline}, the pipeline comprises three stages: \textit{Cognitive Capability Profiling}, where the cognitive capabilities of AI agents are inferred from their performance data on a demand-annotated test battery; \textit{Task Requirements Weighting}, where the relative importance of cognitive capabilities for workplace tasks is elicited from domain experts; and \textit{Suitability Mapping}, where the two data streams are mapped together to produce actionable estimates of how suited an AI agent is to a particular task. The two streams are different in kind. Profiling yields capability \textit{levels}, while requirements gathering produces importance \textit{weights}. By weighting capabilities according to their importance for each task, suitability scores reflect how well an agent matches the capabilities that matter most, not whether it clears a fixed demand threshold (\S\ref{sec:assessing_workplace}).

\subsection{Cognitive Capability profiling}\label{sec:capability_profiling}

Estimating an agent's capability profile requires knowing the level of cognitive demand at which its performance breaks down. However, standard benchmarks are rarely annotated at the item level with the cognitive demands they impose on a system (\S\ref{sec:assessing_workplace}). We therefore build a profiling battery by annotating existing benchmarks with per-item demand levels across multiple capabilities, following the rubric-based annotation methodology of \citet{zhou2026general}, and then infer capability profiles from performance on that battery. The remainder of this subsection specifies the three steps this involves: defining a capability set and demand rubrics (\S\ref{sec:capability_rubrics}), annotating and filtering benchmark items into a targeted battery (\S\ref{sec:benchmark_selection}), and collecting performance data to model capability estimates (\S\ref{sec:profile_estimation}). The full implementation, including code, rubrics, and benchmark annotations, is available in the project repository.\footnote{\url{https://github.com/Kinds-of-Intelligence-CFI/Task-Suitability-Profiles}}

\subsubsection{Capabilities and rubrics}\label{sec:capability_rubrics}

Drawing on existing literature in psychometrics and cognitive science \citep{carroll1993human, spelke2007core}, we identified 18 core cognitive capabilities (Table~\ref{tab:cognitive_capabilities}) that cover a range of cognitive processes relevant to general workplace activity rather than the specialist skills that differentiate one worker from another (\S\ref{sec:assessing_workplace}). These fall into four broad families: \textbf{memory systems}, concerned with how knowledge and skills are retained and retrieved \citep{squire2000memory, tulving1972episodic}; \textbf{executive control}, managing cognitive resources and guiding goal-directed behaviour \citep{diamond2013executive, miyake2000unity, baddeley1992working}; \textbf{object and space understanding}, covering the perception of and interaction with objects in physical and digital environments \citep{spelke2007core, biederman1987recognition, treisman1980feature, gibson1979ecological}; and \textbf{social and communicative} capabilities, supporting interaction with colleagues and clients \citep{premack1978chimpanzee, frith2006neural, clark1996using}. This set converges substantially with other recent capability taxonomies for work and AI evaluation \citep{OECD2025indicators, burnell2026measuring, hernandez2019ai}, reflecting a shared move toward grounding evaluation in underlying constructs rather than surface-level task performance.

For each capability, we developed a scoring rubric (Figure~\ref{fig:rubric}) specifying how to categorise the demands a benchmark item places on a system, on a six-point scale from level 0 (the capability is not required) to level 5 (a very high level of the capability is required). In designing the levels, we were guided intuitively by how large a proportion of a human population would possess the capability at each level: level 1 denoting a demand most adults would meet, level 5 one that only a small minority would. Following the rubric-based annotation methodology of \citet{zhou2026general}, each level is anchored by concrete descriptors of the cognitive demand it represents, along with examples of tasks that require that level of demand. This enables annotators to assign levels consistently across heterogeneous benchmark items. Crucially, demand levels are defined \emph{within} each capability: a level-4 demand on Working Memory and a level-4 demand on Theory of Mind both indicate high demand on their respective capabilities, but are not assumed to be equivalent in absolute terms. Each item is therefore characterised by its profile across all 18 capabilities.

The scale is structured so that successive levels correspond to geometrically increasing demand. In the inference model (\S\ref{sec:profile_estimation}), a demand level $d_{j,k}$ on capability $k$ for item $j$ enters as a raw difficulty $\delta_{j,k} = e^{\lambda d_{j,k}}$, so that each one-level increase multiplies the difficulty by a constant factor $e^{\lambda}$; in our experiments we fix $\lambda = 1$, though this quantity can also be estimated. The levels thus span a wide dynamic range while remaining commensurable with an agent's estimated capability on the same logarithmic scale, allowing demand and capability to be compared directly as a log-ratio. Importantly, the inference depends only on this ordinal structure and geometric spacing, not on the population intuition that motivated the levels: we make no claim that they correspond to fixed proportions of the human population. What the levels provide is an interpretable and internally consistent scale of cognitive demand, whose validation against human performance data we return to in \S\ref{sec:discussion}.

\subsubsection{Benchmark selection and refinement}\label{sec:benchmark_selection}

Having defined the capabilities and rubrics, we surveyed the AI evaluation literature for benchmarks targeting one or more of these abilities, and then assembled a combined battery from them. Our aim was a battery of roughly 20,000 items that balanced breadth and representativeness: covering the full range of demand levels, including rare cases, while reflecting the demands most commonly encountered in real tasks. For this selection process, we annotated the demands of a random sample of 200 items per dataset using GPT-4o. These annotations defined a target distribution for the battery that blends the observed demand distribution (weight 0.7) with a flat distribution (weight 0.3) -- the former keeping coverage representative of common demand levels, the latter ensuring rare, high-demand levels are present. We then selected the number of items we needed from each benchmark in order to approximate this target as closely as possible, subject to the constraints that no single dataset contribute more than 10\% of the battery and that allocations respect each dataset's available sample size. The resulting allocations are given in Table \ref{tab:benchmarks}; some datasets were later removed because performance accuracy could not be scored automatically, leaving 19,576 items.

\begin{table}[h!]
\centering
\caption{Benchmarks}
\label{tab:benchmarks}
\begin{tabular}{p{0.7\linewidth} r}
\toprule
Benchmark & Items \\
\midrule

\href{https://github.com/google/BIG-bench/tree/main/bigbench/benchmark_tasks/abstract_narrative_understanding}{Abstract Narrative Understanding} \citep{srivastava2023beyond} & 303 \\

\href{https://github.com/ruixiangcui/AGIEval}{AGIEval} \citep{zhong2024agieval} & 1896 \\

\href{https://github.com/suzgunmirac/BIG-Bench-Hard}{BigBenchHard} \citep{suzgun2023challenging} & 1265 \\

\href{https://github.com/cicl-stanford/procedural-evals-tom/tree/main}{BigToM} \citep{gandhi2023understanding} & 686 \\

\href{https://github.com/google/BIG-bench/tree/main/bigbench/benchmark_tasks/cause_and_effect}{Cause and Effect} \citep{srivastava2023beyond} & 51 \\

\href{https://github.com/google/BIG-bench/tree/main/bigbench/benchmark_tasks/coqa_conversational_question_answering}{CoQA} \citep{srivastava2023beyond} & 210 \\

\href{https://github.com/Sahandfer/EmoBench}{EmoBench} \citep{sabour2024emobench} & 1200 \\ 

\href{https://github.com/google/BIG-bench/tree/main/bigbench/benchmark_tasks/evaluating_information_essentiality}{Evaluating Information Essentially} \citep{srivastava2023beyond} & 68 \\

\href{https://huggingface.co/datasets/ewok-core/ewok-core-1.0}{EWoK} \citep{ivanova2025elements} & 210 \\

\href{https://github.com/google/BIG-bench/tree/main/bigbench/benchmark_tasks/fantasy_reasoning}{Fantasy Reasoning} \citep{srivastava2023beyond} & 201 \\

\href{https://hyunw.kim/fantom/}{Fantom} \citep{kim2023fantom} & 2177 \\

\href{https://github.com/Kinds-of-Intelligence-CFI/VIGNET}{INTUIT} \citep{prunty2025intuit} & 995 \\

\href{https://github.com/google/BIG-bench/tree/main/bigbench/benchmark_tasks/known_unknowns}{Known Unknowns} \citep{srivastava2023beyond} & 46 \\

\href{https://huggingface.co/datasets/salem-mbzuai/LLM-BabyBench}{LLM BabyBench} \citep{choukrani2025llm} & 600 \\

\href{https://github.com/allenai/MacGyver}{MacGyver} \citep{tian2024macgyver} & 909 \\

\href{https://github.com/maximegmd/MetaMedQA-benchmark}{MetaMedQA} \citep{griot2025large} & 1373 \\

\href{https://github.com/seacowx/OpenToM}{OpenToM} \citep{xu2024opentom} & 2000 \\

\href{https://github.com/karthikv792/LLMs-Planning}{PlanBench} \citep{valmeekam2023planbench} & 2000 \\
\href{https://huggingface.co/datasets/socialnormdataset/social}{SocialNorm} \citep{yuan2024measuring} & 210 \\

\href{https://huggingface.co/datasets/ZhengyanShi/StepGame}{StepGame} \citep{shi2022stepgame} & 1563 \\

\href{https://github.com/google/BIG-bench/tree/main/bigbench/benchmark_tasks/text_navigation_game}{Text Navigation} \citep{srivastava2023beyond} & 1413 \\

\href{https://github.com/TIGER-AI-Lab/MMLU-Pro?utm_source=chatgpt.com}{MMLU-Pro} \citep{wang2024mmlu} & 200 \\

\midrule
\textbf{Total} & \textbf{19{,}576} \\
\bottomrule
\end{tabular}
\end{table}

We next collected two independent sets of demand annotations from distinct model families (GPT-4o and Gemini 3 Flash), evaluating each benchmark item against the demand rubrics. This yielded a demand matrix $D$ for each rater, with $j$ rows (items) and $k$ columns (capabilities). Inter-rater reliability analysis (Table~\ref{tab:interrater}) identified two capability dimensions below our agreement thresholds ($\rho < 0.30$ and $\kappa_w < 0.30$). Raters did not reliably agree on demand levels for \textit{Attention and Inhibitory Control} and \textit{Prospective Memory}; lacking a consistent difficulty signal, they were excluded. The remaining 16 capabilities were retained, with rank correlations from $\rho = 0.38$ to $\rho = 0.81$, well separated from the excluded pair ($\rho \leq 0.09$). Even for the lower-correlation dimensions, raters agreed to within one demand level on the large majority of items (min $= 74\%$), indicating that disagreements were typically small and local and did not reflect inappropriate use of the scale.

To form a single demand matrix, we averaged the two sets of ratings for each item and retained capability. This disagreement between raters largely reflected small, capability-specific differences in scale use rather than random noise. For example, GPT-4o rated Cognitive Flexibility higher, whereas Gemini-3 rated Semantic Memory higher. The mean therefore provides a neutral estimate that reduces rater-specific bias while preserving the original $0$--$5$ scale. The final battery contained 19,535 items once invalid items from raters were removed\footnote{An item was removed if either rater returned an invalid or missing rating on any of the 16 retained capabilities; this affected 41 of the 19,576 items.}.

\begin{figure*}[ht]
\centering
\includegraphics[width=1\linewidth]{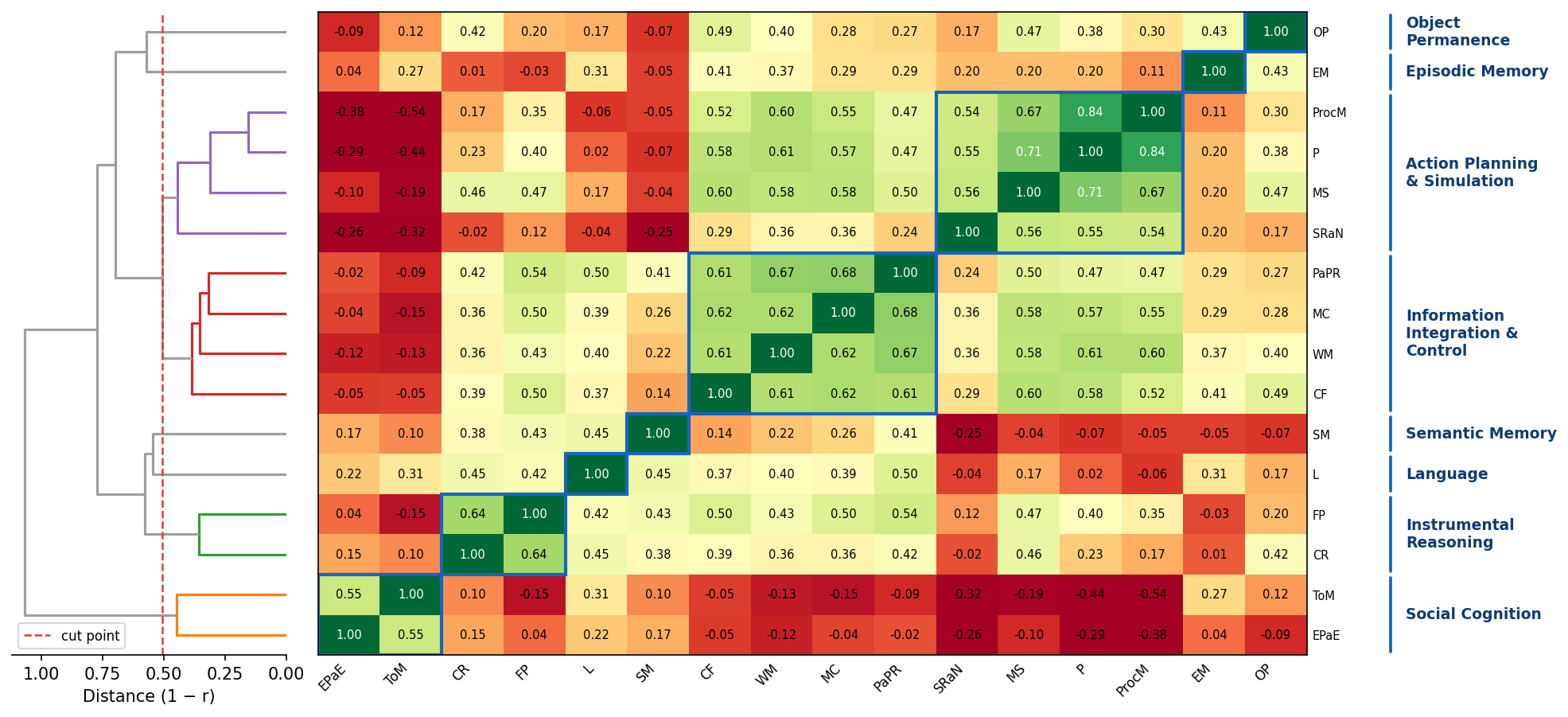}
\caption{Capability clustering. Pearson correlations between the demand profiles of the 16 capabilities retained following reliability screening (Table~\ref{tab:interrater}), revealing dimensions that placed similar demands on the same items. Correlations were computed from annotations z-scored within each rater and capability, controlling for systematic differences between the two raters. The left-hand dendrogram shows hierarchical agglomerative clustering on the distance $1-r$ between dimensions. The red dashed line marks the cut point (distance $=0.5$) that yields eight clusters, each merged into a single composite dimension (named on the right). Full names and definitions of each cognitive capability are given in Table~\ref{tab:cognitive_capabilities}; acronyms and definitions of the resulting composite dimensions are in Table~\ref{tab:clustered_capabilities}.}
\label{fig:demand_correlations}
\end{figure*}

The retained capabilities are not independent in their demand profiles. Cognitively demanding items tend to require many abilities simultaneously, inducing strong positive correlations among the capability columns (Figure~\ref{fig:demand_correlations}) and leaving the inference model little basis for separating the corresponding latent abilities. These correlations do not necessarily reflect conceptual similarity, but the co-occurrence of demands across items: conceptually distinct capabilities may be required together, and the merging step is intended to resolve this collinearity. We therefore merged capabilities with closely aligned demand profiles into composite dimensions using hierarchical agglomerative clustering \citep{rousseeuw1990finding} on the demand-profile correlation matrix, with each rater's demand annotations z-scored within capability to control for differences in rating scale. Cutting the dendrogram (Figure \ref{fig:demand_correlations}, left) at a distance of $1-r=0.5$ yields eight composite dimensions that group capabilities with co-occurring demands while keeping the principal demand families distinct. Each composite's demand on an item is the mean of its constituent capabilities' raw consensus demand scores, preserving the original $0$--$5$ scale.

Two patterns are worth noting in the resulting set of demands. First, no dimension reaches level~5 on any item and only a small proportion reach level~4, reflecting the upper bound of demand present in the selected benchmarks. Second, the merged dimensions vary widely in how their demand is distributed across the battery (Table~\ref{tab:demand_levels}). Some are active on only a subset of items but, where active, span a range of demand levels -- Object Permanence and Social Cognition, for instance, are each required on under half the battery yet carry a tail of higher demands. Others, such as Information Integration \& Control, Semantic Memory, and Language, are near-universal in coverage ($\geq 99\%$), which is to be expected for capabilities that almost any task recruits to some degree. Language, in particular, is required on $100\%$ of items, but $97\%$ are at levels 1–2. This combination of full coverage and narrow range raises a question about identifiability: given near ubiquitous low-level demands, will our inference model be able to correctly estimate these capabilities? We return to this question and assess it directly in our recovery analysis Section~\ref{sec:recovery_results}. Before that, however, we describe how this battery of eight capability dimensions (Table~\ref{tab:clustered_capabilities}) will become the basis for capability profiling.

\newcolumntype{D}{>{\centering\arraybackslash}p{1.5em}}
\begin{table}[h]
  \centering
  \caption{Demand-level distribution across the unified battery dimensions.}
  \label{tab:demand_levels}
  \begin{tabular}{lrDDDDD}
    \toprule
    & > 0 & \multicolumn{5}{c}{\% of active items at level} \\
    \cmidrule(l){3-7}
    Dimension & (\%) & 1 & 2 & 3 & 4 & 5 \\
    \midrule
    IIC & 100 & 23 & 61 & 15 & 1 & 0 \\
    L & 100 & 15 & 82 & 3 & 0 & 0 \\
    SM & 99 & 19 & 64 & 9 & 7 & 0 \\
    APaS & 94 & 54 & 35 & 11 & 0 & 0 \\
    IR & 92 & 50 & 43 & 7 & 0 & 0 \\
    EM & 50 & 44 & 50 & 4 & 2 & 0 \\
    SC & 42 & 22 & 70 & 6 & 2 & 0 \\
    OP & 36 & 42 & 45 & 6 & 7 & 0 \\
    \bottomrule
  \end{tabular}
  \vspace{0.5em}
\parbox{0.9\linewidth}{\footnotesize
\textit{Note.} Merged demands on the 19,535-item battery, rounded to integer levels. Coverage is reported as the \% of items with demand $>0$; the remaining columns give the distribution of demand level among items where the dimension is active. Dimensions: Information Integration \& Control (IIC), Language (L), Semantic Memory (SM), Action Planning \& Simulation (APaS), Instrumental Reasoning (IR), Episodic Memory (EM), Social Cognition (SC), and Object Permanence (OP). Definitions are provided in Table~\ref{tab:clustered_capabilities}.}
\end{table}

\subsubsection{Profile estimation}
\label{sec:profile_estimation}

Estimating a capability profile requires inferring, from a binary record of which battery items an agent passed, how far its competence extends along each capability dimension. We use Measurement Layouts~\cite{burden2023inferring}, a class of Bayesian item-response models that infer latent capabilities jointly from two inputs: an agent's observed per-item performance, and the per-item demand annotations produced in \S\ref{sec:benchmark_selection}. As items typically place demands on several dimensions, the model uses the full pattern of successes and failures across items to disentangle each dimension's contribution. Here we treat capabilities as posterior distributions, not point estimates, which allows sparse or ambiguous coverage of a dimension to be revealed as wider uncertainty.

Following \S\ref{sec:benchmark_selection}, each item $j$ carries an annotated demand vector $D_j \in \{0,\dots,5\}^K$ over the $K$ retained capability dimensions, where $D_{jk}=0$ indicates that item $j$ does not tax dimension $k$. An agent is summarised by a capability vector, which we treat as a ratio-scale quantity $\theta_k > 0$ and estimate in log-space, $c_k = \log\theta_k$, since the log scale is unconstrained and better conditioned for sampling. Each demand level is mapped to a ratio-scale difficulty through the exponential introduced in \S\ref{sec:capability_rubrics}, $\delta_{jk} = e^{\lambda D_{jk}}$, so that successive levels correspond to geometrically increasing difficulty. The agent's position on dimension $k$ relative to what item $j$ demands is then the log-ratio of capability to difficulty: that is, the \emph{margin}:

\begin{equation}\label{eq:margin}
m_{jk} = c_k - \lambda D_{jk} = \log\frac{\theta_k}{\delta_{jk}},
\quad m_{jk}=0 \text{ when } D_{jk}=0.
\end{equation}

A positive margin means the agent comfortably exceeds the item's demand on that dimension; a negative margin means it falls short; inactive dimensions contribute exactly zero and drop out. The per-dimension margins combine into a single item logit $z_j$ (below), which passes through a sigmoid to give the success probability:

\begin{equation}\label{eq:likelihood}
z_j = \alpha + \operatorname{pool}_k(m_{jk}),
\qquad Y_j \sim \mathrm{Bernoulli}\big(\sigma(z_j)\big).
\end{equation}

This is an item-response model \citep{embretson2025item} in which the latent margin is ``capability minus demand,'' and $\sigma(\cdot)$ is the item characteristic curve. To identify the model, each capability dimension enters the margin with unit weight ($\kappa=1$), and the common slope is fixed at $\lambda=1$. We set these defaults due to the constraints of single-agent profiling. With observations from a single agent, the model cannot distinguish whether differences in success arise because a dimension is more heavily weighted or because the agent has greater capability on that dimension, nor can it separately identify the overall slope from the scale of the latent capabilities and demands. Both could in principle be relaxed in a hierarchical population model where many agents share these parameters (\S\ref{app:assumptions}); we leave this to future work.

The central modelling choice is therefore how an item's several capability demands aggregate into $z_j$. Tasks differ in this respect: some are compensatory, where strength on one ability can offset a shortfall on another (one might compensate for poor prospective memory through structured planning, for instance), while others are bottlenecked, where the weakest required capability limits performance regardless of strengths elsewhere (for example, a person with excellent spatial reasoning cannot operate a vehicle effectively without the procedural memory required to execute the learned actions involved in driving). This can be seen as a single continuum. A fully compensatory baseline averages across active margins,

\begin{equation}\label{eq:pool-mean}
z_j = \alpha + \frac{1}{K_j^{\mathrm{on}}}\sum_k m_{jk},
\qquad K_j^{\mathrm{on}} = \#\{k : D_{jk}>0\},
\end{equation}

with the mean, rather than a raw sum, ensuring that logits remain comparable across items that recruit different numbers of dimensions. A soft-minimum rule with temperature $\tau$ generalises this:

\begin{equation}\label{eq:pool-softmin}
z_j = \alpha - \frac{1}{\tau}\log\!\left(\frac{1}{K_j^{\mathrm{on}}}\sum_k e^{-\tau m_{jk}}\right),
\end{equation}

which recovers mean pooling as $\tau \to 0$ and approaches the weakest-link minimum $\min_k m_{jk}$ as $\tau \to \infty$, with intermediate values interpolating between these extremes. A single parameter therefore controls the transition from ``abilities compensate for each other'' to ``the weakest ability is the bottleneck.'' Here, we use the soft-minimum pooling rule (Equation \ref{eq:pool-softmin}) and determine the appropriate temperature $\tau$ through the recovery analysis in \S\ref{sec:results}.

Equations~\eqref{eq:margin}--\eqref{eq:pool-softmin} run in the \emph{generative} direction: given an agent's capabilities and an item's demands, they assign a probability to each outcome $Y_j$. Profiling requires the reverse: inferring an agent's latent capabilities from a binary record of performance on the testing battery. An agent's capabilities can be recovered by inverting the generative model with Bayes' rule, as in the Measurement Layouts framework~\cite{burden2023inferring} (see Equation \ref{eq:posterior} in \S\ref{app:posterior}). 

For a single agent, the demand matrix and slope $\lambda$ are fixed, leaving only the $K$ log-capabilities $c$ and intercept $\alpha$ to estimate. Each $c_k$ receives an identical weakly informative $\mathcal{N}(3.0,0.8^2)$ prior, chosen to place broad mass over the ratio-scale capability range spanned by the battery without strongly favouring a particular location; $\alpha$ receives a weak $\mathcal{N}(0,5^2)$ prior (\S\ref{app:posterior}). Combining these priors with the likelihood in \eqref{eq:likelihood} gives a posterior over $c$ based only on attempted items, with missing responses contributing no likelihood terms. Because this posterior is not available in closed form, we obtain samples using MCMC (the No-U-Turn Sampler) \citep{hoffman2014no} and summarise each dimension by its posterior mean and credible interval. 

When an interpretable magnitude is required, log-capabilities are transformed to the ratio scale as $\theta_k=e^{c_k}$. Crucially, we retain the full posterior. Dimensions supported by sparse or collinear evidence remain uncertain, and this uncertainty propagates into downstream capability profiles and suitability estimates. The result is therefore not a fixed capability score, but a posterior distribution over what the agent can do that reflects the uncertainty in each capability estimate.

\subsection{Requirements weighting}
\label{sec:requirements_weighting}

The capability profiles of \S\ref{sec:capability_profiling} describe what an agent can do, but to assess suitability we also need a structured account of what each workplace task requires. As argued in \S\ref{sec:assessing_workplace}, we operationalise this through expert elicitation of the \textit{relative importance} of each core capability for a given activity. Direct annotation of task demands would be infeasible at organisational scale; instead, this subsection describes the instrument used to decompose workplace activities into their constituent capability requirements.

We administered the questionnaire to participants recruited through collaborating organisations and supplemented these responses with additional online participants to broaden coverage beyond the participating workforce. To capture the range of work activities in which AI systems are likely to be deployed, we organised responses around six job domains (Table \ref{tab:domains}) corresponding to departments within a typical product-based company. These domains span manual and physical roles (Warehouse or logistics; Manufacture, maintenance, or repair), data and analytical roles (Numerical, data, or programming), office-based organisational roles (Administration, organisational, or planning; Customer service, marketing, or HR), and client-facing roles (Hospitality, sales, or client care).

\begin{table}[h]
\centering
\caption{Job domains}
\begin{tabular}{c l l l}
\toprule
\# & Acronym & Job domain  \\
\midrule
1 & WL   & Warehouse or logistics \\
2 & MMR  & Manufacture, maintenance, or repair \\
3 & NDP  & Numerical, data, or programming \\
4 & AOP  & Administration, organisational, or planning \\
5 & CMH  & Customer service, marketing, or HR \\
6 & HSC  & Hospitality, sales, or client care \\
\bottomrule
\end{tabular}
\label{tab:domains}
\end{table}

The questionnaire was designed as a practical instrument for widespread requirements gathering. It proceeded in four stages. \textit{Demographics} recorded each participant's job domain and role experience. \textit{Task selection} asked participants to identify the five work activities most important to their role from a list of 18 general activities (Table~\ref{tab:work_activities}), adapted from O*NET work activity categories~\citep{onet2026} and chosen to be commensurable across domains. Participants then ranked their selected activities by importance and reported the hours per week spent on each, enabling importance to be distinguished from time allocation. \textit{Capability familiarisation} introduced all 18 capabilities individually (each with a definition, cross-domain examples, and an illustrative image), followed by a matching quiz that both reinforced learning and served as an attention-based quality-control check. \textit{Capability weighting} then elicited the core requirement profile: for each selected activity, participants built a ``robot helper'' by first selecting the five capabilities they judged most essential, then distributing 100 points across them to reflect relative importance (Figure \ref{fig:questionnaire_ranking}).

Constraining the capability-weighting task to the five most important capabilities for each of the five most important activities kept completion time feasible (approximately 30 minutes, including demographic and familiarisation sections), limiting fatigue while sustaining engagement. Administering familiarisation before weighting ensured that judgements were informed rather than purely intuitive, partially mitigating the introspective limits of employee self-report for eliciting capability importance scores.

Aggregation proceeds as follows: because participants select only their five most important activities (and, within each, five essential capabilities), unselected items are assigned a zero rather than treated as missing. Task importance is therefore \textit{frequency-adjusted}: respondents assign 5 points to their highest-ranked activity, down to 1 point for their fifth, and these scores are averaged within each domain. The same procedure applies to capability importance, with weights averaged across all respondents rather than only those who selected a given capability. This preserves both how often an item is selected and how highly it is ranked when selected. Averaging these allocations across participants yields an \textit{importance matrix} mapping capabilities to activities, with responses aggregated at multiple levels of granularity: cross-domain (Figure~\ref{fig:ability_matrix_all}), domain-specific (Figure~\ref{fig:ability_matrix_by_domain}), or company-specific. These matrices provide the activity-level weights $w_{tk}$ used in the suitability mapping of \S\ref{sec:suitability_mapping}.

In parallel, we conducted expert interviews to obtain role-specific capability profiles. Each interview produces a capability vector for a single role and, separately, for three core duties within that role, enabling analyses at a finer level of granularity than the survey's activity-level matrices (we return to the interview analysis in \S\ref{sec:company_case_study}).

\subsection{Suitability mapping}
\label{sec:suitability_mapping}

The capability profiling of \S\ref{sec:capability_profiling} yields, for each agent, a posterior over its capability vector; the requirements gathering of \S\ref{sec:requirements_weighting} yields, for each activity, a weighted profile of the capabilities it draws on. Suitability mapping combines the two into a task-by-agent score, addressing the question: how well does an agent's capability profile fit an activity's requirements?

Suitability is a weighted aggregation of an agent's capabilities using the activity-specific importance weights elicited in \S\ref{sec:requirements_weighting}. Unlike the item-response model of \S\ref{sec:profile_estimation}, it yields a comparative score rather than a calibrated probability of success.

For activity $t$, let $r_t \in \mathbb{R}^{K}$ denote its row of the importance matrix, containing the mean capability allocations from the requirements questionnaire. These allocations are transformed into a weight vector by applying an optional sharpening exponent $s \geq 1$ and normalising to sum to one:

\begin{equation}
w_{tk} = \frac{r_{tk}^{\,s}}{\sum_{k'} r_{tk'}^{\,s}},
\label{eq:weights}
\end{equation}

The resulting score is therefore a weighted mean that is invariant to the raw magnitude of the importance row, with capabilities assigned zero importance receiving zero weight. The exponent $s \geq 1$ optionally controls the \emph{sharpness} of the weighting: $s=1$ preserves the elicited importance weights, while larger values increasingly concentrate weight on the capabilities an activity emphasises most. Throughout this work we use the default $s=1$, although \S\ref{sec:suitability_stability} examines the effect of varying this parameter.\footnote{An optional preprocessing step is to normalise each capability column across activities to $[0,1]$ before forming weights. This converts absolute into relative importance and can increase contrast when activities have broadly similar importance profiles. We do not apply this normalisation here.}

Given weights $w_{tk}$ and capability levels $\theta_{ak}=e^{c_{ak}}$ (on the ratio scale of \S\ref{sec:profile_estimation}), the suitability of agent $a$ for activity $t$ is the weighted power mean (generalised mean) of order $p$. Because capability profiles are inferred over the clustered dimensions (\S\ref{sec:benchmark_selection}) whereas importance is elicited over the original capabilities, we first expand each inferred cluster onto its constituent capabilities, assigning every constituent the capability level inferred for its cluster,

\begin{equation} 
S_{at} = \Big( \sum_{k=1}^{K} w_{tk}\, \theta_{ak}^{\,p} \Big)^{1/p}. \label{eq:suitability} 
\end{equation}

The parameter $p$ controls the degree of compensation between capabilities. Larger values allow strengths to offset weaknesses, whereas smaller values increasingly treat weak capabilities as bottlenecks. Throughout this work we use the geometric mean ($p=0$), which is the natural midpoint for ratio-scale capabilities because it averages in the model's native log scale. The effect of varying $p$ is examined in \S\ref{sec:suitability_stability}.

Capabilities are inferred as a posterior (\S\ref{sec:profile_estimation}), meaning each posterior draw is mapped independently through \eqref{eq:suitability}. Suitability is therefore itself a posterior distribution, summarised by its mean and 95\% credible interval. Uncertainty in capability estimates propagates directly into downstream suitability scores. We additionally propagate uncertainty in the elicited importance weights by sampling weight vectors from a Dirichlet distribution centred on the estimated profile, $w_t \sim \mathrm{Dirichlet}(\kappa_t \bar{w}_t)$, where the concentration parameter $\kappa_t$ controls confidence in the elicited weights.

The parameter $p$ controls how capability strengths and weaknesses combine in the overall suitability score. When $p=1$, suitability is the weighted arithmetic mean, so a higher level in one capability can offset a lower level in another in direct proportion (compensatory). As $p$ decreases, the score becomes increasingly sensitive to low capability levels. In the limit as $p\to-\infty$, suitability is determined by the lowest capability level among those with positive weight (non-compensatory). As $p$ increases above 1, high capability levels have progressively greater influence, allowing a standout strength to dominate the score (ultra compensatory). Thus, lower values of $p$ are appropriate when an activity requires competence across all important capabilities, whereas higher values are appropriate when strong performance in one capability can compensate for weakness in others. 

The parameter $p$ affects how capability levels are aggregated, while the sharpness parameter $s$ affects how concentrated the capability weights are. These represent separate assumptions: $p$ governs substitutability between capabilities, whereas $s$ governs the distribution of importance across them. Throughout this work, we use $p=0$, the weighted geometric mean, which is the natural baseline for ratio-scale capabilities because it corresponds to averaging in the model's native log coordinate while providing a weakly compensatory aggregation. Unlike the capability profiling pooling temperature $\tau$ (\S\ref{sec:profile_estimation}, Equation \ref{eq:pool-softmin}), which governs how capabilities combine to explain observed item outcomes, $p$ is a decision-stage assumption governing how capability profiles are aggregated for comparison, not a property of the underlying measurement model.

\section{Validation and testing}
\label{sec:results}

In this section, we validate the pipeline and apply it end to end. We first assess recovery of latent capabilities in simulation, then estimate capability profiles for a range of modern AI systems. We next elicit work requirements across job domains, identifying the tasks that matter most and the capabilities they draw upon. Finally, we combine capability estimates and requirement profiles to produce suitability scores for each system across work activities.

\subsection{Recovery analysis}
\label{sec:recovery_results}

Before applying the profiling method to real systems, we first establish whether the inference procedure can recover known latent capability profiles. This cannot be assessed directly on real models, whose true capabilities are unknown, so we validate the procedure using synthetic agents. We sample 20 agents from the prior with known capability profiles (Table \ref{tab:synthetic_agent_profiles}), simulate their responses on the benchmark battery, and test whether inference reconstructs the ``ground truth'' profiles. This controlled setting allows us to evaluate two modelling choices: how an item's multiple demands combine to determine success (the pooling temperature $\tau$; Equation \ref{eq:pool-softmin} in \S\ref{sec:profile_estimation}), and how capability scales are anchored across agents (the use of a per-agent or shared intercept $\alpha$). The remainder of this section presents the recovery analysis and motivates our choice of $\tau=1$ and a shared $\alpha$.

Recovery is assessed primarily using between-agent recovery $r$, which is the Pearson correlation between true and posterior-mean capability values across synthetic agents for each dimension. We additionally examine posterior contraction and log-scale mean absolute error (MAE) as complementary diagnostics of informativeness and accuracy (see \S\ref{app:recovery}).

\subsubsection{Pooling rule}\label{sec:tau_sweep}

We first sweep the soft-min pooling temperature $\tau \in \{0, 0.25, 0.5, 1, 2\}$, where $\tau = 0$ is fully compensatory (demands averaged) and larger values increasingly approximate weakest-link pooling, so performance is driven by the hardest demand. For each value of $\tau$, we adjust the simulation intercept $\alpha$ so that mean simulated accuracy remains fixed at $0.60$. This allows us to examine the effect of the pooling rule without confounding it with changes in the simulation’s overall difficulty. 

\begin{table}[h]
\centering
\caption{Parameter-recovery $r$ per battery dimension across the soft-min temperature $\tau$.}
\label{tab:tau_sweep}
\begin{tabular}{lccccc}
\toprule
Dimension & $\tau{=}0$ & 0.25 & 0.5 & 1 & 2 \\
\midrule
IIC (100\%) & 0.61 & 0.63 & 0.73 & 0.83 & \textbf{0.90} \\
L (100\%) & 0.67 & 0.81 & 0.84 & \textbf{0.90} & 0.86 \\
SM (99\%) & 0.73 & 0.84 & 0.90 & \textbf{0.92} & 0.91 \\
APS (94\%) & 0.86 & 0.87 & 0.86 & 0.86 & \textbf{0.91} \\
IR (92\%) & \textbf{0.89} & 0.85 & 0.80 & 0.76 & 0.72 \\
EM (50\%) & 0.82 & \textbf{0.83} & 0.78 & 0.80 & 0.79 \\
SC (42\%) & 0.81 & 0.81 & \textbf{0.84} & 0.82 & 0.82 \\
OP (36\%) & \textbf{0.89} & 0.85 & 0.85 & 0.84 & 0.88 \\
\midrule
Mean & 0.79 & 0.81 & 0.82 & 0.84 & \textbf{0.85} \\
Worst dimension & 0.61 & 0.63 & 0.73 & \textbf{0.76} & 0.72 \\
\bottomrule
\end{tabular}

\vspace{0.5em}
{\raggedright\footnotesize\textit{Note.} A fixed population of synthetic agents is simulated and re-fit self-consistently at each $\tau$. The simulation intercept is recalibrated so mean accuracy remains $0.60$ (item difficulty held constant), isolating pooling effects from changes in overall difficulty. $\tau{=}0$ corresponds to the normalized-additive (compensatory) model; larger $\tau$ values move toward weakest-link pooling. Bold indicates the highest recovery in each row. Dimensions: Information Integration \& Control (IIC), Language (L), Semantic Memory (SM), Action Planning \& Simulation (APS), Instrumental Reasoning (IR), Episodic Memory (EM), Social Cognition (SC), Object Permanence (OP); item coverage in brackets.\par}
\end{table}

Under the compensatory baseline ($\tau = 0$) the three most frequently required dimensions -- Information Integration \& Control, Language, and Semantic Memory -- show the weakest recovery ($r = 0.61, 0.67, 0.73$; Table~\ref{tab:tau_sweep} and Figure~\ref{fig:recovery_scatter}). Because these dimensions contribute to almost every item, the data rarely isolate their individual contributions, making them difficult to distinguish from the other capabilities recruited alongside them. Increasing $\tau$ substantially improves recovery of these dimensions: by $\tau = 1$, recovery rises to $0.83$, $0.90$, and $0.92$, respectively (Table~\ref{tab:tau_sweep}). The main trade-off is a gradual decline in Instrumental Reasoning recovery ($0.89 \to 0.76$). Although mean recovery increases slightly further at $\tau = 2$, the worst-recovered dimension is recovered best at $\tau = 1$ ($r=0.76$) before Instrumental Reasoning deteriorates further. We therefore adopt $\tau = 1$ as the best compromise.

\subsubsection{Intercept treatment}\label{sec:alpha_choice}

The recovery analysis also motivates the second modelling choice introduced above: how the intercept $\alpha$ should be treated. Because the pooling rule depends only on the margins $c_k - \lambda D_{jk}$, the data cannot distinguish a higher overall capability level from a lower intercept; increasing every capability and decreasing $\alpha$ by the same amount produces exactly the same predictions. The battery therefore directly identifies the \emph{shape} of an agent's capability profile (its relative strengths and weaknesses), whereas its \textit{overall level} is identified only relative to the intercept. This suggests two alternatives: estimate a separate intercept for each agent, or estimate a single intercept shared across agents that anchors the common capability scale. Table~\ref{tab:alpha_analysis} compares these alternatives.

\begin{table}[h]
  \centering
    \caption{Recovery of capability level and shape, within-agent posterior collinearity, and downstream suitability recovery, under alternative intercept treatments ($\tau = 1$).}
  \label{tab:alpha_analysis}
  \begin{tabular}{lccc}
    \toprule
     & Free $\alpha$ & Shared $\alpha$ & Fixed $\alpha$   \\
    \midrule
    Recovery shape $r$ & 0.92 & 0.92 & 0.92 \\
    Recovery level $r$ & 0.12 & 0.98 & 0.98 \\
    \midrule
    Effective dims & 2.7 & 6.7 & 6.7 \\
    Within-agent $|r|$ & 0.50 & 0.13 & 0.13 \\
    Worst-dim contraction & 0.71 & 0.81 & 0.82 \\
    \midrule
    Suitability $\rho$ & 0.14 & 0.91 & 0.91 \\
    Suitability (PAcc) & 0.53 & 0.90 & 0.89 \\
    \bottomrule
  \end{tabular}

  \vspace{0.5em}
{\raggedright\footnotesize
\textit{Note.} Twenty synthetic agents were simulated and re-fit self-consistently using the soft-min choice rule ($\tau=1$); the simulation intercept was calibrated to give mean accuracy $0.60$. \emph{Shape $r$} and \emph{Level $r$} are across-agent Pearson correlations between true and recovered values: \emph{Shape} measures recovery of each agent's capability contrasts (deviations from its mean), and \emph{Level} its mean capability across dimensions. The middle block summarises within-agent posterior geometry: \emph{Effective dims}, participation ratio of the posterior capability-correlation eigenvalues (maximum $8$; higher indicates more independently identified capabilities); \emph{Within-agent $|r|$}, mean absolute off-diagonal posterior correlation (lower indicates less confounding); \emph{Worst-dim contraction} $=\min_k[1-\mathrm{Var}_{\text{post}}(c_k)/\sigma_c^2]$, where $\sigma_c^2$ is the prior variance (higher indicates greater information gain). The lower block reports recovery of downstream suitability metrics (\S\ref{sec:suitability_mapping}), averaged over elicited task-importance profiles (Spearman's $\rho$; PAcc = pairwise accuracy, chance $=0.5$). Columns: \emph{Free}, independent intercept per agent; \emph{Shared}, one intercept estimated jointly across agents; \emph{Fixed}, intercept fixed to its true simulated value (oracle upper bound).\par}
\end{table}

Across all three treatments, profile shape is recovered equally well ($r = 0.92$), indicating that the relative capability profile is identified regardless of how the intercept is handled. However, estimating a separate intercept for each agent leaves the overall capability level effectively unanchored ($r = 0.12$), whereas sharing a single intercept across agents anchors the common capability scale, increasing level recovery to $0.98$. The shared-intercept model also yields a much better conditioned posterior (6.7 versus 2.7 effective dimensions; mean within-agent $|r| = 0.13$ versus $0.50$) and substantially improves downstream suitability recovery ($\rho = 0.91$ versus $0.14$; pairwise accuracy $0.90$ versus $0.53$). The shared-intercept model performs essentially identically to the oracle analysis in which $\alpha$ is fixed to its true simulated value, indicating that it recovers nearly all of the identifiable information without requiring knowledge of the ground truth. 

We therefore adopt soft-min pooling with $\tau = 1$ and a shared intercept in all subsequent analyses. More generally, the modelling choices in this recovery analysis were selected to maximize the identifiability of capability profiles on the present benchmark. However, the same modelling parameters are used for data simulation and inference, and thus the analysis establishes internal recoverability, but it does not establish the empirical validity of those modelling assumptions. Assessing whether performance on \textit{real tasks} follows the same pooling behaviour is left to future work (\S\ref{sec:discussion}).

\subsection{AI capability profiles}
\label{sec:capability_results}

\begin{figure*}[ht]
\centering
\includegraphics[width=0.85\linewidth]{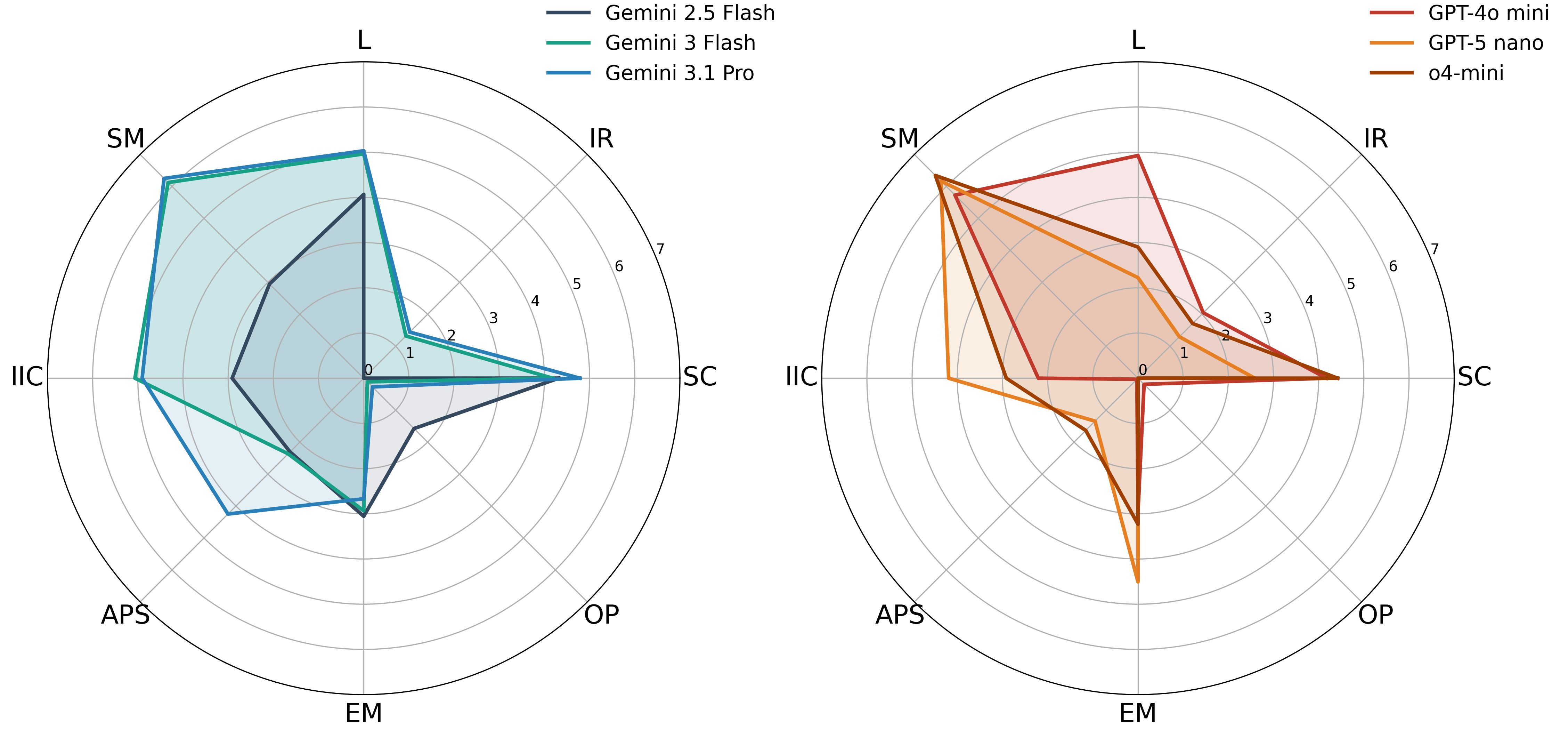}
\vspace{6pt}
\caption{Capability profiles by model family. Posterior-mean log-capability ($c_k = \log\theta_k$) across the eight clustered battery dimensions for six LLMs, grouped by developer: Google (Gemini 2.5 Flash, Gemini 3 Flash, Gemini 3.1 Pro; left) and OpenAI (GPT-4o-mini, GPT-5-nano, o4-mini; right). These profiles were produced using soft-min pooling ($\tau=1$) and a shared intercept across systems. Axes: SC = Social Cognition, IR = Instrumental Reasoning, L = Language, SM = Semantic Memory, IIC = Information Integration \& Control, APS = Action Planning \& Simulation, EM = Episodic Memory, OP = Object Permanence. Spokes show posterior means only; per-dimension uncertainty (95\% HDIs) is given in the forest plots (Figure \ref{fig:forest_grid}).}
\label{fig:family_radar}
\end{figure*}

Using the model defined in \S\ref{sec:profile_estimation} and the parameters identified in \S\ref{sec:recovery_results}, we estimated capability profiles across the eight clustered dimensions (\S\ref{sec:benchmark_selection}) for six AI systems from two developer families (Google and OpenAI). Figure~\ref{fig:family_radar} displays the profiles by family, with per-dimension uncertainty given in the forest plots of Figure~\ref{fig:forest_grid}. Table~\ref{tab:by_capability} displays performance and inferred capability estimates by dimensions across the six systems, while Tables~\ref{tab:by_system}--\ref{tab:capability_breakdown} report the full posterior estimates for each system.

The two Gemini 3 models exhibit the highest overall capability levels: Gemini 3.1 Pro ($\bar{c}=3.70$) and Gemini 3 Flash ($\bar{c}=3.38$), while the remaining four systems cluster at similar levels ($\bar{c}\approx2.6$--$2.9$). Despite these differences in overall level, all six systems share a remarkably similar profile shape. Variation between capability dimensions substantially exceeds variation between systems (Table~\ref{tab:by_capability}).

\begin{table}[h]
  \centering
  \caption{Benchmark and inference summaries for each capability dimension across the six systems.}
  \label{tab:by_capability}
  \begin{tabular}{lrrrrrrrr}
    \toprule
     & Accuracy & Coverage & $\bar c$ & $SD$ & $\max|r|$  \\
    \midrule
    SM  & 59.0\% & 99\% & 5.59 & 0.37 & 0.27  \\
    SC  & 65.2\% & 42\% & 4.08 & 0.42 & 0.23  \\
    L   & 58.8\% & 100\% & 4.02 & 0.38 & 0.40  \\
    IIC & 58.8\% & 100\% & 3.70 & 0.38 & 0.33  \\
    EM  & 56.0\% & 50\% & 3.23 & 0.39 & 0.31  \\
    APS & 57.3\% & 94\% & 1.99 & 0.19 & 0.61 \\
    IR  & 56.8\% & 92\% & 1.22 & 0.15 & 0.66  \\
    OP  & 45.8\% & 36\% & 0.29 & 0.15 & 0.73 \\
    \bottomrule
  \end{tabular}
  
  \vspace{0.5em}
{\raggedright\footnotesize
\textit{Note.} \emph{Accuracy} and \emph{Coverage} describe the battery: \emph{Accuracy} is mean accuracy on items requiring that capability ($D>0$), and \emph{Coverage} is the fraction of items requiring that capability. The remaining columns summarise the inferred latent capability estimates: $\bar c$, mean posterior log-capability across systems; $SD$, mean posterior standard deviation; and $\max|r|$, the maximum absolute posterior correlation with any other capability (mean over systems), with values approaching $1$ indicating poorer identifiability.\par}
\end{table}

Across the catalogue, Semantic Memory, Language, and Social Cognition are consistently the strongest capabilities, reflecting the language-based communication, factual recall, and interpersonal interaction at which contemporary chatbots excel. Semantic Memory is the highest-scoring dimension for five of the six systems ($c\approx5.7$--$6.3$); the exception is Gemini 2.5 Flash, which instead peaks on Social Cognition, with Language close behind. At the opposite end of the profiles, Action Planning \& Simulation, Instrumental Reasoning, and Object Permanence are consistently the weakest dimensions. Current systems thus remain strongest in capabilities centred on knowledge, language, and social understanding, whereas planning and reasoning about actions and objects remain their principal weaknesses.

The Gemini 3 models distinguish themselves not through uniformly higher capability, but through selective improvements in planning and cognitive control (Table~\ref{tab:capability_breakdown}). Their largest advantage is in Information Integration \& Control, where Gemini 3 Flash ($5.06$) and Gemini 3.1 Pro ($4.90$) separate clearly from the remainder of the catalogue ($c\approx2$--$4$). The gap is larger still in Action Planning \& Simulation: Gemini 3.1 Pro reaches $4.25$, compared with $2.37$ for the next-highest system (Gemini 3 Flash). By contrast, knowledge- and language-related capabilities differ relatively little across models. Overall, what distinguishes the strongest systems is not superior factual knowledge or communication ability, but stronger planning, integration, and control. Even so, Object Permanence and Instrumental Reasoning remain among their weakest capabilities, suggesting that recent progress toward agentic behaviour has been driven more by higher-level coordination than by robust reasoning about objects and their interactions.

\subsection{Task importance}
\label{sec:questionnaire_results}

Having estimated capability profiles for our catalogue of agents, we now turn to the employee questionnaire and interviews (\S\ref{sec:requirements_weighting}), which define how these capabilities should be weighted when compared against workplace tasks, at differing levels of granularity.

\subsubsection{Participant sample and demographics}
\label{sec:participant_sample}

Our human questionnaire sample consisted of participants recruited directly through collaborating companies ($N=125$), alongside a larger supplementary pool recruited online through the platform \textit{Prolific} ($N=414$). In total, $539$ questionnaire responses were collected across the six job domains (Table \ref{tab:domains}), of which $410$ remained after quality control.\footnote{Responses were retained only if participants completed the questionnaire, scored $\ge 50\%$ on the capability quiz, and spent $\ge 10$ minutes completing the questionnaire; these filters removed 59, 66, and 4 respondents respectively.} Table~\ref{tab:participants} provides the distribution of responses across domains, and across company and online recruitment sources.

Demographically, respondents averaged $40.3$ years of age, were $49.8\%$ female, and typically held a degree (Figure~\ref{fig:questionnaire_demographics}, Table~\ref{tab:demographics}). Experience was right-skewed, with respondents averaging $5.7$ years in their current role but $11.8$ years in their wider field. Attitudes toward AI leaned mildly positive (mean $\approx 58/100$), with substantial variation across respondents. The company and online sub-samples were broadly comparable, although company respondents were slightly younger and more likely to be female.

\subsubsection{Sample validation}
\label{sec:sample_validation}

Recruitment sources were unevenly distributed across job domains (Table~\ref{tab:participants}), with manual-physical occupations in particular represented almost entirely by the online sample. Before combining the two sources, it is therefore important to establish that the recovered capability requirements reflect the work itself not differences in the respondent sample. We validated this by comparing the task~$\times$~capability importance matrices derived from the company and online samples across the 16 retained questionnaire capabilities (Figure~\ref{fig:ability_matrix_all}). The eight clustered dimensions are only used later when mapping to capability profiles (\S\ref{sec:requirements_weighting}).

Agreement was assessed using two complementary measures. Pearson correlation tests whether the two samples assign similar \emph{relative} importance to the capabilities, independent of differences in their overall rating level. Cosine similarity instead measures agreement in the complete task--capability importance profiles, retaining differences in both relative and absolute importance. As the matrices are selection-weighted, capabilities not selected for an activity are assigned a zero instead of a missing value, making cosine similarity an appropriate measure of overall profile agreement.

Agreement is high on both measures (Table~\ref{tab:validation_summary}). When comparing activities, the two samples show very similar capability-importance profiles (mean cosine $0.91$, Pearson $0.80$), indicating that the same activities are judged to rely on similar combinations of capabilities. When comparing capabilities, agreement is somewhat lower (cosine $0.85$, Pearson $0.52$), primarily because capabilities that are selected only rarely are estimated from fewer observations. More frequently selected capabilities show near-perfect agreement between the two samples (Table~\ref{tab:validation_combined}).

The remaining differences are consistent with measurement reliability rather than systematic disagreement between the samples. A split-half noise ceiling (Table~\ref{tab:validation_summary}) supports this interpretation. Within-source reliability is high (up to $0.94$), and the observed between-source Pearson correlation reaches the expected ceiling (disattenuated $r \approx 1$). Thus, the company and online samples agree about as closely as two random halves of the same sample would. The small residual differences are concentrated in sparsely sampled capabilities, with some additional variation likely reflecting genuine differences in how broad activity labels are interpreted across domains. Overall, the recovered capability map replicates robustly across recruitment sources in both structure and detail, justifying the pooling of both samples for all subsequent analyses.

\subsubsection{Which tasks matter, by domain}
\label{sec:task_importance}

Ranking activities by frequency-adjusted importance (see \S\ref{sec:requirements_weighting}) reveals a common core of activities that are important across all six domains, alongside activities that distinguish particular occupations (Figure~\ref{fig:task_importance_lolipop}). We denote the resulting task-importance score for activity $t$ by $I_t$. Averaged across domains, the highest-ranked activities are Problem solving ($I_t=2.04$), Decision making ($1.78$), Checking ($1.44$), Researching ($1.35$), and Computer use ($0.96$). At the opposite end are Listening ($0.29$), Coding ($0.38$), Managing resources ($0.39$), and Data manipulation ($0.39$) (Table~\ref{tab:work_activities}). Because the weighting combines perceived importance with selection frequency, broadly applicable activities rank above those regarded as important only within specific occupational groups.

Problem solving and Decision making rank at or near the top in every domain, forming a domain-general core of judgement-intensive work. The principal differences lie in the supporting activities. Manual-physical occupations (WL and MMR), for example, place much greater emphasis on Checking ($2.43$ and $2.45$, compared with an overall value of $1.44$) and Tool use ($1.32$ and $1.88$, compared with $0.67$). By contrast, the numerical-digital domain (NDP) prioritises Computer use ($1.89$) and Analysing data ($1.89$), while Coding ($0.68$) and Data manipulation ($1.06$) become substantially more important than in the overall ranking. The organisational domain (AOP) uniquely elevates Long-term planning ($0.94$, compared with $0.47$ overall), whereas the client-facing domains (CMH and HSC) place much greater emphasis on Building rapport, reaching its highest value in Hospitality, Sales, and Client Care ($1.83$). Together, these departures define the distinctive capability requirements of each occupational domain.

The weekly-hours analysis (Figure~\ref{fig:task_hours_lolipop}) provides a complementary perspective. Task importance and time allocation are related but distinct: Checking ranks among the most important activities while occupying only a moderate share of the working week ($10.2$ hrs/wk), whereas Computer use accounts for the most time ($15.7$ hrs/wk) despite only moderate importance. Tool use ($14.7$ hrs/wk), Managing people ($13.4$ hrs/wk), and Admin ($10.8$ hrs/wk) likewise occupy substantial time relative to their importance rankings. Importance therefore identifies the activities that workers regard as defining their roles, whereas hours identify where effort is concentrated, highlighting complementary opportunities for AI deployment through either augmenting high-value tasks or reducing time spent on high-volume activities.

\subsubsection{Which capabilities matter, by task}
\label{sec:capability_importance}

\begin{figure}[ht]
\centering
\includegraphics[width=1\linewidth]{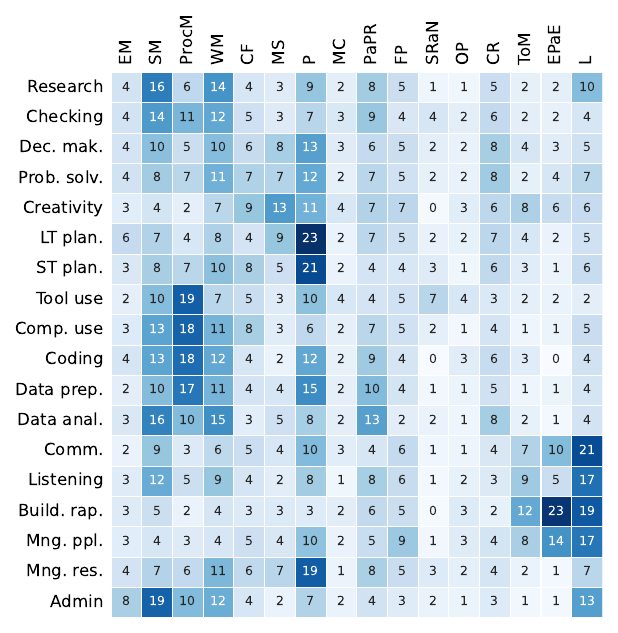}
\vspace{6pt}
\caption{Capability importance by work activity across the full sample ($N=410$). Each cell shows the frequency-adjusted importance of a cognitive capability (columns) for a work activity (rows): the mean number of points allocated from a respondent's 100-point budget for that activity, averaged across all respondents who performed it, with unselected capabilities treated as zeros. Scores therefore reflect both how frequently a capability is selected and how many points it receives when selected. Rows sum to approximately $93$ (from an original 100) across the 16 capabilities retained after the inter-rater reliability screen (Table~\ref{tab:interrater}). Full names and definitions are given in Table~\ref{tab:cognitive_capabilities} for capabilities, and in Table~\ref{tab:work_activities} for work activities.}
\label{fig:ability_matrix_all}
\end{figure}

After selecting the five most important activities in their role, participants identified the five most important capabilities for each and distributed 100 ``ability points'' across them to reflect their relative importance. Averaging these allocations across respondents (with unselected capabilities treated as zeros not missing values; \S\ref{sec:requirements_weighting}) yields the task--capability importance matrix shown in Figure~\ref{fig:ability_matrix_all}, in which each row represents the frequency-adjusted capability profile of a work activity.\footnote{Rows sum to approximately $93$ across the 16 capabilities retained after the inter-rater reliability screen (Table~\ref{tab:interrater}).}

The matrix is dominated by a shared cognitive core. Planning, Semantic Memory, Working Memory, Language, and Procedural Memory receive the greatest weight across nearly every activity, reflecting the knowledge, planning, control, and communication demands common to most work. Activities differentiate themselves primarily through the secondary capabilities they recruit: interpersonal work places greater emphasis on social cognition, analytical work on pattern recognition, and creative work on mental simulation. The capability profile of an activity is therefore best understood as a common cognitive core tuned by task-specific secondary demands.

This same structure persists across job domains (Appendix Figure~\ref{fig:ability_matrix_by_domain}). The domain-specific matrices are more alike than different, correlating cell-for-cell at $r=0.53$--$0.77$ (mean $0.63$). The common cognitive core remains intact, while the largest differences again appear in the secondary capabilities: Warehouse and Logistics places greater emphasis on Spatial Reasoning \& Navigation ($+2.1$ relative to the pooled average), Manufacture, Maintenance, and Repair on Planning ($+2.5$), and Hospitality, Sales, and Client Care on Theory of Mind ($+1.7$) and Emotion Perception \& Empathy ($+1.2$). Overall, capability requirements appear to be organised around a stable cognitive core, with occupational differences arising primarily from the secondary capabilities associated with particular forms of work.

\subsection{Task--capability mapping}
\label{sec:task_capability_mapping}

After inferring the capability profiles of a catalogue of AI systems (\S\ref{sec:capability_results}) and eliciting which capabilities matter for a range of workplace tasks (\S\ref{sec:questionnaire_results}), we now map one onto the other to estimate AI system suitability. 

\subsubsection{Suitability scores}
\label{sec:suitability_results}

The suitability mapping (Equations~\ref{eq:weights}--\ref{eq:suitability}) contains two user-specified parameters: the compensatory power $p$ and the demand sharpness $s$. Unlike the capability estimates and task-importance weights, these are not inferred from data. Instead, they specify how capabilities should be combined when judging suitability for a task, and are thus defined based on the intended use case.

\begin{figure}[h]
\centering
\includegraphics[width=1\linewidth]{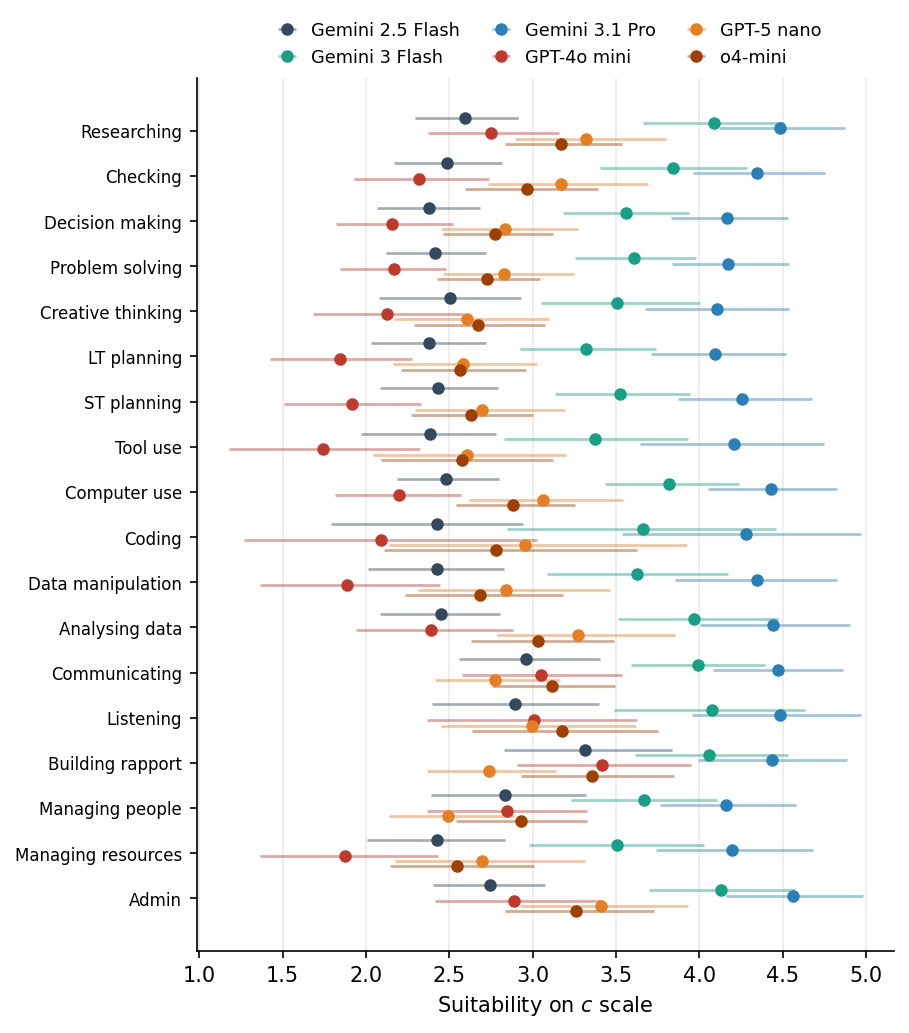}
\vspace{6pt}
\caption{Task suitability of the six AI systems across the 18 work activities on the log-capability ($c=\log\theta$) scale (points: posterior means; bars: $95\%$ credible intervals). Suitability is the importance-weighted power mean of ratio-scale capabilities ($\theta=e^c$), reported as $\log S$ (the weighted mean of $c$ at $p=0$). Estimates use the full sample ($n=410$, all domains), with sharpness $s=1$, no column normalisation, and empirical weight uncertainty ($\kappa_t\approx n_t$).}
\label{fig:suitability_scores}
\end{figure}

Throughout the main analysis we use neutral defaults: $p=0$, the weighted geometric mean, which allows modest compensation across capability dimensions, and $s=1$, which uses the elicited task-importance weights without further sharpening.\footnote{All remaining suitability-mapping settings are held at their defaults (\S\ref{sec:suitability_mapping}): capability weights are not normalised across tasks, and uncertainty in the elicited weights is propagated using a task-specific concentration parameter, $\kappa_t \approx n_t$, where $n_t$ is the number of respondents who rated task $t$.} Figure~\ref{fig:suitability_scores} shows the resulting suitability scores across the 18 work activities. Gemini 3.1 Pro is the most suitable system for every activity ($\log S\approx4.1$--$4.6$), followed consistently by Gemini 3 Flash ($\approx3.3$--$4.1$); the remaining four systems form a lower, closely overlapping group ($\approx1.7$--$3.4$). Notably, the ranking changes very little across activities. Systems differ from one another far more than activities differentiate the systems.

This stability follows directly from the shared cognitive core identified in \S\ref{sec:questionnaire_results}. Almost every activity places substantial weight on knowledge, language, planning, and cognitive control, so suitability is determined by broadly similar capability priorities regardless of the task. Systems differ relatively little in the knowledge- and language-related capabilities, but much more in Information Integration \& Control and Action Planning \& Simulation, where the Gemini~3 models hold their largest advantage (Table~\ref{tab:by_system}). As a result, the ordering changes very little from one activity to the next.\footnote{The aggregated results above use the overall task-importance profiles, but the same mapping can be applied using each domain's elicited weights, yielding domain-specific suitability estimates (Table~\ref{tab:task-domain-suitability}).}

The principal exceptions are socially oriented activities such as Building rapport, Listening, and Communicating. By placing greater weight on Social Cognition and Language than most other activities, they partially break the otherwise stable ordering of systems. GPT-4o mini, in particular, rises to the top of the mid-pack on Building rapport ($\log S\approx3.4$), reflecting its comparatively strong performance on these capabilities.

The uncertainty intervals in Figure~\ref{fig:suitability_scores} tell a complementary story. They reflect uncertainty in both the inferred capability profiles and the elicited task-importance weights. Activities supported by fewer questionnaire responses therefore produce wider intervals, most notably Coding, whose importance profile was estimated from only around 20 respondents. By contrast, well-sampled activities such as Problem solving yield correspondingly narrower intervals.

Suitability alone, however, indicates only whether a system \emph{can} perform a task, not whether automating that task would have the greatest impact. We therefore combine suitability with the task-importance score, $I_t$, from \S\ref{sec:task_importance}, defining a deployment-priority score,

\begin{equation}
P_{at}=I_tS_{at},
\label{eq:deployment_priority}
\end{equation}

which we report on the log scale as $\log P_{at}$ (Table~\ref{tab:task-domain-priority}). This reorders activities towards those that are both well suited to current AI systems and central to the role: Researching, Admin, and Analysing data emerge as the highest-priority deployment targets overall, while tasks of lower importance fall in priority regardless of their suitability. Figure~\ref{fig:automation_full} visualises the resulting trade-off between task importance and suitability, identifying activities that are both important and well suited to current AI systems. Figure~\ref{fig:automation_hours} instead replaces task importance with task frequency (hours per week), shifting the focus from work that is perceived to be most valuable to work that occupies the greatest share of employee time.

\subsubsection{Sensitivity to policy assumptions}
\label{sec:suitability_stability}

The suitability scores above use the neutral defaults ($p=0$, $s=1$), but these parameters represent deployment preferences and are not inferred from data. Alternative choices will inevitably produce different suitability scores; one question to consider is whether moderate changes in these assumptions substantially alter the overall conclusions. We therefore vary both parameters across a broad range and examine the resulting suitability rankings.

Across a wide range of settings, the suitability mapping behaves smoothly (Table~\ref{tab:suit_pxs_invariance}). On average, nearly 12 of the 15 pairwise system orderings remain unchanged, and each task exhibits only three to seven distinct rankings across the full compensatory sweep. Where reordering does occur, it is largely confined to adjacent systems whose capability profiles already overlap substantially (\S\ref{sec:capability_results}). In the present catalogue, Gemini~3.1 Pro remains the highest-ranked system for 17 of the 18 work activities, with only Admin changing under the most extreme compensatory setting ($p=2$).

The full parameter sweep (Table~\ref{tab:suit_pxs_kendall}) shows the same pattern, with substantial reordering appearing only under deliberately extreme policy settings. Overall, the suitability mapping behaves predictably across a wide range of reasonable policy choices.

\subsubsection{Single company case study}
\label{sec:company_case_study}

The analyses above aggregate questionnaire responses across all participants, but in practice the suitability mapping will be most valuable when applied using an individual organisation's own data. We illustrate this with a single company case study, anonymised as ``Company X''.

\begin{figure}[h]
\centering
\includegraphics[width=1\linewidth]{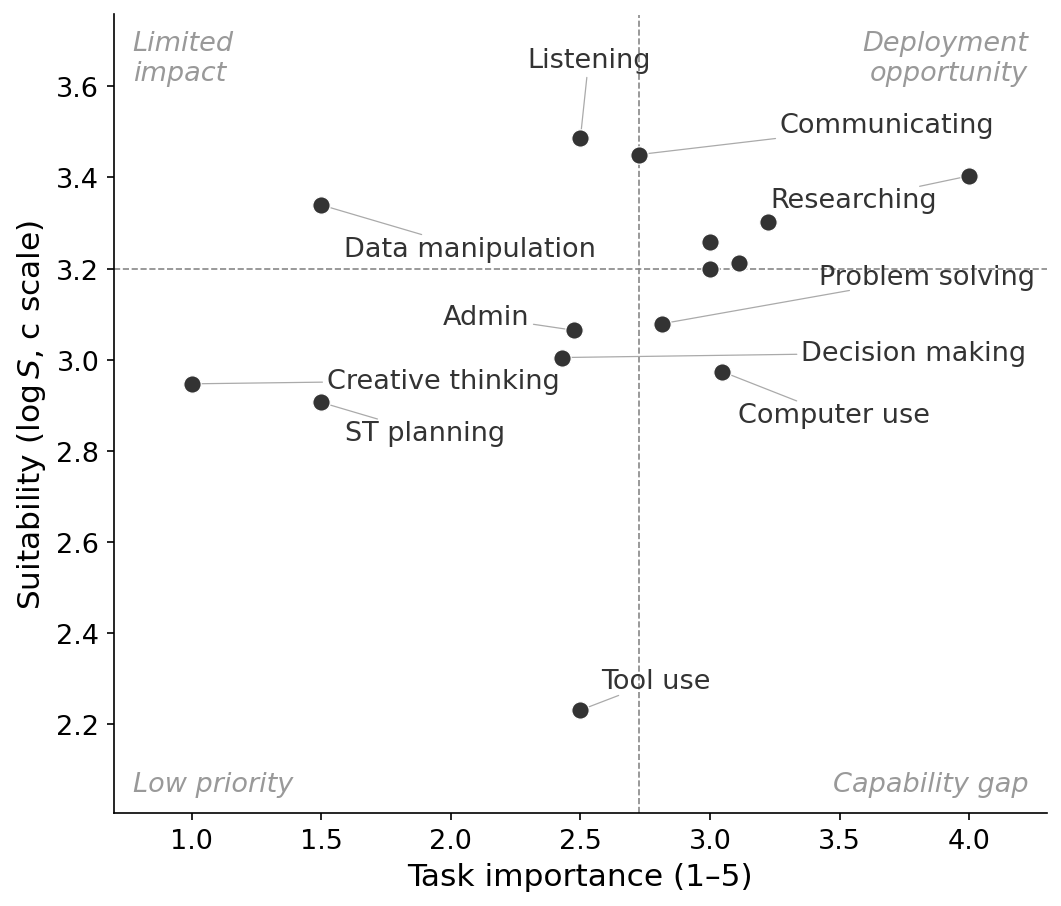}
\vspace{6pt}
\caption{Task importance (x; mean rank-derived rating, 1--5) versus mean AI suitability (y; log-capability scale) for Company X ($N=35$). Suitability is computed as the equally weighted geometric mean across the six profiled AI systems (equivalently, the arithmetic mean on the log-capability scale). Dashed lines show the medians, defining the development-opportunity (top-right), capability-gap (bottom-right), limited-impact (top-left), and low-priority (bottom-left) quadrants. Labels are shown only where a task's quadrant placement is confident (posterior probability $\geq 0.75$); unlabelled points have credible intervals that straddle the suitability median. Data manipulation is also labelled despite straddling the median because of its conspicuous position.}
\label{fig:companyX_automation}
\end{figure}

Company X contributed 35 questionnaire responses (after quality control), drawn predominantly from the \textit{Administration, Organisational, or Planning} and \textit{Customer Service, Marketing, or HR} domains. Re-estimating the task--capability importance weights, $w_{tk}$ (Equation~\ref{eq:weights}), together with the task-importance scores, $I_t$, from these responses yields deployment recommendations specific to the organisation. Figure~\ref{fig:companyX_automation} shows the resulting importance--suitability map. Relative to the full-sample analysis (Figure~\ref{fig:automation_full}), Communicating and Researching emerge as the clearest deployment opportunities for Company X, whereas Problem solving and Computer use remain high-priority tasks where much of the current AI catalogue still falls short.

For this illustration, suitability is averaged across the six profiled AI systems, providing a catalogue-level view of current AI capability; the same map can instead be generated for any individual model. The quadrant boundaries are defined by the median suitability and importance scores across the tasks shown. The quadrants are therefore \emph{relative}: `high'' and `low'' indicate whether a task falls above or below the median within this set of tasks. In practical applications, these boundaries could instead be set using policy-defined thresholds where absolute criteria for importance or suitability are preferred.

The same approach can be applied at an even finer level of granularity. As part of our collaboration with companies, we conducted structured interviews with experienced employees, who rated the importance of all 18 original cognitive capabilities both for their role overall and for three core duties within that role. Because capability weights are elicited directly for each duty, the resulting suitability estimates capture differences within a role that cannot be represented by the activity-level questionnaire. Figure~\ref{fig:interview_plot} illustrates this using a senior Customer Service employee at Company X, with duties such as `informing customers about their rights'' and `helping customers with their applications''. Together, the questionnaire and interview variants demonstrate that the framework can support deployment decisions at multiple levels of granularity, from occupational averages to individual duties within a specific organisation.

\section{Discussion}\label{sec:discussion}

This report introduced a capability-based framework for assessing the suitability of AI systems for workplace tasks. The framework combines two complementary components: \textit{cognitive capability profiling}, which infers an agent's capabilities from performance on a demand-annotated benchmark battery, and \textit{task requirements weighting}, which elicits the importance of those same capabilities for workplace tasks. Expressing both agents and tasks in terms of a shared set of cognitive constructs allows them to be compared directly, translating benchmark performance into a psychometric representation that can be mapped onto the cognitive demands of work.

The principal contribution of this work is therefore methodological. Building on \citet{zhou2026general}, we used item-level demand annotation to reorganise existing AI benchmarks around the cognitive capabilities they recruit rather than the task domains they represent, refining these demands into eight latent capability dimensions and adapting the Bayesian profiling model of \citet{burden2023inferring} to recover capability profiles from benchmark performance. In parallel, we developed a lightweight requirements-gathering procedure that organisations can use to estimate both the capabilities required for their work and the relative importance of different tasks. Together, these components produce deployment recommendations at the level of occupational domains, organisations, roles, or individual duties, while allowing capability profiling and requirements gathering to evolve independently as AI systems and workplaces change.

Applying the framework to a catalogue of contemporary AI systems revealed several substantive findings. Despite differences in overall capability level, systems exhibited remarkably similar profile \emph{shapes}, varying much more across cognitive dimensions than across model families or versions (Figure~\ref{fig:family_radar}). All performed strongly on the capabilities underpinning current chatbot performance -- language, semantic memory, and social cognition -- while remaining weakest on capabilities associated with more embodied or agentic behaviour, including planning, causal reasoning, affordance perception, and object permanence. The advantage held by the strongest systems was concentrated not in knowledge or communication, but in high-level planning and cognitive control. Future progress may therefore depend less on expanding knowledge and more on improving how knowledge is integrated, maintained, and deployed during goal-directed behaviour, while the capabilities underlying embodied agency remain comparatively underdeveloped.

On the requirements side, respondents across six occupational domains identified a similarly stable cognitive core. Semantic, procedural, and working memory, planning, and language were consistently rated as important across a wide range of workplace tasks, with interpersonal activities placing greater emphasis on social cognition and individual domains exhibiting predictable secondary specialisations. This common set of requirements spans both the capabilities shared across AI systems and those that distinguish them. Semantic memory and language are strengths of every model, whereas planning and cognitive control are the dimensions on which systems differ most. As a result, models with stronger planning and control capabilities (e.g., Gemini~3.1 Pro) achieved consistently higher suitability across almost all workplace activities, not only within particular domains. These findings should, however, be interpreted in the context of the present test battery, which samples agentic capabilities less extensively than other cognitive dimensions (\S\ref{sec:capability_results}); true differences between systems on those dimensions may therefore be larger than the present profiles indicate.

\subsection{Limitations and future directions}
\label{sec:limitations}

The principal value of this pipeline lies not in the current prototype but in the approach as a whole. The framework separates \textit{capability profiling} and \textit{requirements gathering} into distinct measurement processes, expressing both agents and workplace tasks in terms of a shared space of cognitive capabilities and demands. These constructs describe properties of the agent and the work, not the current technology. Consequently, the approach avoids extrapolating from existing AI deployments or relying on expert forecasts of future AI capabilities. As AI systems improve and workplace requirements evolve, the resulting profiles will need updating, but the framework that produces and compares them remains unchanged.

\subsubsection{Requirements measurement}

On the requirements side, the principal limitation is that the current pipeline measures task \emph{importance} not task \emph{demand}. The resulting suitability scores are therefore comparative rather than calibrated. They indicate which agents are better matched to a task, but not the probability that an agent will successfully perform it. Importance and demand are fundamentally different quantities. Extending the rubric-based demand annotation used for benchmark items (\S\ref{sec:capability_profiling}) to workplace tasks would place requirements and capabilities on the same measurement scale, allowing suitability to be interpreted as a calibrated prediction of task performance.

Such an extension would also address a limitation of the current requirements-gathering process: its reliance on human introspection. Employees were asked to identify the cognitive capabilities most important for their work, yet many cognitive processes are automatic and difficult to report accurately. Capabilities such as metacognition, spatial reasoning, and object permanence were selected relatively infrequently, plausibly because they are less salient to conscious reflection rather than because they contribute little to performance. This limitation is compounded by restricting respondents to five capabilities per activity. Demand annotation shifts these judgements away from introspection and towards structured rubric-based assessment of the work itself, mirroring the approach already adopted for capability profiling.

Even with demand annotation, however, one important challenge would remain. Both approaches ultimately describe what a task demands of a \emph{human}, whereas AI systems may reach the same outcome through different cognitive strategies. Coding provides an illustrative example. Human programmers rely heavily on planning and procedural memory, whereas contemporary AI systems often succeed through statistical pattern matching. Human-derived task requirements may therefore mischaracterise the capabilities AI systems actually rely upon.\footnote{The compensatory parameter $p$ partially accommodates this possibility by allowing strengths in some capabilities to offset weaknesses in others, making suitability estimates less sensitive to differences in strategy between humans and AI (\S\ref{sec:suitability_mapping}).} This limitation cannot be resolved from within the framework itself, but instead requires validation against observed deployment outcomes to establish where human-centred accounts of task requirements fail to generalise to non-human agents \citep{zhou2026predictable, schellaert2025analysing}.

\subsubsection{Capability measurement}

On the capability side, demand annotation provides a principled way of reorganising existing benchmark performance around cognitive constructs. Human psychometric tests assume a human participant and often transfer poorly to non-human agents, whereas conventional AI benchmarks report performance by task domain rather than the capabilities underlying that performance. Demand annotation combines the breadth and ecological validity of existing benchmarks with construct-based evaluation, allowing capability profiles to be recovered from existing testing materials.

The present implementation, however, is constrained by the available benchmark ecosystem. The benchmark battery is assembled primarily from text-based evaluations and therefore under-represents multimodal perception, long-horizon planning, and interactive tool use -- precisely the capabilities on which contemporary systems appear weakest and most differentiated (\S\ref{sec:capability_results}). Likewise, we profile foundation models in isolation, whereas practical deployments increasingly involve scaffolded agents whose external memory, planning modules, and tools compensate for limitations of the underlying model. For example, a monitor that tracks background state effectively supplies object permanence, while well-specified tooling reduces the demand on affordance perception by making available actions explicit. The resulting capability profiles therefore reflect base models in isolation, not deployed agentic AI systems. While these limitations are properties of the current benchmark battery, they are not limitations of the framework itself. As richer multimodal and agentic benchmarks become available, the capability battery can be extended accordingly, allowing the framework to characterise a broader range of cognitive capabilities while retaining the same construct-based approach.

\subsubsection{Future directions}

Beyond improving the current measurement instruments, the framework naturally extends in several directions. As AI systems become broadly capable across many cognitive dimensions, deployment decisions may depend increasingly on behavioural \emph{propensities} rather than capabilities alone. Characteristics such as risk tolerance, consistency, persistence, and responses to ambiguity are themselves latent constructs that could, in principle, be measured within the same framework alongside cognitive capabilities \citep{romero2026capabilities}.

The most consequential extension, however, is to profile human workers using the same capability framework as AI systems. Appropriate subsets of the annotated benchmark battery could serve as psychometric assessments for people, producing capability profiles directly comparable with those of AI systems. Task requirements could then be mapped onto humans and AI in exactly the same way, allowing the framework to inform evidence-based task allocation across heterogeneous workforces: identifying which work is best performed by people, by AI systems, or by combinations of the two \citep{prunty2026reverse}. Expressing both humans and AI systems within the same capability space would provide a common construct-based language for workforce planning in an era of increasingly hybrid teams.

\subsection{Conclusion}
\label{sec:conclusion}

In practice, few organisations assess AI suitability systematically. Decisions rely largely on human judgement and informal experimentation rather than structured evaluation \citep{pan2025measuring, dsit2026aiadoption}, and where empirical methods are used, they typically depend on aggregate benchmark scores that reveal little about where a system will actually fail \citep{fodor2025line, mcintosh2025inadequacies}. This report has presented an alternative. By expressing both AI systems and workplace tasks in terms of a shared space of cognitive capabilities, the framework links benchmark performance to the cognitive demands of work, producing suitability estimates that are systematic, transparent, and readily updated as models and organisations change. Because task requirements can be elicited at whatever level of granularity a decision requires, the same approach supports assessment at the level of occupational domains, organisations, roles, or individual duties.

The broader contribution, however, is methodological. Separating capability profiling from requirements gathering provides a reusable framework that is independent of any particular AI system, benchmark battery, or workplace. As both AI capabilities and patterns of work continue to evolve, the specific profiles will change, but the construct-based approach to measuring and comparing them remains. In the longer term, the same framework provides a common language for representing AI systems, human workers, and hybrid teams within a shared capability space, supporting more systematic and evidence-based workforce planning.

\subsection*{Acknowledgements}
The authors acknowledge the support of \href{https://www.accenture.com/}{Accenture} to this research. They also thank the participating organisations for their collaboration and the time generously contributed by their employees and experts in completing the questionnaires and interviews. Finally, the authors are grateful to all other respondents who participated in the broader study for their valuable contributions.

\bibliographystyle{unsrtnat}
\bibliography{main} %

\clearpage
\onecolumn
\appendix

\section{Battery and inference procedure}\label{app:battery}
\renewcommand{\thetable}{A\arabic{table}}
\renewcommand{\thefigure}{A\arabic{figure}}
\setcounter{table}{0}
\setcounter{figure}{0}

\subsection{Refining capability dimensions}
\label{app:refining}

\begin{figure}[ht]
\centering
\fbox{\includegraphics[width=0.8\linewidth]{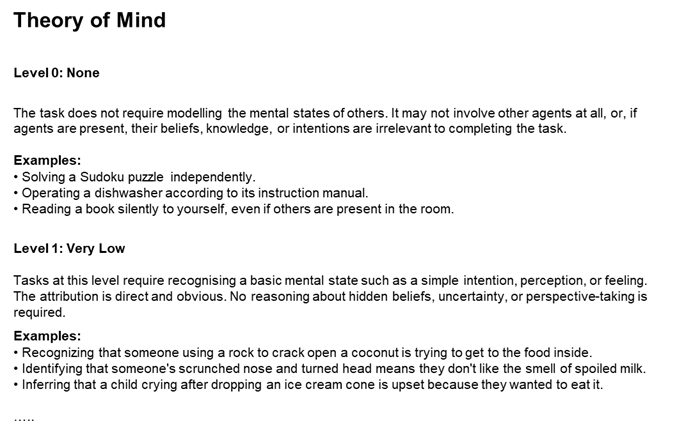}}
\vspace{6pt}
\caption{A snippet from the \textit{Theory of Mind} (ToM) rubric used to annotate benchmark items. The rubric defines how to categorise tasks from level 0 (no ToM capability required) to level 5 (very high ToM capability required), and concrete task examples are provided to help ground the demand levels. The full rubrics for all 18 capabilities are included in the \href{https://github.com/Kinds-of-Intelligence-CFI/Task-Suitability-Profiles/tree/main/Benchmarks/Annotations/rubric_files}{project repository}.}
\label{fig:rubric}
\end{figure}

Table~\ref{tab:cognitive_capabilities} lists the 18 capabilities drawn from the cognitive science literature. For each, we developed a scoring rubric (Figure~\ref{fig:rubric}) that rates a given task item by the level of demand it places on that capability. Using these rubrics, we collected two independent sets of demand annotations across the full battery (Table~\ref{tab:benchmarks}) from two LLM annotators, GPT-4o and Gemini~3 Flash.

We then refined the dimension set in two steps. First, we assessed inter-rater reliability between the two annotators (Table~\ref{tab:interrater}) and excluded the two dimensions on which they failed to agree reliably (Attention and Inhibitory Control, and Prospective Memory). Second, to address redundancy among the remaining dimensions, we examined the correlations between their demand profiles and performed dimensionality reduction, clustering together dimensions that tended to co-occur across the battery (Figure~\ref{fig:demand_correlations}). The resulting eight dimensions are summarised in Table~\ref{tab:clustered_capabilities}.

\begin{table}[ht!]
\centering
\caption{Core cognitive capabilities}
\label{tab:cognitive_capabilities}
\renewcommand{\arraystretch}{1.25}
\begin{tabular}{@{}p{0.3\linewidth} p{0.07\linewidth} >{\footnotesize}p{0.6\linewidth}@{}}
\toprule
\textbf{Capability} & & \textbf{Definition} \\
\midrule
\multicolumn{3}{@{}l}{\textit{Memory Systems}} \\
Episodic Memory & EM & Remembering previous events \\
Semantic Memory & SM & Remembering facts and information \\
Procedural Memory & ProcM & Remembering how to perform learned tasks or skills \\
Prospective Memory & ProsM & Remembering to do what you had planned to do \\
\addlinespace[8pt]
\multicolumn{3}{@{}l}{\textit{Executive Control}} \\
Working Memory & WM & Holding multiple pieces of information in your mind at once \\
Attention and Inhibitory Control & AaIC & Controlling behaviours or thoughts to focus on the task at hand \\
Cognitive Flexibility & CF & Switching between tasks or adapting to changing circumstances \\
Planning & P & Mapping out a strategy or a sequence of actions to achieve a goal \\
\addlinespace[8pt]
\multicolumn{3}{@{}l}{\textit{Object and Space Understanding}} \\
Perception and Pattern Recognition & PaPR & Using prior experience to notice relevant details about people, objects or data \\
Functional Perception & FP & Recognising the appropriate roles that objects, people, or information can play in achieving a goal \\
Spatial Reasoning and Navigation & SRaN & Reasoning about size, space and distance and how you should move from one location to another \\
Object Permanence & OP & Realising that people and objects continue to exist and have impact even if you cannot see them \\
\addlinespace[8pt]
\multicolumn{3}{@{}l}{\textit{Social and Communicative}} \\
Theory of Mind & ToM & Reasoning about the goals, beliefs and desires of others \\
Emotion Perception and Empathy & EPaE & Considering and connecting with the feelings and emotions of others \\
Language & L & Using language to communicate information between you and another person \\
\addlinespace[8pt]
\multicolumn{3}{@{}l}{\textit{Domain-general}} \\
Mental Simulation & MS & Imagining possible future scenarios that might result from different actions \\
Metacognition & M & Thinking about or assessing your own thoughts, ability or performance \\
Causal Reasoning & CR & Understanding that an outcome is the result of a previous action or event \\
\bottomrule
\addlinespace[2pt]
\multicolumn{3}{@{}p{\linewidth}@{}}{\footnotesize\textit{Note.} Mental Simulation, Metacognition, and Causal Reasoning are domain-general capacities recruited across families and are not assigned to a single grouping.} \\
\end{tabular}
\end{table}

\begin{table}[t]
\centering
\begin{threeparttable} %
\caption{Inter-rater reliability of demand annotations between GPT-4o and Gemini~3}
\label{tab:interrater}
\begin{tabular}{@{}ll @{\hspace{2em}} S[table-format=-1.2]S[table-format=1.2]S[table-format=1.2]S[table-format=2.0]@{}}
\toprule
Capability & Acronym & {$\rho$} & {$\kappa_w$} & {MAE} & {\%$_{\pm1}$} \\
\midrule
Theory of Mind & ToM & 0.81 & 0.68 & 0.45 & 88 \\
Emotion Perception and Empathy & EPaE & 0.72 & 0.72 & 0.16 & 97 \\
Spatial Reasoning and Navigation & SRaN & 0.72 & 0.71 & 0.53 & 89 \\
Causal Reasoning & CR & 0.67 & 0.59 & 0.53 & 94 \\
Procedural Memory & ProcM & 0.65 & 0.65 & 0.64 & 82 \\
Object Permanence & OP & 0.54 & 0.51 & 0.49 & 85 \\
Planning & P & 0.53 & 0.56 & 0.51 & 89 \\
Episodic Memory & EM & 0.50 & 0.44 & 0.53 & 87 \\
Semantic Memory & SM & 0.47 & 0.46 & 0.74 & 85 \\
Language & L & 0.46 & 0.40 & 0.50 & 96 \\
Metacognition & MC & 0.44 & 0.40 & 0.69 & 90 \\
Perception and Pattern Recognition & PaPR & 0.43 & 0.40 & 0.68 & 87 \\
Functional Perception & FP & 0.41 & 0.45 & 0.81 & 82 \\
Cognitive Flexibility & CF & 0.40 & 0.40 & 0.85 & 74 \\
Working Memory & WM & 0.39 & 0.33 & 0.93 & 77 \\
Mental Simulation & MS & 0.38 & 0.37 & 0.82 & 78 \\
\midrule
\multicolumn{6}{@{}l}{\textit{Excluded}} \\
Attention and Inhibitory Control & AaIC & 0.09 & 0.07 & 0.87 & 79 \\
Prospective Memory & ProsM & -0.03 & 0.15 & 0.66 & 82 \\
\bottomrule
\end{tabular}

\begin{tablenotes}[flushleft]
\footnotesize
\item \textit{Note.} Annotations by GPT-4o and Gemini~3 on the 19,531 items rated by both; items with an invalid response from either rater were dropped, giving fewer than the 19,576 in Table~\ref{tab:benchmarks}, and four fewer than the 19,535-item battery, which requires valid ratings only on the 16 retained capabilities. $\rho$: Spearman rank correlation; $\kappa_w$: quadratic-weighted Cohen's kappa; MAE: mean absolute error (in demand levels); \%$_{\pm1}$: \% of items where raters agree to within one level. Lower-panel capabilities are excluded given raters did not reliably agree ($\rho<0.30$ and $\kappa_w<0.30$).
\end{tablenotes}
\end{threeparttable}
\end{table}

\begin{table}[ht]
\centering
\caption{Clustered capability dimensions}
\label{tab:clustered_capabilities}
\vspace{6pt}
\small
\begin{tabular}{p{3cm}lp{5.5cm}p{5.5cm}}
\toprule
\textbf{Capability} & \textbf{} & \textbf{Definition} &
\textbf{Constituent abilities} \\
\midrule
Episodic Memory & EM & Remembering previous events & --- \\
\addlinespace
Semantic Memory & SM & Remembering facts and information & --- \\
\addlinespace
Object Permanence & OP & Realising that people and objects continue to exist and have impact even if you cannot see them & --- \\
\addlinespace
Language & L & Using language to communicate information between you and another person & --- \\
\addlinespace
Social Cognition & SC & Understanding and reasoning about emotions,
beliefs, desires and social cues in others & \textit{Emotion Perception \&
Empathy} (EPaE): Considering and connecting with the feelings and emotions of
others; \textit{Theory of Mind} (ToM): Reasoning about the goals, beliefs
and desires of others \\
\addlinespace
Instrumental Reasoning & IR & Reasoning about how objects, actions, and events can bring about desired outcomes & \textit{Causal Reasoning} (CR): Inferring how outcomes arise from prior actions or events; \textit{Functional Perception} (FP): Recognising the appropriate roles that objects, people, or information can play in achieving a goal \\
\addlinespace
Information Integration \& Control & IIC & Combining and manipulating relevant information about yourself and the environment in service of the current task & \textit{Working Memory} (WM): Holding multiple pieces of information in your mind at once; \textit{Cognitive Flexibility} (CF): Switching between tasks or adapting to changing
circumstances; \textit{Metacognition} (MC): Thinking about or assessing your own
thoughts, ability or performance; \textit{Perception \& Pattern Recognition}
(PaPR): Detecting structure and relevant regularities in people,
objects or data \\
\addlinespace
Action Planning \& Simulation & APaS & Planning, simulating, and reasoning about future sequences of actions & \textit{Planning} (P): Mapping out a strategy or a sequence of actions to achieve a goal; \textit{Procedural Memory} (ProcM): Remembering how to perform learned tasks or skills; \textit{Mental Simulation} (MS): Imagining possible future scenarios that might result from different actions; \textit{Spatial Reasoning \& Navigation} (SRaN): Reasoning about size, space and distance and how you should move from one location to another \\
\bottomrule
\addlinespace[6pt]
\multicolumn{4}{@{}p{\linewidth}@{}}{\footnotesize\textit{Note.} The eight capability dimensions used for profiling. Inter-rater reliability screening and dimensionality reduction produced eight capabilities from the original 18. The capabilities that were not clustered retain their original definitions, while the definitions of clustered capabilities are synthesised from their constituent abilities.}\\
\end{tabular}
\end{table}

\subsection{Posterior for capability estimation}
\label{app:posterior}

Profile estimation (\S\ref{sec:profile_estimation}) inverts the generative model of Equations~\eqref{eq:margin}--\eqref{eq:pool-softmin} with Bayes' rule. For a single agent the demand matrix $D$ and the slope $\lambda$ are fixed; the unknowns are the $K$ log-capabilities $c$ and the intercept $\alpha$. With a $\mathcal{N}(\mu_c, \sigma_c)$ prior on each $c_k$ and a weak $\mathcal{N}(0, \sigma_\alpha)$ prior on $\alpha$, the posterior is

\begin{equation}\label{eq:posterior}
p(c, \alpha \mid Y, D) \;\propto\;
\underbrace{\prod_{j} \sigma(z_j)^{Y_j}\big(1-\sigma(z_j)\big)^{1-Y_j}}_{\text{likelihood}}
\;\cdot\;
\underbrace{\Big[\textstyle\prod_k \mathcal{N}(c_k \mid \mu_c, \sigma_c)\Big]\,\mathcal{N}(\alpha \mid 0, \sigma_\alpha)}_{\text{prior}},
\end{equation}

where each item logit $z_j = z_j(c, \alpha)$ is the pooled margin of Equation~\eqref{eq:pool-softmin} and the likelihood is the Bernoulli observation model of Equation~\eqref{eq:likelihood}. The product runs only over items the agent attempted, so missing responses drop out. The sigmoid likelihood and the pooling nonlinearity admit no closed-form posterior, so we draw samples by MCMC using the No-U-Turn Sampler, and summarise each dimension by the posterior over $\theta_k = e^{c_k}$.

\subsection{Modelling assumptions and extensions}
\label{app:assumptions}

The capability-inference model in \S\ref{sec:profile_estimation} makes several assumptions to keep its parameters identifiable. These concern the demand slope, discrimination weights, and intercept. All arise from the same issue. Data from a single agent cannot separately identify parameters that trade off against its capability estimates.

First, the slope $\lambda$, which determines how demand affects the success logit (Equations~\ref{eq:margin}--\ref{eq:pool-softmin}), is shared across capability dimensions. This assumes that a one-level increase in demand has the same effect in every dimension. In principle, dimensions such as Object Permanence and Language could have different slopes $\lambda_k$. For a single agent, however, a dimension-specific slope cannot be separated from capability $c_k$, since the model depends on the margin $m_{jk} = c_k - \lambda_k D_{jk}$. Changes in $\lambda_k$ can therefore trade off against changes in $c_k$. The agent's responses cannot distinguish a steeper demand scale from greater capability on that dimension, so we use a single shared $\lambda$.

Second, each dimension enters the item logit with a fixed discrimination weight of one. A freely estimated discrimination parameter would similarly be confounded with the capability scale and $\lambda$ in single-agent data. Fixing discrimination to one, as in a Rasch-type model, keeps capabilities on a common and interpretable scale. The trade-off is that differences in how strongly performance depends on each dimension are absorbed into the capability estimates rather than modelled separately.

Third, the intercept $\alpha$ must provide a common reference point across agents. Because the pooling rule depends only on the margins $c_k - \lambda D_{jk}$, shifting all capabilities for an agent by a constant and offsetting that shift through its intercept leaves the predictions unchanged. A free intercept for each agent therefore makes its \emph{overall capability level} unidentified, although its \emph{profile shape} -- the deviations of individual dimensions from its mean capability -- remains recoverable (\S\ref{sec:alpha_choice}, Table~\ref{tab:alpha_analysis}). We instead estimate a single intercept shared across the catalogue, which anchors agents relative to one another. In synthetic experiments, this improves recovery of overall capability level from $r = 0.12$ to $r = 0.98$.

This shared anchor also limits how overall capability levels should be interpreted. They are identified \emph{relative} to the other agents in the catalogue, not on an absolute scale. Adding or removing agents may shift the common reference point, so absolute levels should not be compared across separately fitted catalogues. Within a catalogue, however, comparisons of overall level remain meaningful, as do comparisons of profile shape.

These identifiability constraints suggest a natural extension: fitting many agents jointly in a hierarchical model. Individual capability profiles could be drawn from a shared population distribution, while structural parameters such as $\lambda$ (or dimension-specific $\lambda_k$), discrimination weights, and the intercept could be estimated at the population level. Variation across agents would then provide information for separating these structural parameters from individual capabilities. Such a model could estimate profiles for a population of agents, or human participants (\S\ref{sec:discussion}), in a single fit and place them on a common scale by construction. We leave this extension to future work. The present model, with individually estimated capability profiles and a shared intercept, provides a simpler identifiable alternative.

\clearpage

\section{Questionnaire and interviews}\label{app:questionnaire}
\renewcommand{\thetable}{B\arabic{table}}
\renewcommand{\thefigure}{B\arabic{figure}}
\setcounter{table}{0}
\setcounter{figure}{0}

\begin{figure}[ht]
\centering
\includegraphics[width=1\linewidth]{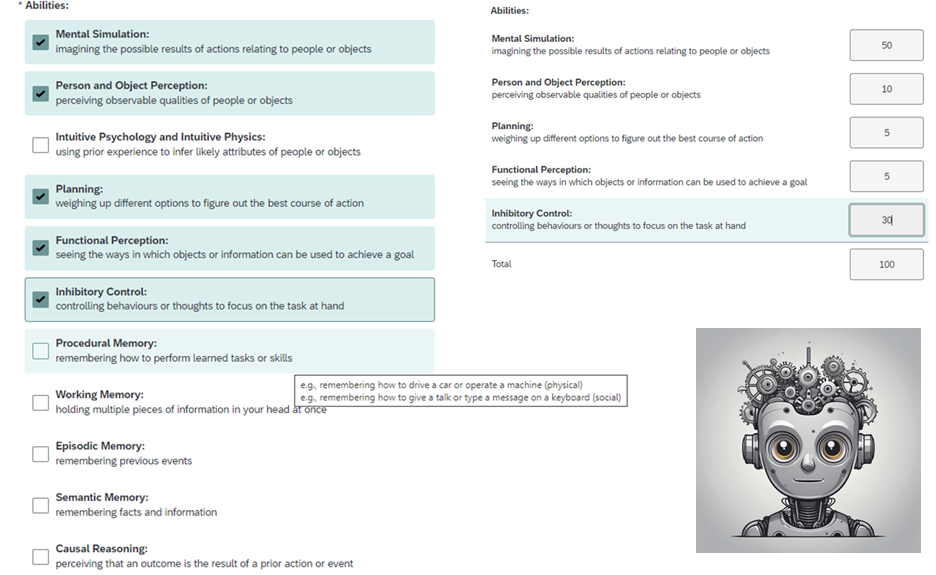}
\vspace{6pt}
\caption{The task-capability weighting process. \textit{Left:} Participants select the five capabilities they consider most essential for performing the target work activity, framed as equipping a robot helper for that role; hovering over any activity reveals a tooltip with concrete examples. \textit{Right:} Participants distribute 100 ability points across their five selected capabilities, tuning the robot to the configuration they judge optimal -- yielding a weighted profile of the most valuable capabilities for that activity. \textit{Note}: capabilities and examples shown here are illustrative only -- the actual questionnaire used the capabilities in Table \ref{tab:cognitive_capabilities}}
\label{fig:questionnaire_ranking}
\end{figure}

\begin{figure}[ht]
\centering
\includegraphics[width=1\linewidth]{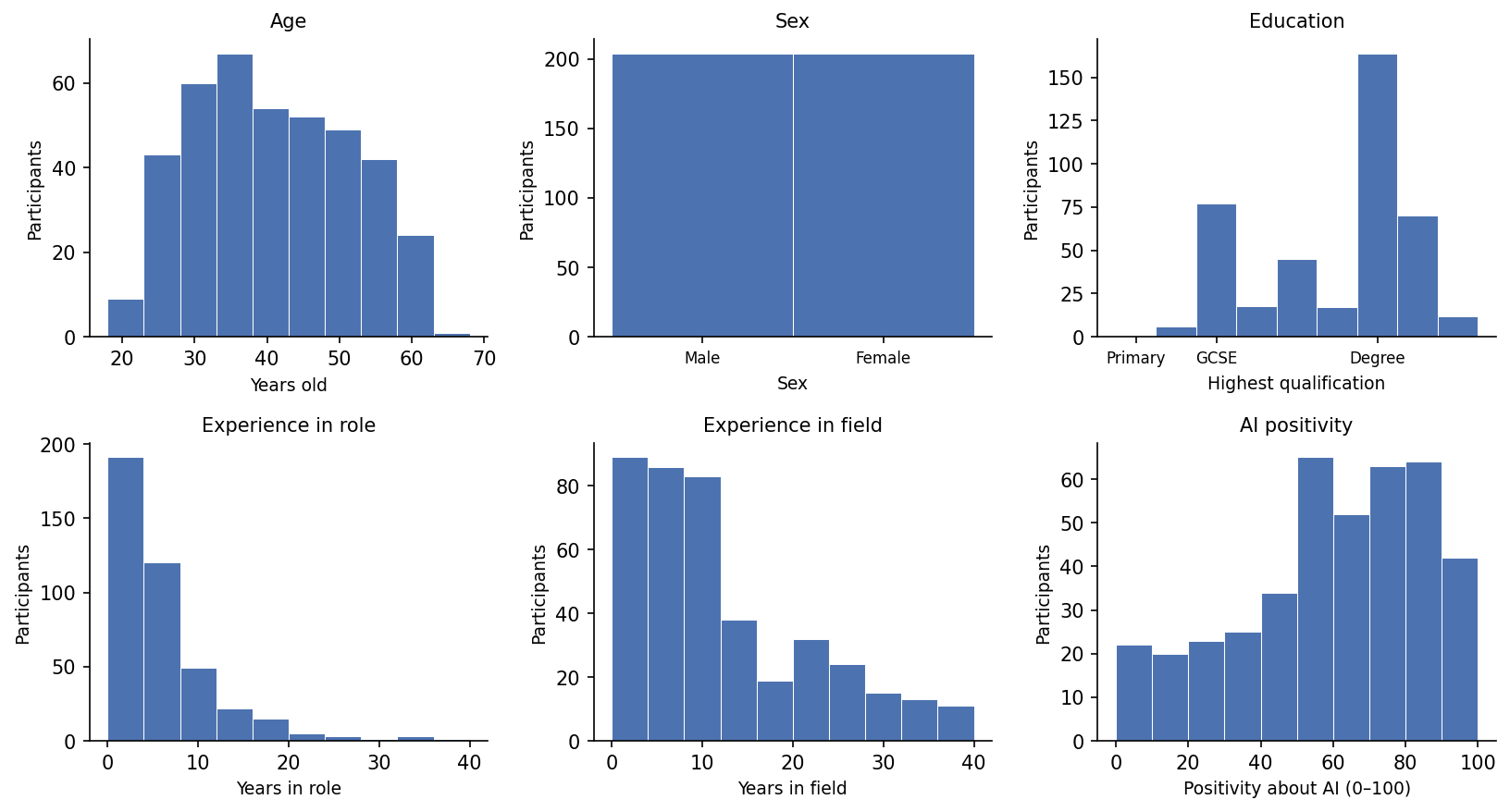}
\vspace{6pt}
\caption{Participant demographic histograms for the full sample after exclusions ($N = 410$).}
\label{fig:questionnaire_demographics}
\end{figure}

This section provides supplementary materials relating to the questionnaire and interviews. Table \ref{tab:work_activities} describes the 18 work activities adapted from O*NET, while Figure \ref{fig:questionnaire_ranking} illustrates how participants selected and distributed importance weights across capabilities for individual tasks. Table \ref{tab:participants} summarises questionnaire respondents by recruitment source (company or online), and Tables \ref{tab:validation_summary}--\ref{tab:validation_combined} present the sample validation analysis, comparing capability importance ratings between the company and online samples. Finally, Table \ref{tab:demographics} and Figure \ref{fig:questionnaire_demographics} summarise the demographic characteristics of the full sample.

\begin{table}[ht]
\centering
\caption{Work activities}
\vspace{6pt}
\label{tab:work_activities}
\renewcommand{\arraystretch}{1.3}
\begin{tabular}{p{0.2\linewidth} p{0.75\linewidth}}
\toprule
\textbf{Work activity} & \textbf{Description} \\
\midrule
Researching & Learning about latest developments, researching a product or client \\
Checking & Inspecting products or documents, interviewing staff or reviewing their performance, ensuring compliance with regulations or guidelines \\
Decision making & Deciding project goals, choosing between different plans or options \\
Problem solving & Finding ways to overcome roadblocks, fix issues, or resolve conflict \\
Creative thinking & Coming up with new ideas for products or marketing, innovating \\
Long-term planning & Considering the overall goals for a project, setting objectives, strategising \\
Short-term planning & Scheduling your time, deciding which tasks are most important and the order they should be completed \\
Tool use & Using physical tools or machines, using your hands to move, alter or affect objects \\
Computer use & Interacting with computer software, email or apps to communicate or complete a task \\
Writing code & Creating programmes or interacting with a computer using a programming language, instead of a user interface \\
Data manipulation & Structuring, processing or cleaning data for storage or later use \\
Analysing data& Running statistical tests, consulting graphs, drawing insights \\
Communicating & Speaking to colleagues, writing articles, delivering a sales pitch \\
Listening & Taking instructions, reading documents, listening to feedback \\
Building rapport & Developing rapport with clients or colleagues, nurturing existing relationships \\
Managing people & Training new or junior staff, delivering advice, supervising, mentoring \\
Managing resources & Monitoring expenses, working within a budget, ensuring adequate supplies \\
Admin & Keeping records on what was said or done or how resources were used, cataloguing \\
\bottomrule
\addlinespace[2pt]
\multicolumn{2}{@{}p{\linewidth}@{}}{\footnotesize\textit{Note.} The 18 work activities were adapted from O*NET work activity categories \cite{onet2026}.}\\
\end{tabular}
\end{table}

\begin{table}[ht]
\centering
\caption{Participant questionnaire sample.}
\begin{tabular}{lrrrrrr}
\toprule
 & \multicolumn{2}{c}{Companies} & \multicolumn{2}{c}{Online} & \multicolumn{2}{c}{Total} \\
\cmidrule(lr){2-3} \cmidrule(lr){4-5} \cmidrule(lr){6-7}
Domain & Pre-QC & Post-QC & Pre-QC & Post-QC & Pre-QC & Post-QC \\
\midrule
Warehouse or logistics (WL) & 2 & 1 & 47 & 36 & 49 & 37 \\
Manufacture, maintenance, or repair (MMR) & 0 & 0 & 48 & 33 & 48 & 33 \\
Numerical, data, or programming (NDP) & 18 & 12 & 59 & 50 & 77 & 62 \\
Administration, organisational, or planning (AOP) & 46 & 37 & 125 & 102 & 171 & 139 \\
Customer service, marketing or HR (CMH) & 38 & 24 & 67 & 55 & 105 & 79 \\
Hospitality, sales, or client care (HSC) & 21 & 11 & 68 & 49 & 89 & 60 \\
\midrule
Total & 125 & 85 & 414 & 325 & 539 & 410 \\
\bottomrule
\end{tabular}

\vspace{0.5em}
{\raggedright\footnotesize
\textit{Note.} Participant numbers by job domain before and after quality control (QC), split by recruitment source (company vs online) with the combined total.\par}
\label{tab:participants}
\end{table}

\begin{table}[htbp]
  \centering
  \caption{Sample demographics, overall and by source.}
  \label{tab:demographics}
  \begin{tabular}{lrrrrrrr}
    \toprule
    Group & $n$ & Age (mean $\pm$ sd) & \% female & Median education & Yrs in role & Yrs in field & AI positivity \\
    \midrule
    Overall & 410 & $40.3 \pm 10.7$ & 49.8 & Degree & 5.7 & 11.8 & 57.6 \\
    Companies & 85 & $37.7 \pm 12.0$ & 65.9 & Degree & 4.2 & 8.4 & 62.1 \\
    Online & 325 & $40.9 \pm 10.3$ & 45.5 & Degree & 6.1 & 12.7 & 56.4 \\
    \bottomrule
  \end{tabular}
\end{table}

\begin{table}[h]
  \centering
  \caption{Companies vs online agreement: sample validation.}
  \label{tab:validation_summary}
  \begin{tabular}{lrcrrrr}
    \toprule
    & Cosine & Pearson & Reliability & Reliability & Noise & Disattenuated \\
    & & [95\% CI] & (companies) & (online) & ceiling & $r$ \\
    \midrule
    Per capability & 0.846 & 0.620 [0.508, 0.732] & 0.497 & 0.733 & 0.603 & 1.000 \\
    Per work activity & 0.905 & 0.876 [0.830, 0.923] & 0.788 & 0.935 & 0.858 & 1.000 \\
    \bottomrule
  \end{tabular}

  \vspace{0.5em}
{\raggedright \footnotesize
  \textit{Note.} Values are aggregated across the 16 capabilities and 18 work activities retained after the inter-rater reliability check (Table~\ref{tab:interrater}). Cosine and Pearson are the mean cosine similarity and mean (Fisher $z$-averaged) Pearson correlation between the company and online importance vectors; bracketed values are bootstrap 95\% CIs for the mean Pearson. Reliability is the Spearman--Brown split-half correlation of each source's own respondents; the noise ceiling $\sqrt{\mathrm{rel}_\mathrm{comp}\cdot\mathrm{rel}_\mathrm{online}}$ is the largest between-source Pearson attainable given that noise. Disattenuated $r$ is mean Pearson $\div$ ceiling, capped at 1 (uncapped 1.03 and 1.02).\par}
\end{table}

\begin{table}[htbp]
  \centering
  \caption{Companies vs online agreement per capability and per work activity.}
  \label{tab:validation_combined}
  \begin{tabular}{lrrr @{\hspace{2.5em}} lrrr}
    \toprule
    \multicolumn{3}{c}{Capabilities} & \multicolumn{3}{c}{Work activities} \\
    \cmidrule(lr){1-3} \cmidrule(lr){4-6}
    Capability & Cosine & Pearson & Work activity & Cosine & Pearson \\
    \midrule
    EM & 0.893 & 0.599 & Researching & 0.982 & 0.964 \\
    SM & 0.875 & 0.483 & Checking & 0.889 & 0.779 \\
    ProcM & 0.874 & 0.689 & Decision making & 0.953 & 0.805 \\
    WM & 0.916 & 0.548 & Problem solving & 0.950 & 0.816 \\
    CF & 0.894 & 0.383 & Creative thinking & 0.910 & 0.694 \\
    MS & 0.961 & 0.879 & LT planning & 0.960 & 0.933 \\
    P & 0.948 & 0.825 & ST planning & 0.959 & 0.905 \\
    MC & 0.838 & 0.660 & Tool use & 0.728 & 0.336 \\
    PaPR & 0.896 & 0.239 & Computer use & 0.994 & 0.987 \\
    FP & 0.676 & -0.190 & Coding & 0.496 & 0.227 \\
    SRaN & 0.415 & 0.002 & Data manipulation & 0.957 & 0.923 \\
    OP & 0.786 & 0.483 & Analysing data & 0.970 & 0.926 \\
    CR & 0.712 & -0.069 & Communicating & 0.966 & 0.927 \\
    ToM & 0.904 & 0.754 & Listening & 0.977 & 0.954 \\
    EPaE & 0.973 & 0.958 & Building rapport & 0.985 & 0.972 \\
    L & 0.979 & 0.943 & Managing people & 0.942 & 0.856 \\
    && & Managing resources & 0.729 & 0.431 \\
    && & Admin & 0.950 & 0.893 \\
    \midrule
    Mean & 0.846 & 0.512 & Mean & 0.905 & 0.796 \\
    \bottomrule
  \end{tabular}

    \vspace{0.5em}
  {\raggedright \footnotesize
  \textit{Note.} Agreement is assessed using cosine similarity and Pearson correlation of the importance vectors. We validate on the 16 capabilities retained after the inter-rater reliability check (Table \ref{tab:interrater}). Full capability names are provided in Table \ref{tab:cognitive_capabilities} and work activity descriptions in Table \ref{tab:work_activities}.\par}
\end{table}

\clearpage

\section{Supplementary results}\label{app:results}
\renewcommand{\thetable}{C\arabic{table}}
\renewcommand{\thefigure}{C\arabic{figure}}
\setcounter{table}{0}
\setcounter{figure}{0}

In this section we provide supplementary results relating to the three aspects of the pipeline: capability estimation (\S\ref{app:capability_profiles}), task requirements from the questionnaire analysis (\S\ref{app:results_questionnaire}), and the for the resulting suitability scores (\S\ref{app:results_suitability}).

\subsection{Recovery analysis}\label{app:recovery}

\begin{figure}[ht]
\centering
\includegraphics[width=1\linewidth]{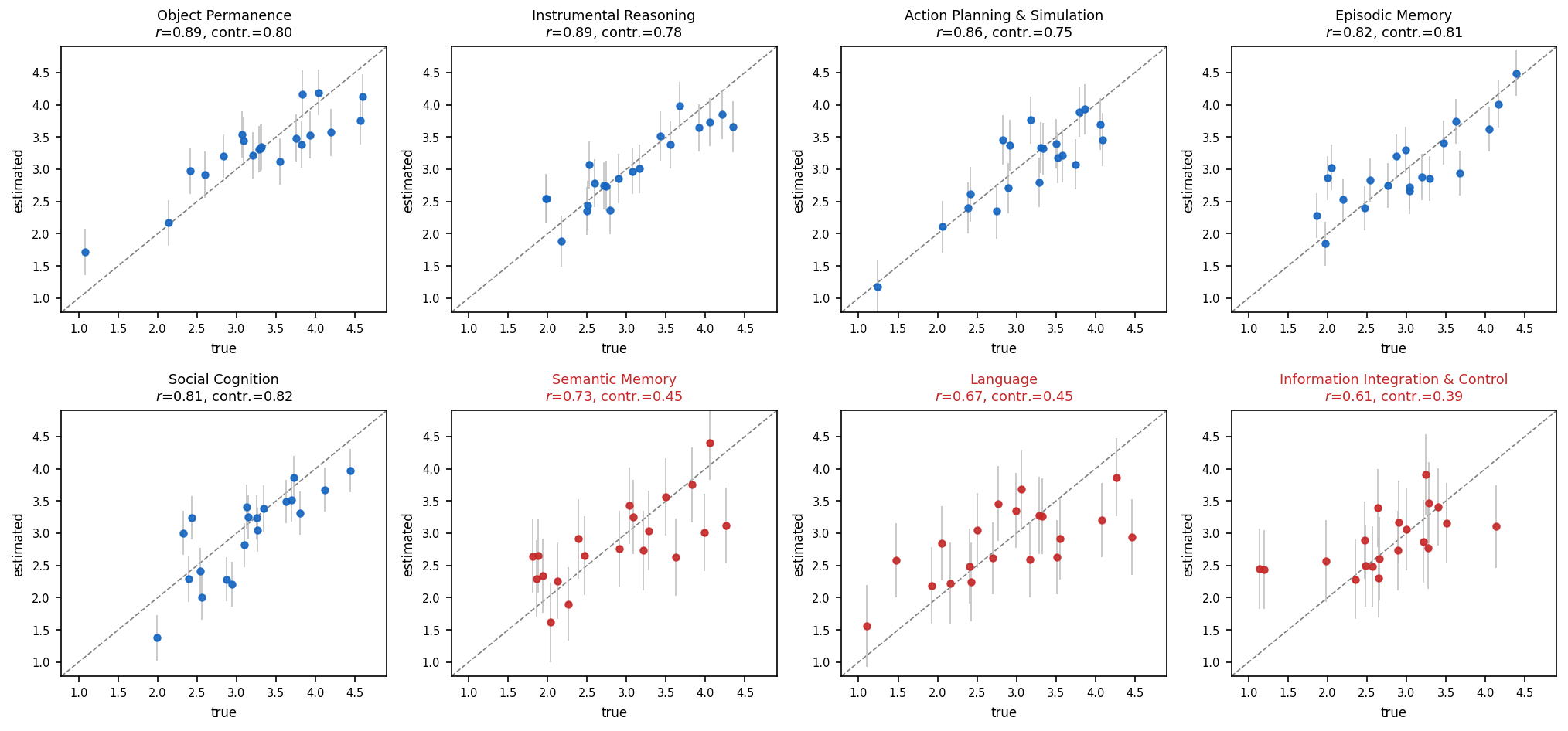}
\vspace{6pt}
\caption{Fully compensatory capability recovery across dimensions. Each panel plots posterior-mean estimates $\hat{c}_k$ ($\pm 1$\,SD) against ground-truth capability levels $c_k$ across synthetic agents ($N = 20$), under a fully compensatory model ($\tau = 0$), with the recovery correlation $r$ reported per dimension. Contraction (contr.), a measure of how identifiable each dimension is, is also reported; dimensions with contraction $< 0.5$ are highlighted in red.}
\label{fig:recovery_scatter}
\end{figure}

This appendix section reports the full recovery analysis underlying the results of \S\ref{sec:recovery_results}. Figure~\ref{fig:recovery_scatter} and Tables~\ref{tab:tau_sweep_contraction}--\ref{tab:tau_sweep_mae} show recovery across the soft-min pooling temperature $\tau$. For each value of $\tau$, a fixed population of synthetic agents with known capability profiles (Table~\ref{tab:synthetic_agent_profiles}) is simulated and re-fit using the same inference procedure. Recovery is assessed primarily using the between-agent correlation of true and posterior-mean capability values for each dimension (Figure~\ref{fig:recovery_scatter}, Table~\ref{tab:tau_sweep}). Posterior contraction (Table~\ref{tab:tau_sweep_contraction}) and log-scale mean absolute error (Table~\ref{tab:tau_sweep_mae}) provide complementary measures of informativeness and absolute error. Together, these diagnostics support the choice of $\tau=1$: recovery of the highest-coverage dimensions improves substantially as pooling moves away from the fully compensatory baseline toward weakest-link behaviour, with the principal trade-off being a modest decline in Instrumental Reasoning recovery.

The recovery analysis also reveals an identifiability confound between an agent's overall capability level and the intercept $\alpha$. Because the pooling rule depends only on the margins $c_k-\lambda D_{jk}$, adding a constant to every capability and subtracting the same constant from $\alpha$ leaves all predicted outcomes unchanged. A free per-agent intercept therefore leaves the overall capability level unidentified while preserving the relative profile shape. Table~\ref{tab:posterior_collinearity_per_agent} examines the resulting posterior geometry. With independent intercepts, the eight capability posteriors collapse onto approximately $2.8$ effective dimensions and exhibit strong cross-capability correlations. Estimating a single intercept jointly across agents approximately doubles effective dimensionality, reduces posterior correlations, and substantially improves conditioning. In synthetic recovery, this shared anchor increases level recovery from $r=0.12$ to $0.98$ and improves downstream suitability recovery from near chance to $\rho=0.91$. We therefore use a shared intercept in all subsequent analyses.

\vspace{20pt}

\begin{table}[h]
  \centering
  \caption{Synthetic agent capability profiles used in the recovery analysis}
  \label{tab:synthetic_agent_profiles}
  \begin{tabular}{lrrrrrrrr}
    \toprule
    Agent & SC & IR & L & SM & IIC & APaS & EM & OP \\
    \midrule
    A1 & 3.10 & 2.89 & 3.51 & 3.08 & 2.57 & 3.29 & 4.04 & 3.76 \\
    A2 & 2.44 & 1.99 & 2.50 & 3.03 & 1.14 & 2.82 & 2.00 & 2.41 \\
    A3 & 2.56 & 2.75 & 3.33 & 3.83 & 2.90 & 4.09 & 2.47 & 3.28 \\
    A4 & 3.72 & 3.08 & 2.41 & 2.26 & 2.63 & 3.18 & 2.19 & 2.83 \\
    A5 & 2.87 & 3.43 & 3.17 & 3.28 & 2.48 & 2.90 & 3.63 & 4.19 \\
    A6 & 1.99 & 4.21 & 4.08 & 3.63 & 3.21 & 2.75 & 4.17 & 4.57 \\
    A7 & 4.44 & 4.05 & 3.29 & 2.03 & 3.00 & 3.53 & 1.97 & 3.32 \\
    A8 & 3.34 & 3.56 & 2.05 & 2.47 & 2.65 & 2.06 & 4.39 & 2.60 \\
    A9 & 3.26 & 2.79 & 4.27 & 4.06 & 3.51 & 1.24 & 3.04 & 3.55 \\
    A10 & 3.80 & 2.51 & 4.46 & 1.94 & 2.47 & 3.75 & 3.04 & 4.60 \\
    A11 & 3.15 & 2.49 & 2.70 & 2.13 & 1.98 & 3.50 & 3.46 & 4.04 \\
    A12 & 2.40 & 4.35 & 2.77 & 4.26 & 2.65 & 2.41 & 3.20 & 3.83 \\
    A13 & 3.13 & 2.53 & 1.93 & 1.88 & 3.40 & 3.79 & 2.87 & 2.14 \\
    A14 & 3.70 & 1.98 & 2.43 & 3.50 & 1.20 & 3.31 & 2.53 & 3.09 \\
    A15 & 2.94 & 3.16 & 3.56 & 2.39 & 4.14 & 3.58 & 3.67 & 3.93 \\
    A16 & 3.63 & 3.68 & 3.06 & 1.86 & 2.89 & 2.38 & 1.86 & 3.21 \\
    A17 & 2.55 & 2.18 & 2.17 & 3.21 & 3.29 & 4.06 & 2.99 & 3.83 \\
    A18 & 4.12 & 3.92 & 1.11 & 3.98 & 3.27 & 3.34 & 3.30 & 3.31 \\
    A19 & 3.26 & 2.71 & 1.48 & 2.91 & 2.36 & 3.86 & 2.77 & 3.07 \\
    A20 & 2.32 & 2.59 & 2.99 & 1.81 & 3.24 & 2.92 & 2.05 & 1.08 \\
    \midrule
    Mean & 3.14 & 3.04 & 2.86 & 2.88 & 2.75 & 3.14 & 2.98 & 3.33 \\
    SD & 0.64 & 0.73 & 0.89 & 0.82 & 0.73 & 0.72 & 0.76 & 0.85 \\
    \bottomrule
  \end{tabular}

  \vspace{0.5em}
  {\raggedright \footnotesize
  \textit{Note.} A fixed population of $N=20$ synthetic agents is drawn once from the log-scale capability prior $c_{ak}\sim\mathcal{N}(\mu_c,\sigma_c^2)$ with $\mu_c=3.0$ and $\sigma_c=0.8$, then simulated and re-fit self-consistently at each pooling temperature $\tau$ (Table~\ref{tab:tau_sweep}). Values are the log-scale capabilities $c_{ak}$ used to simulate performance; the corresponding ratio-scale capabilities are $\theta_k=e^{c_k}$. Columns are the $K=8$ clustered battery dimensions (Table~\ref{tab:clustered_capabilities}): SC (Social Cognition), IR (Instrumental Reasoning), L (Language), SM (Semantic Memory), IIC (Information Integration \& Control), APaS (Action Planning \& Simulation), EM (Episodic Memory), OP (Object Permanence). \par}
\end{table}

\begin{table}[h]
  \centering
  \caption{Posterior contraction per battery dimension across the soft-min temperature $\tau$.}
  \label{tab:tau_sweep_contraction}
  \begin{tabular}{lrrrrr}
    \toprule
    Dimension (coverage) & $\tau{=}0$ & $0.25$ & $0.5$ & $1$ & $2$ \\
    \midrule
    Information Integration \& Control (100\%) & 0.39 & 0.55 & 0.68 & 0.77 & \textbf{0.80} \\
    Language (100\%) & 0.45 & 0.57 & 0.69 & \textbf{0.75} & 0.72 \\
    Semantic Memory (99\%) & 0.45 & 0.71 & 0.81 & \textbf{0.84} & \textbf{0.84} \\
    Action Planning \& Sim.\ (94\%) & \textbf{0.75} & 0.74 & 0.73 & 0.72 & 0.71 \\
    Object Permanence (36\%) & \textbf{0.80} & 0.77 & 0.74 & 0.71 & 0.68 \\
    Social Cognition (42\%) & \textbf{0.82} & 0.81 & 0.80 & 0.78 & 0.76 \\
    Episodic Memory (50\%) & \textbf{0.81} & 0.79 & 0.77 & 0.73 & 0.66 \\
    Instrumental Reasoning (92\%) & \textbf{0.78} & 0.76 & 0.76 & 0.74 & 0.66 \\
    \midrule
    Mean & 0.66 & 0.71 & 0.75 & \textbf{0.76} & 0.73 \\
    Worst dimension & 0.39 & 0.55 & 0.68 & \textbf{0.71} & 0.66 \\
    \bottomrule
  \end{tabular}

  \vspace{0.5em}
  {\raggedright \footnotesize
  \textit{Note.} Contraction $=1-\mathrm{Var}_{\text{post}}/\mathrm{Var}_{\text{prior}}$ measures how much the data sharpen the prior (1 = fully informed by data, 0 = prior only). As in Table \ref{tab:tau_sweep}, a fixed population of synthetic agents is simulated and re-fit self-consistently at each $\tau$. The simulation intercept is recalibrated so mean accuracy remains $0.60$ (item difficulty held constant), isolating pooling effects from changes in overall difficulty. $\tau{=}0$ corresponds to the normalized-additive (compensatory) model; larger $\tau$ values move toward weakest-link pooling. Bold indicates the highest contraction in each row. \par}
\end{table}

\begin{table}[t]
  \centering
  \caption{Log-scale recovery error per battery dimension across the soft-min temperature $\tau$.}
  \label{tab:tau_sweep_mae}
  \begin{tabular}{lrrrrr}
    \toprule
    Dimension (coverage) & $\tau{=}0$ & 0.25 & 0.5 & 1 & 2 \\
    \midrule
    Information Integration \& Control (100\%) & 0.42 & 0.42 & 0.40 & 0.33 & \textbf{0.26} \\
    Language (100\%) & 0.51 & 0.43 & 0.38 & \textbf{0.30} & 0.34 \\
    Semantic Memory (99\%) & 0.45 & 0.39 & 0.32 & \textbf{0.29} & 0.30 \\
    Action Planning \& Sim.\ (94\%) & 0.29 & 0.29 & 0.32 & 0.33 & \textbf{0.25} \\
    Instrumental Reasoning (92\%) & \textbf{0.27} & 0.31 & 0.35 & 0.39 & 0.39 \\
    Episodic Memory (50\%) & 0.34 & \textbf{0.33} & 0.39 & 0.34 & 0.34 \\
    Social Cognition (42\%) & 0.35 & 0.35 & 0.35 & \textbf{0.33} & 0.37 \\
    Object Permanence (36\%) & \textbf{0.34} & 0.38 & 0.40 & 0.41 & 0.34 \\
    \midrule
    Mean & 0.37 & 0.36 & 0.36 & 0.34 & \textbf{0.32} \\
    Worst dimension & 0.51 & 0.43 & 0.40 & 0.41 & \textbf{0.39} \\
    \bottomrule
  \end{tabular}

  \vspace{0.5em}
{\raggedright \footnotesize
  \textit{Note.} MAE is the mean over agents of $|c_k-\hat c_k|$ on the log (capability) scale. As in Table \ref{tab:tau_sweep}, a fixed population of synthetic agents is simulated and re-fit self-consistently at each $\tau$. The simulation intercept is recalibrated so mean accuracy remains $0.60$ (item difficulty held constant), isolating pooling effects from changes in overall difficulty. $\tau{=}0$ corresponds to the normalized-additive (compensatory) model; larger $\tau$ values move toward weakest-link pooling. Bold indicates the lowest MAE in each row.\par}
\end{table}

\begin{table}[t]
  \centering
  \caption{Per-agent within-agent posterior collinearity of the eight log-capabilities
    $c_k$, with free and shared intercepts.}
  \label{tab:posterior_collinearity_per_agent}
  \small
  \setlength{\tabcolsep}{5pt}
  \begin{tabular}{l rrrrr rrrrr}
    \toprule
    & \multicolumn{5}{c}{Free intercept} & \multicolumn{5}{c}{Shared intercept} \\
    \cmidrule(lr){2-6}\cmidrule(lr){7-11}
    Agent & Eff.\ dims & $\overline{|r|}$ & $\max|r|$ & $\mathrm{cond}(R)$ & worst contr.
          & Eff.\ dims & $\overline{|r|}$ & $\max|r|$ & $\mathrm{cond}(R)$ & worst contr. \\
    \midrule
    GPT-4o-mini            & 3.07 & 0.42 & 0.95 & 98  & 0.59 & 5.68 & 0.16 & 0.75 & 12 & 0.64 \\
    o4-mini                & 2.58 & 0.50 & 0.94 & 180 & 0.62 & 5.02 & 0.22 & 0.72 & 23 & 0.66 \\
    Gemini 3 Flash         & 2.66 & 0.48 & 0.98 & 229 & 0.60 & 4.84 & 0.22 & 0.86 & 25 & 0.70 \\
    Gemini 3.1 Pro & 3.06 & 0.42 & 0.98 & 224 & 0.56 & 5.15 & 0.20 & 0.88 & 25 & 0.72 \\
    GPT-5-nano             & 2.32 & 0.53 & 0.96 & 214 & 0.57 & 4.54 & 0.24 & 0.76 & 24 & 0.59 \\
    Gemini 2.5 Flash       & 3.02 & 0.44 & 0.82 & 91  & 0.58 & 6.00 & 0.18 & 0.41 & 12 & 0.58 \\
    \midrule
    Mean                   & 2.79 & 0.47 & 0.94 & 173 & 0.58 & 5.21 & 0.20 & 0.73 & 20 & 0.65 \\
    \bottomrule
  \end{tabular}

  \vspace{0.5em}
    {\raggedright\footnotesize
    \textit{Note.} For each agent, posterior draws of the eight $c_k$ define a correlation matrix $R$. Reported diagnostics are the effective dimensionality (participation ratio, $(\sum_i\lambda_i)^2/\sum_i\lambda_i^2$, maximum $8$), the mean and maximum absolute off-diagonal correlations ($\overline{|r|}$, $\max|r|$; lower is better), the condition number $\mathrm{cond}(R)$ (multicollinearity; lower is better), and the worst contraction,
    $\min_k[1-\mathrm{Var}_{\mathrm{post}}(c_k)/\sigma_c^2]$, with $\sigma_c=0.8$. Both models use a soft-min choice rule ($\tau=1$), the same behavioural battery, and the same prior.\par}
\end{table}

\clearpage

\subsection{AI capability profiles}\label{app:capability_profiles}

\begin{figure*}[ht]
\centering
\includegraphics[width=1\linewidth]{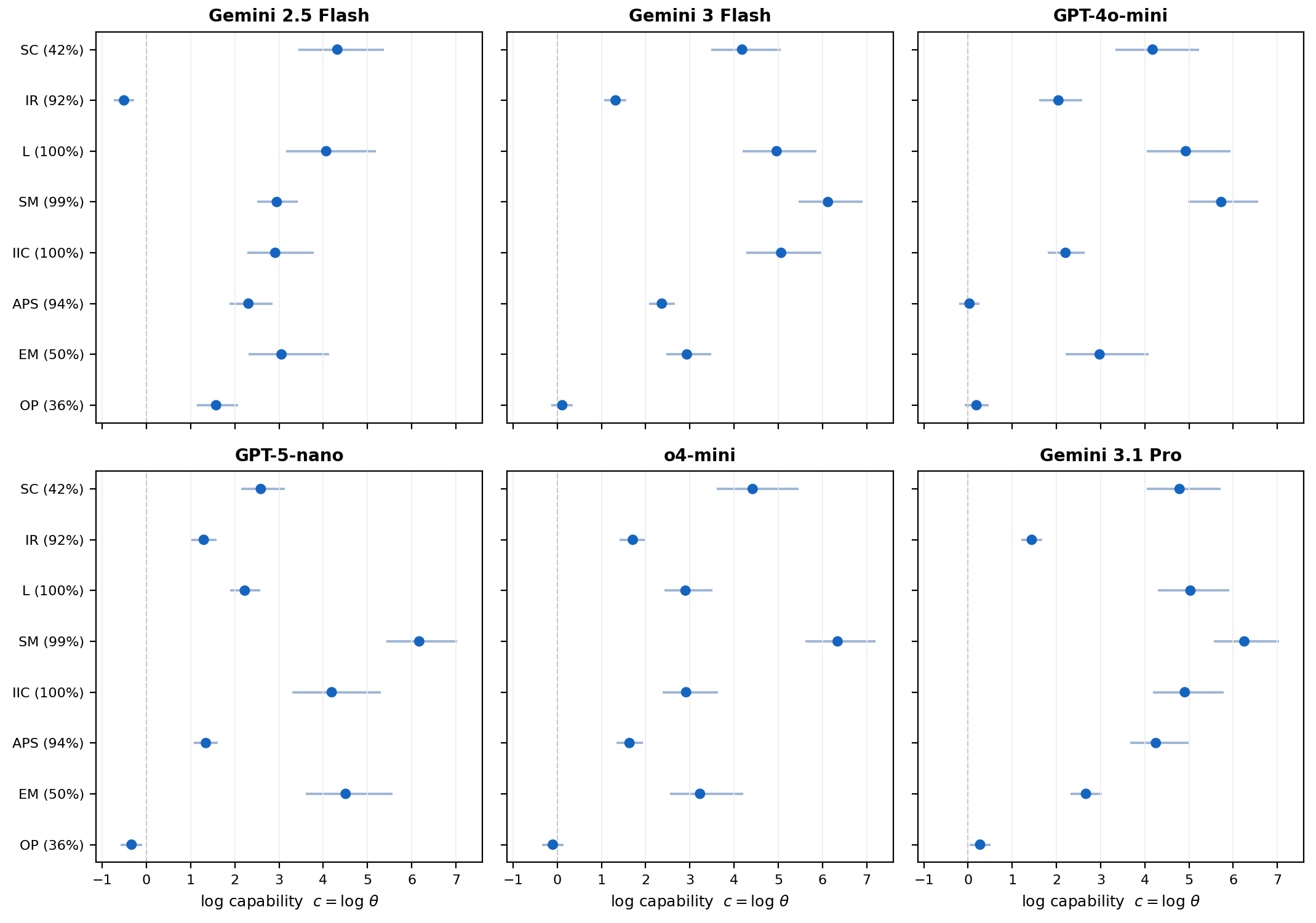}
\vspace{6pt}
\caption{Per-capability posteriors by system. Each panel is a forest plot of the eight clustered battery dimensions for a single AI system. Dots are posterior means of the log-capability $c_k = \log\theta_k$ and bars are 95\% highest-density intervals, using soft-min pooling ($\tau=1$) and a shared intercept across systems. The dashed vertical line marks $c=0$ ($\theta=1$). The axis labels for dimensions display coverage (the share of battery items demanding it) in parentheses. Axes: SC = Social Cognition, IR = Instrumental Reasoning, L = Language, SM = Semantic Memory, IIC = Information Integration \& Control, APS = Action Planning \& Simulation, EM = Episodic Memory, OP = Object Permanence.}
\label{fig:forest_grid}
\end{figure*}

This section provides additional detail on the capability profiling results reported in \S\ref{sec:capability_results}. Figure~\ref{fig:forest_grid} displays the posterior capability estimates for each AI system across the eight clustered dimensions. Table~\ref{tab:by_system} aggregates across dimensions to provide each system's overall estimated capability level, while Table~\ref{tab:capability_breakdown} reports the complete system-by-capability breakdown.

The inferred profiles are broadly consistent in shape but differ in overall level. The two Gemini~3 models achieve the highest aggregate capability estimates among the evaluated systems. The relative ordering of dimensions is also largely preserved across systems, with Semantic Memory, Language, and Social Cognition among the strongest estimated capabilities, and Action Planning \& Simulation, Instrumental Reasoning, and Object Permanence among the weakest.

\begin{table}[t]
  \centering
  \caption{Per-system summary, aggregated across the eight capability dimensions.}
  \label{tab:by_system}
  \begin{tabular}{lrrrrrr}
    \toprule
    System & Acc. & $N$ & $\bar c$ & $SD$ & $\hat R_{\max}$ & $\mathrm{ESS}_{\min}$ \\
    \midrule
    Gemini 3.1 Pro & 73.3\% & 19,535 & 3.70 & 0.30 & 1.013 & 363 \\
    Gemini 3 Flash & 68.4\% & 19,535 & 3.38 & 0.29 & 1.013 & 376 \\
    o4-mini & 61.8\% & 19,535 & 2.88 & 0.29 & 1.011 & 377 \\
    GPT-4o-mini & 48.7\% & 19,535 & 2.79 & 0.32 & 1.013 & 362 \\
    GPT-5-nano & 56.0\% & 19,535 & 2.75 & 0.28 & 1.013 & 382 \\
    Gemini 2.5 Flash & 44.8\% & 19,535 & 2.58 & 0.34 & 1.013 & 348 \\
    \bottomrule
  \end{tabular}

    \vspace{0.5em}
    {\raggedright\footnotesize
    \textit{Note.} All systems are fit jointly with a shared intercept, which anchors each system's overall level against the others. \emph{Accuracy} is the overall fraction of benchmark items answered correctly. \emph{Capability} is the mean posterior log-capability $\bar c=\overline{\log\theta}$; $SD$ is the mean posterior standard deviation. $\hat R_{\max}$ and $\mathrm{ESS}_{\min}$ are the worst-case MCMC convergence diagnostics across dimensions: at 2{,}000 draws/chain (4 chains) all fits are well converged, with $\hat R_{\max}\!\le\!1.013$ and bulk $\mathrm{ESS}_{\min}\!\gtrsim\!350$.\par}
\end{table}

\begin{table}[t]
  \centering
  \footnotesize
  \caption{Posterior capability estimates for every system $\times$ capability.}
  \label{tab:capability_breakdown}
  \begin{tabular}{lrrrrrrrr}
    \toprule
    System & IIC & L & SM & APS & IR & EM & SC & OP \\
    \midrule
    Gemini 3.1 Pro & 4.90 (0.41) & 5.03 (0.41) & 6.25 (0.38) & 4.25 (0.34) & 1.44 (0.12) & 2.67 (0.18) & 4.79 (0.42) & 0.27 (0.13) \\
    Gemini 3 Flash & 5.06 (0.44) & 4.96 (0.43) & 6.12 (0.37) & 2.37 (0.15) & 1.32 (0.13) & 2.93 (0.26) & 4.18 (0.40) & 0.11 (0.13) \\
    o4-mini & 2.92 (0.32) & 2.90 (0.27) & 6.34 (0.40) & 1.63 (0.15) & 1.71 (0.15) & 3.23 (0.42) & 4.42 (0.47) & -0.10 (0.13) \\
    GPT-4o-mini & 2.21 (0.21) & 4.93 (0.48) & 5.73 (0.41) & 0.03 (0.12) & 2.05 (0.25) & 2.98 (0.48) & 4.18 (0.48) & 0.19 (0.14) \\
    GPT-5-nano & 4.19 (0.51) & 2.22 (0.18) & 6.17 (0.41) & 1.34 (0.14) & 1.30 (0.15) & 4.51 (0.50) & 2.59 (0.25) & -0.34 (0.13) \\
    Gemini 2.5 Flash & 2.91 (0.39) & 4.07 (0.52) & 2.95 (0.23) & 2.31 (0.25) & -0.51 (0.12) & 3.06 (0.47) & 4.32 (0.50) & 1.58 (0.24) \\
    \midrule
    \textit{Mean} & 3.70 & 4.02 & 5.59 & 1.99 & 1.22 & 3.23 & 4.08 & 0.29 \\
    \bottomrule
  \end{tabular}

  \vspace{0.5em}
    {\raggedright\footnotesize
    \textit{Note.} Capability estimates are $c=\log\theta$ (posterior SD in parentheses). Columns are ordered by coverage (left = best covered). All values are separately identified (mean max posterior $|$corr$|\le0.90$; see Table \ref{tab:posterior_collinearity_per_agent}).\par}
\end{table}

\clearpage
\subsection{Questionnaire analysis}\label{app:results_questionnaire}

\begin{figure}[ht]
\centering
\includegraphics[width=1\linewidth]{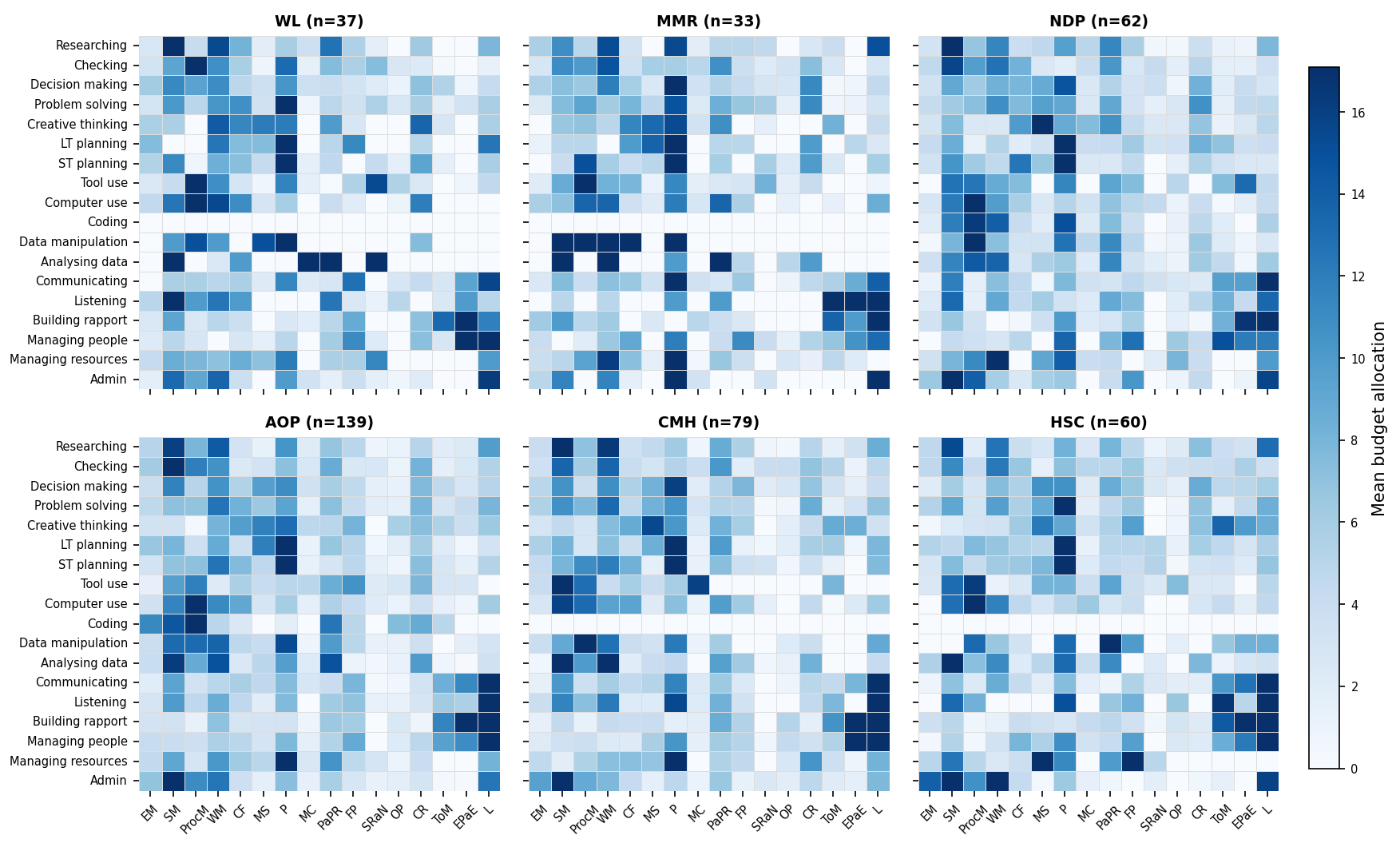}
\vspace{6pt}
\caption{Capability importance per work activity, by occupational domain. Panels give the capability-importance matrix (scored as in Figure~\ref{fig:ability_matrix_all}) estimated separately within each of the six occupational domains: Warehouse and Logistics (WL), Manufacture, Maintenance, or Repair (MMR), Numerical, Data, or Programming (NDP); Administration, Organisational, or Planning (AOP); Customer Service, Marketing, or HR (CMH); Hospitality, Sales, or Client Care (HSC).}
\label{fig:ability_matrix_by_domain}
\end{figure}

Section~\ref{sec:questionnaire_results} describes how the questionnaire responses are aggregated into task-ability importance matrices. Complementing the pooled matrix of Figure~\ref{fig:ability_matrix_all}, Figure~\ref{fig:ability_matrix_by_domain} presents these matrices separately for each of the six occupational domains, showing how the shared capability core is retuned by each domain's characteristic secondary demands (\S\ref{sec:questionnaire_results}).

\begin{figure}[h]
\centering
\includegraphics[width=1\linewidth]{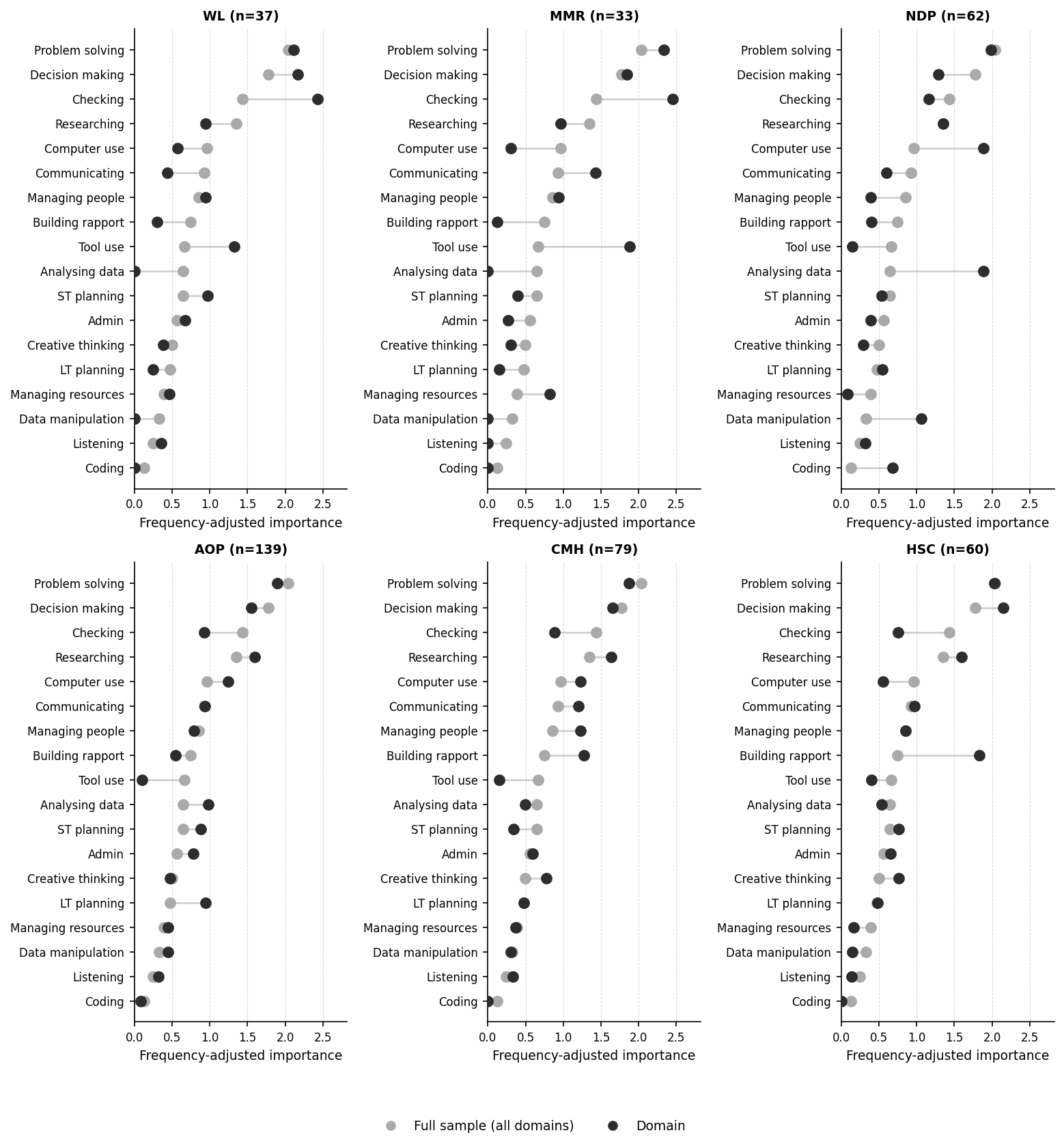}
\vspace{6pt}
\caption{Work-activity importance by occupational domain. Each panel shows one domain ($n$ = respondents in that domain). Dark dots give the domain's \emph{frequency-adjusted} importance for each activity -- the mean rank score across all respondents in the domain, scoring 5 points for a respondent's top-ranked activity down to 1 for the fifth and 0 for the unranked, so that both how often an activity is chosen and how highly it is ranked contribute. Grey dots give the full-sample reference (the equal-weight average across the six domains), identical across panels. Activities are ordered top-to-bottom by that overall reference, and the connecting line marks each domain's departure from it. Domains: Warehouse and Logistics (WL), Manufacture, Maintenance, or Repair (MMR), Numerical, Data, or Programming (NDP); Administration, Organisational, or Planning (AOP); Customer Service, Marketing, or HR (CMH); Hospitality, Sales, or Client Care (HSC).}
\label{fig:task_importance_lolipop}
\end{figure}

\begin{figure}[ht]
\centering
\includegraphics[width=1\linewidth]{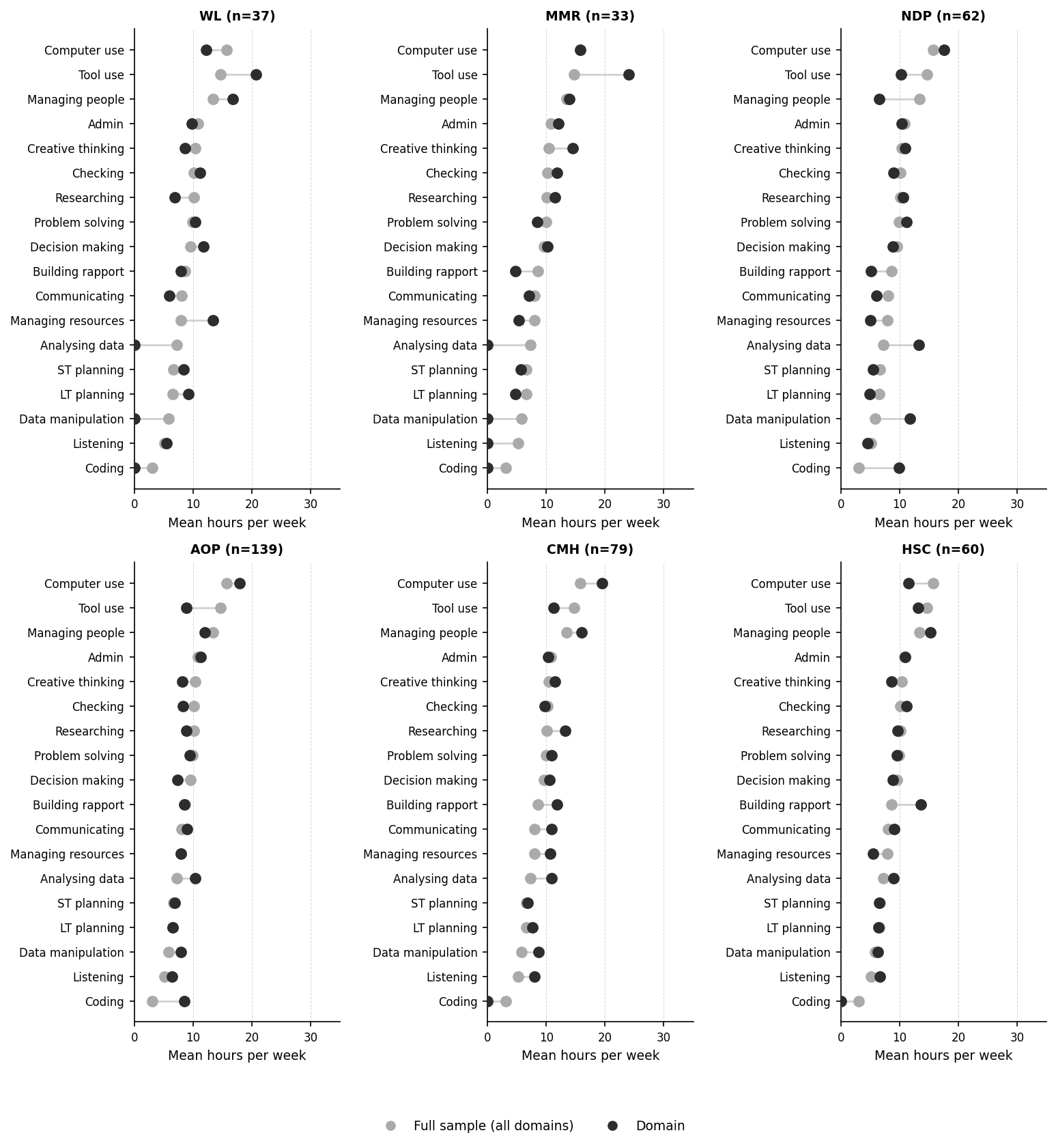}
\vspace{6pt}
\caption{Weekly hours per work activity, by occupational domain. Each panel shows one domain ($n$ = respondents in that domain). Dark dots give the mean hours per week spent on each activity \emph{among respondents who perform it}; grey dots give the full-sample reference (the equal-weight average across the six domains), identical across panels. Hours are conditional means among performers (no zero-imputation), so activities not rated in a domain are omitted; activities are ordered by the overall reference. Domains: Warehouse and Logistics (WL), Manufacture, Maintenance, or Repair (MMR), Numerical, Data, or Programming (NDP); Administration, Organisational, or Planning (AOP); Customer Service, Marketing, or HR (CMH); Hospitality, Sales, or Client Care (HSC).}
\label{fig:task_hours_lolipop}
\end{figure}

Figures~\ref{fig:task_importance_lolipop}--\ref{fig:task_hours_lolipop} break down, by domain, how respondents valued each work activity, using the two measures the questionnaire collected. Figure~\ref{fig:task_importance_lolipop} reports frequency-adjusted importance from the selection-and-ranking task, while Figure~\ref{fig:task_hours_lolipop} reports the hours per week respondents reported spending on each activity. Presenting both makes explicit the gap noted in the main text between what workers treat as important and where their time actually accumulates: high-value judgement activities tend to be weighted heavily but performed intermittently, whereas execution-oriented activities consume more of the working week than their importance ranking alone would suggest.

\clearpage
\subsection{Suitability mapping}\label{app:results_suitability}

This appendix collects the supplementary suitability analyses underlying \S\ref{sec:suitability_results}. Table~\ref{tab:suit_pxs_invariance} summarises the robustness of the suitability rankings across the compensatory sweep, reporting for each task whether the top-ranked system remains unchanged, how many distinct rankings are observed, how many system pairs preserve their ordering, and how many remain statistically distinguishable after propagating uncertainty in the elicited task-importance weights. Table~\ref{tab:suit_pxs_kendall} provides the complementary global view, reporting the mean Kendall's $\tau$ between each parameter setting and the neutral configuration across the joint $(p,s)$ sweep. Together, these analyses show that the principal deployment recommendations are robust to reasonable policy choices, with meaningful changes appearing only under deliberately extreme settings.

\begin{figure}[h]
\centering
\includegraphics[width=0.8\linewidth]{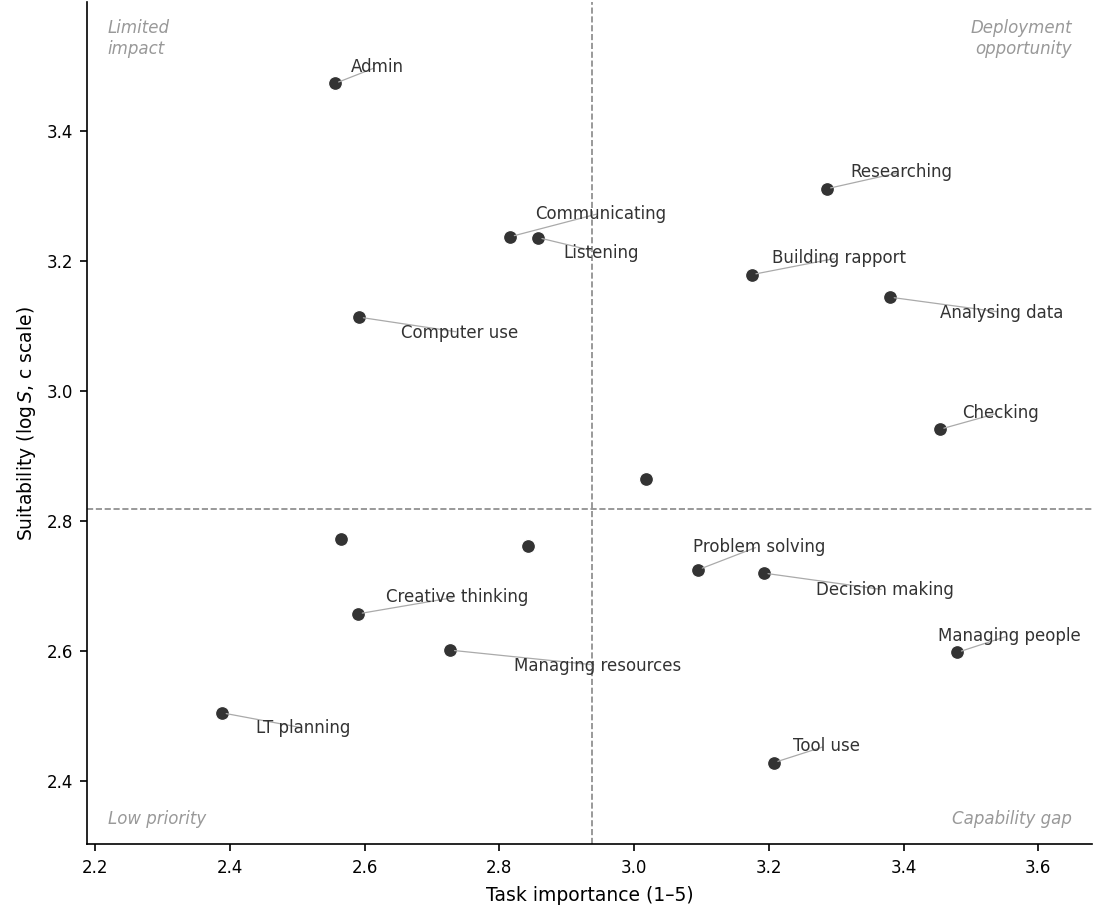}
\vspace{6pt}
\caption{Task importance (x; mean rank-derived rating, 1–5) vs. AI suitability (y; log-capability scale), pooled across six models for the full participant sample ($N = 410$). Dashed lines are the medians, defining development opportunity (top-right), capability-gap (bottom-right), limited-impact (top-left), and low-priority (bottom-left) quadrants. Labels shown only where a task's quadrant placement is confident (posterior probability $\ge 0.75$); unlabelled points straddle the suitability median.}
\label{fig:automation_full}
\end{figure}

\begin{figure}[h]
\centering
\includegraphics[width=0.8\linewidth]{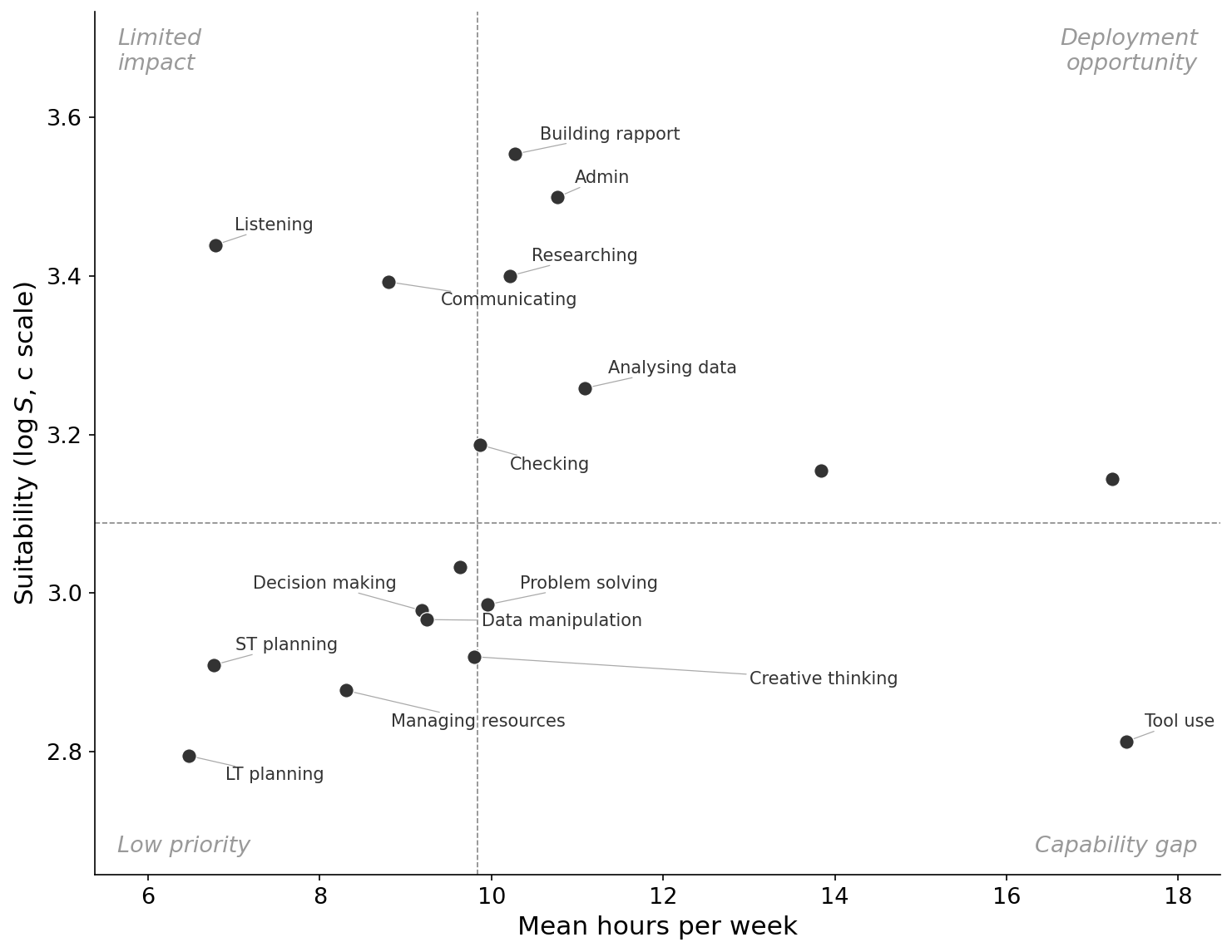}
\vspace{6pt}
\caption{Task frequency (x; mean hours per week) vs. AI suitability (y; log-capability scale), pooled across six models for the full participant sample ($N = 410$). Dashed lines are the medians, defining development opportunity (top-right), capability-gap (bottom-right), limited-impact (top-left), and low-priority (bottom-left) quadrants. Labels shown only where a task's quadrant placement is confident (posterior probability $\ge 0.75$); unlabelled points straddle the suitability median.}
\label{fig:automation_hours}
\end{figure}

\begin{figure}[h]
\centering
\includegraphics[width=0.7\linewidth]{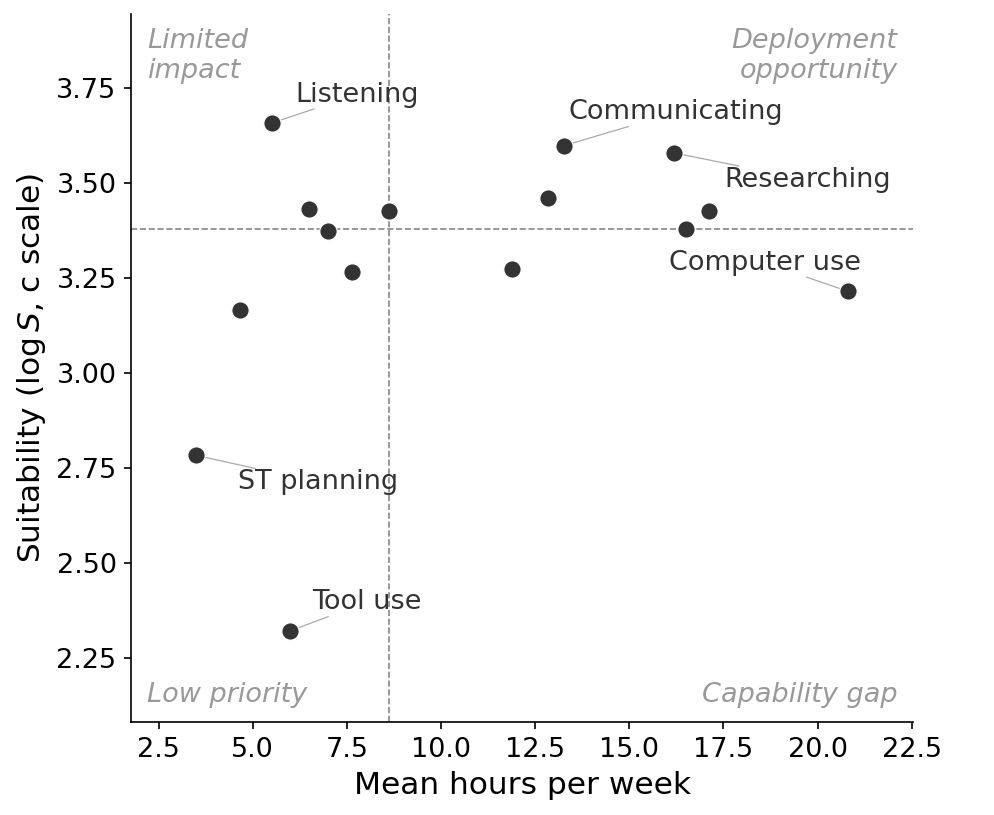}
\vspace{6pt}
\caption{Task frequency (x; mean hours per week) vs. AI suitability (y; log-capability scale) for Company X ($N=35$), pooled across six models. Dashed lines are the medians, defining development opportunity (top-right), capability-gap (bottom-right), limited-impact (top-left), and low-priority (bottom-left) quadrants. Labels shown only where a task's quadrant placement is confident (posterior probability $\ge 0.75$); unlabelled points straddle the suitability median.}
\label{fig:automation_hours_company_x}
\end{figure}

\begin{figure}[h]
\centering
\includegraphics[width=0.7\linewidth]{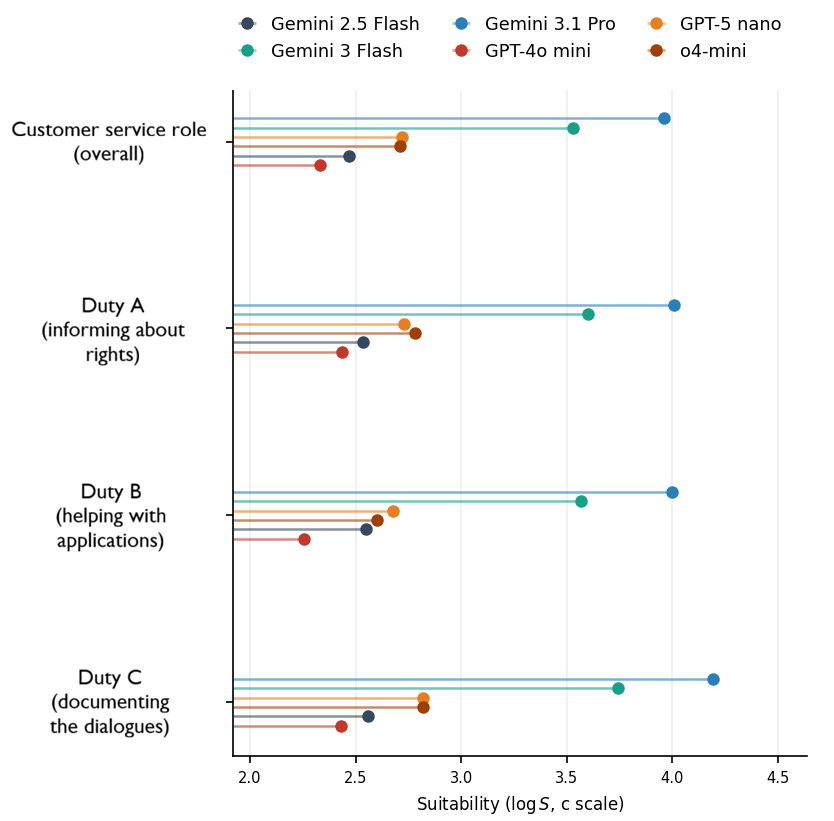}
\vspace{6pt}
\caption{Suitability scores for six AI systems computed after interviewing a senior Customer Service employee within Company X. Suitability is the importance-weighted power mean of ratio-scale capabilities ($\theta=e^c$), reported as $\log S$, and computed for the role overall and for three core duties within it: (A) informing about rights; (B) helping with applications; and (C) documenting the dialogues.}
\label{fig:interview_plot}
\end{figure}

Tables~\ref{tab:task-domain-suitability} and~\ref{tab:task-domain-priority} break the pooled suitability scores out by occupational domain. The former reports suitability ($\log S$) computed using each domain's own elicited importance weights, while the latter combines suitability with task importance to give the deployment-priority score, $\log(\mathrm{importance}\times S)$. Finally, Figures~\ref{fig:automation_full}--\ref{fig:interview_plot} present the deployment maps referenced in the main text: the full-sample importance-versus-suitability and frequency-versus-suitability plots, the corresponding frequency-versus-suitability analysis for Company~X, and the finer-grained duty-level suitability estimates derived from a senior Customer Service interview at Company~X.

\begin{table}[h]
\centering
\caption{Robustness of suitability rankings across compensatory settings.}
\label{tab:suit_pxs_invariance}
\begin{tabular}{lccccc}
\toprule
Task &
Winner
stable &
Unique
rankings &
Invariant
pairs &
\multicolumn{2}{c}{Separated pairs} \\
\cmidrule(lr){5-6}
&&&&
$\kappa\!\to\!\infty$ &
$\kappa_t\!\approx\!n_t$ \\
\midrule
Researching         & Yes & 4 & 12/15 &  9/15 & 6/15 \\
Checking            & Yes & 5 & 12/15 &  9/15 & 6/15 \\
Decision making     & Yes & 5 & 12/15 & 10/15 & 6/15 \\
Problem solving     & Yes & 4 & 13/15 & 10/15 & 8/15 \\
Creative thinking   & Yes & 3 & 13/15 &  9/15 & 6/15 \\
LT planning         & Yes & 4 & 12/15 & 10/15 & 6/15 \\
ST planning         & Yes & 4 & 12/15 & 10/15 & 6/15 \\
Tool use            & Yes & 6 & 11/15 & 10/15 & 6/15 \\
Computer use        & Yes & 4 & 12/15 &  9/15 & 7/15 \\
Coding              & Yes & 4 & 12/15 & 10/15 & 2/15 \\
Data manipulation   & Yes & 4 & 12/15 & 10/15 & 6/15 \\
Analysing data      & Yes & 5 & 12/15 & 10/15 & 6/15 \\
Communicating       & Yes & 7 & 10/15 &  7/15 & 7/15 \\
Listening           & Yes & 4 & 12/15 &  7/15 & 5/15 \\
Building rapport    & Yes & 3 & 11/15 &  5/15 & 5/15 \\
Managing people     & Yes & 5 & 11/15 &  8/15 & 5/15 \\
Managing resources  & Yes & 4 & 13/15 & 11/15 & 6/15 \\
Admin      & No  & 4 & 11/15 &  8/15 & 5/15 \\
\midrule
\textit{Mean / total} &
17/18 &
4.4 &
11.8/15 &
9.0/15 &
5.8/15 \\
\bottomrule
\end{tabular}

\vspace{0.5em}
{\raggedright\footnotesize
\textit{Note.} Rankings are evaluated across the compensatory sweep
$p\in\{-2,-1,-0.5,0,0.5,1,2\}$ with demand sharpness fixed at $s=1$.
Gemini~3.1 Pro remains the highest-ranked system for every task except Admin, which changes only at the most extreme setting ($p=2$). \emph{Winner stable} indicates whether the top-ranked system is unchanged across the sweep. \emph{Unique rankings} counts the number of distinct six-system rankings observed (1 = fully invariant). \emph{Invariant pairs} counts system pairs (out of 15) whose relative ordering never changes. \emph{Separated pairs} additionally require non-overlapping 95\% credible intervals at the operating point ($p=0$, $s=1$); the two columns compare analyses without ($\kappa\rightarrow\infty$) and with ($\kappa_t\approx n_t$) uncertainty in the elicited task-importance weights.
\par}
\end{table}

\begin{table}[h]
  \centering
  \caption{Ranking stability across suitability-policy settings.}
  \label{tab:suit_pxs_kendall}
  \begin{tabular}{lccccccc}
    \toprule
    & \multicolumn{7}{c}{$p$} \\
    \cmidrule(lr){2-8}
    $s$ & $-2$ & $-1$ & $-0.5$ & $0$ & $0.5$ & $1$ & $2$ \\
    \midrule
    1 & 0.80 & 0.81 & 0.82 & \fbox{1.00} & 0.88 & 0.85 & 0.65 \\
    2 & 0.82 & 0.79 & 0.79 & 0.87 & 0.89 & 0.81 & 0.64 \\
    4 & 0.73 & 0.75 & 0.75 & 0.76 & 0.80 & 0.79 & 0.61 \\
    8 & 0.72 & 0.72 & 0.70 & 0.70 & 0.71 & 0.65 & 0.61 \\
    \bottomrule
  \end{tabular}

  \vspace{0.5em}
{\raggedright \footnotesize
\textit{Note.} Entries are the mean Kendall's $\tau$ between each cell's per-task model ranking and the neutral operating point ($p=0$, geometric mean; $s=1$, raw importance weights; boxed cell is the neutral anchor, $\tau=1$ by construction). Here, $p$ controls the degree to which strong capabilities compensate for weaker ones, and $s$ controls how strongly tasks emphasise their highest-weighted capabilities. All other settings are fixed at the operating configuration (shared-intercept soft-min pooling at $\tau=1$, no column normalisation). Higher Kendall's $\tau$ indicates greater agreement with the neutral ranking.
\par}
\end{table}

\begin{table*}[t]
\centering
\footnotesize
\setlength{\tabcolsep}{5pt}
\caption{AI suitability per task and work domain}
\label{tab:task-domain-suitability}
\begin{tabular}{lccccccc}
\toprule
Task & WL & MMR & NDP & AOP & CMH & HSC & All \\
\midrule
Admin & \makecell{3.41\\{\scriptsize[3.17,\,3.64]}} & \makecell{3.41\\{\scriptsize[3.19,\,3.64]}} & \makecell{3.07\\{\scriptsize[2.84,\,3.30]}} & \makecell{3.34\\{\scriptsize[3.11,\,3.57]}} & \makecell{3.18\\{\scriptsize[2.95,\,3.40]}} & \makecell{3.64\\{\scriptsize[3.41,\,3.87]}} & \makecell{3.47\\{\scriptsize[3.24,\,3.71]}} \\
Researching & \makecell{3.47\\{\scriptsize[3.23,\,3.71]}} & \makecell{3.08\\{\scriptsize[2.85,\,3.30]}} & \makecell{3.37\\{\scriptsize[3.13,\,3.61]}} & \makecell{3.22\\{\scriptsize[2.99,\,3.44]}} & \makecell{3.15\\{\scriptsize[2.91,\,3.38]}} & \makecell{3.16\\{\scriptsize[2.93,\,3.39]}} & \makecell{3.31\\{\scriptsize[3.07,\,3.55]}} \\
Communicating & \makecell{2.74\\{\scriptsize[2.51,\,2.96]}} & \makecell{2.91\\{\scriptsize[2.68,\,3.13]}} & \makecell{3.25\\{\scriptsize[3.03,\,3.47]}} & \makecell{3.25\\{\scriptsize[3.02,\,3.48]}} & \makecell{3.22\\{\scriptsize[2.99,\,3.46]}} & \makecell{3.32\\{\scriptsize[3.08,\,3.54]}} & \makecell{3.24\\{\scriptsize[3.01,\,3.46]}} \\
Listening & \makecell{3.27\\{\scriptsize[3.04,\,3.51]}} & \makecell{3.86\\{\scriptsize[3.59,\,4.11]}} & \makecell{3.29\\{\scriptsize[3.06,\,3.52]}} & \makecell{3.16\\{\scriptsize[2.92,\,3.38]}} & \makecell{3.28\\{\scriptsize[3.05,\,3.50]}} & \makecell{2.97\\{\scriptsize[2.73,\,3.20]}} & \makecell{3.23\\{\scriptsize[3.00,\,3.46]}} \\
Building rapport & \makecell{3.40\\{\scriptsize[3.17,\,3.63]}} & \makecell{3.69\\{\scriptsize[3.46,\,3.92]}} & \makecell{3.29\\{\scriptsize[3.06,\,3.53]}} & \makecell{3.33\\{\scriptsize[3.08,\,3.56]}} & \makecell{2.98\\{\scriptsize[2.74,\,3.22]}} & \makecell{3.40\\{\scriptsize[3.17,\,3.64]}} & \makecell{3.18\\{\scriptsize[2.93,\,3.42]}} \\
Analysing data & \makecell{3.72\\{\scriptsize[3.45,\,4.00]}} & \makecell{2.73\\{\scriptsize[2.48,\,2.96]}} & \makecell{3.05\\{\scriptsize[2.81,\,3.28]}} & \makecell{3.13\\{\scriptsize[2.88,\,3.37]}} & \makecell{2.94\\{\scriptsize[2.70,\,3.18]}} & \makecell{3.15\\{\scriptsize[2.92,\,3.38]}} & \makecell{3.15\\{\scriptsize[2.90,\,3.39]}} \\
Computer use & \makecell{2.98\\{\scriptsize[2.74,\,3.21]}} & \makecell{2.95\\{\scriptsize[2.73,\,3.19]}} & \makecell{2.88\\{\scriptsize[2.66,\,3.11]}} & \makecell{3.00\\{\scriptsize[2.77,\,3.25]}} & \makecell{3.09\\{\scriptsize[2.84,\,3.32]}} & \makecell{3.32\\{\scriptsize[3.07,\,3.56]}} & \makecell{3.11\\{\scriptsize[2.88,\,3.35]}} \\
Checking & \makecell{2.76\\{\scriptsize[2.52,\,2.99]}} & \makecell{2.74\\{\scriptsize[2.50,\,2.97]}} & \makecell{3.19\\{\scriptsize[2.95,\,3.42]}} & \makecell{3.03\\{\scriptsize[2.81,\,3.26]}} & \makecell{2.82\\{\scriptsize[2.59,\,3.04]}} & \makecell{3.05\\{\scriptsize[2.81,\,3.29]}} & \makecell{2.94\\{\scriptsize[2.70,\,3.17]}} \\
Data manipulation & \makecell{2.58\\{\scriptsize[2.35,\,2.83]}} & \makecell{3.46\\{\scriptsize[3.22,\,3.71]}} & \makecell{2.78\\{\scriptsize[2.53,\,3.02]}} & \makecell{2.95\\{\scriptsize[2.72,\,3.19]}} & \makecell{2.82\\{\scriptsize[2.59,\,3.05]}} & \makecell{2.95\\{\scriptsize[2.71,\,3.19]}} & \makecell{2.86\\{\scriptsize[2.63,\,3.11]}} \\
ST planning & \makecell{3.02\\{\scriptsize[2.79,\,3.25]}} & \makecell{2.17\\{\scriptsize[1.94,\,2.40]}} & \makecell{2.87\\{\scriptsize[2.64,\,3.10]}} & \makecell{2.73\\{\scriptsize[2.50,\,2.96]}} & \makecell{2.85\\{\scriptsize[2.61,\,3.09]}} & \makecell{2.85\\{\scriptsize[2.62,\,3.08]}} & \makecell{2.77\\{\scriptsize[2.54,\,3.00]}} \\
Coding & \textemdash & \textemdash & \makecell{3.01\\{\scriptsize[2.78,\,3.24]}} & \makecell{2.57\\{\scriptsize[2.35,\,2.80]}} & \textemdash & \textemdash & \makecell{2.76\\{\scriptsize[2.52,\,2.99]}} \\
Problem solving & \makecell{2.95\\{\scriptsize[2.72,\,3.18]}} & \makecell{2.44\\{\scriptsize[2.21,\,2.67]}} & \makecell{2.70\\{\scriptsize[2.47,\,2.93]}} & \makecell{2.84\\{\scriptsize[2.61,\,3.08]}} & \makecell{2.80\\{\scriptsize[2.57,\,3.03]}} & \makecell{2.92\\{\scriptsize[2.68,\,3.14]}} & \makecell{2.72\\{\scriptsize[2.49,\,2.95]}} \\
Decision making & \makecell{3.10\\{\scriptsize[2.87,\,3.32]}} & \makecell{2.54\\{\scriptsize[2.30,\,2.77]}} & \makecell{2.75\\{\scriptsize[2.52,\,2.98]}} & \makecell{2.91\\{\scriptsize[2.67,\,3.14]}} & \makecell{2.56\\{\scriptsize[2.34,\,2.78]}} & \makecell{2.73\\{\scriptsize[2.49,\,2.95]}} & \makecell{2.72\\{\scriptsize[2.48,\,2.95]}} \\
Creative thinking & \makecell{3.00\\{\scriptsize[2.77,\,3.23]}} & \makecell{3.09\\{\scriptsize[2.85,\,3.33]}} & \makecell{2.81\\{\scriptsize[2.57,\,3.04]}} & \makecell{2.55\\{\scriptsize[2.32,\,2.78]}} & \makecell{2.84\\{\scriptsize[2.60,\,3.07]}} & \makecell{2.93\\{\scriptsize[2.69,\,3.16]}} & \makecell{2.66\\{\scriptsize[2.42,\,2.89]}} \\
Managing resources & \makecell{3.09\\{\scriptsize[2.86,\,3.32]}} & \makecell{2.64\\{\scriptsize[2.42,\,2.87]}} & \makecell{2.52\\{\scriptsize[2.29,\,2.75]}} & \makecell{2.88\\{\scriptsize[2.65,\,3.12]}} & \makecell{2.40\\{\scriptsize[2.18,\,2.62]}} & \makecell{2.63\\{\scriptsize[2.40,\,2.86]}} & \makecell{2.60\\{\scriptsize[2.37,\,2.83]}} \\
Managing people & \makecell{3.10\\{\scriptsize[2.88,\,3.33]}} & \makecell{2.53\\{\scriptsize[2.30,\,2.75]}} & \makecell{2.64\\{\scriptsize[2.41,\,2.88]}} & \makecell{2.93\\{\scriptsize[2.70,\,3.16]}} & \makecell{2.62\\{\scriptsize[2.39,\,2.84]}} & \makecell{3.17\\{\scriptsize[2.93,\,3.40]}} & \makecell{2.60\\{\scriptsize[2.37,\,2.82]}} \\
LT planning & \makecell{2.80\\{\scriptsize[2.58,\,3.03]}} & \makecell{2.39\\{\scriptsize[2.15,\,2.62]}} & \makecell{2.59\\{\scriptsize[2.36,\,2.81]}} & \makecell{2.59\\{\scriptsize[2.36,\,2.81]}} & \makecell{2.86\\{\scriptsize[2.63,\,3.09]}} & \makecell{2.62\\{\scriptsize[2.40,\,2.85]}} & \makecell{2.50\\{\scriptsize[2.28,\,2.73]}} \\
Tool use & \makecell{2.25\\{\scriptsize[2.02,\,2.48]}} & \makecell{2.52\\{\scriptsize[2.29,\,2.75]}} & \makecell{3.11\\{\scriptsize[2.87,\,3.34]}} & \makecell{2.54\\{\scriptsize[2.31,\,2.76]}} & \makecell{3.76\\{\scriptsize[3.52,\,3.99]}} & \makecell{2.58\\{\scriptsize[2.34,\,2.81]}} & \makecell{2.43\\{\scriptsize[2.19,\,2.67]}} \\
\bottomrule
\end{tabular}

\vspace{0.5em}
{\raggedright \footnotesize
\textit{Note.} Each cell reports pooled AI suitability across the six evaluated models on the log-capability scale ($\log S$), shown as the posterior mean with a 95\% credible interval and computed using the task-ability importance weights for that domain.  Rows are ordered by overall $\log S$. Column keys: WL = Warehouse or logistics; MMR = Manufacture, maintenance, or repair; NDP = Numerical, data, or programming; AOP = Administration, organisational, or planning; CMH = Customer service, marketing or HR; HSC = Hospitality, sales, or client care. A dash (\textemdash{}) indicates that a task was absent from that domain's data.
\par}
\end{table*}

\begin{table*}[t]
\centering
\footnotesize
\setlength{\tabcolsep}{5pt}
\caption{AI deployment priority score by task and work domain}
\label{tab:task-domain-priority}
\begin{tabular}{lccccccc}
\toprule
Task & WL & MMR & NDP & AOP & CMH & HSC & All \\
\midrule
Researching & \makecell{4.63\\{\scriptsize[4.38,\,4.87]}} & \makecell{4.14\\{\scriptsize[3.92,\,4.37]}} & \makecell{4.47\\{\scriptsize[4.23,\,4.71]}} & \makecell{4.45\\{\scriptsize[4.22,\,4.67]}} & \makecell{4.34\\{\scriptsize[4.11,\,4.58]}} & \makecell{4.39\\{\scriptsize[4.16,\,4.62]}} & \makecell{4.50\\{\scriptsize[4.26,\,4.74]}} \\
Admin & \makecell{4.14\\{\scriptsize[3.91,\,4.37]}} & \makecell{4.51\\{\scriptsize[4.29,\,4.74]}} & \makecell{3.94\\{\scriptsize[3.72,\,4.17]}} & \makecell{4.39\\{\scriptsize[4.16,\,4.62]}} & \makecell{3.94\\{\scriptsize[3.71,\,4.16]}} & \makecell{4.67\\{\scriptsize[4.44,\,4.89]}} & \makecell{4.41\\{\scriptsize[4.18,\,4.65]}} \\
Analysing data & \makecell{4.97\\{\scriptsize[4.70,\,5.25]}} & \makecell{3.98\\{\scriptsize[3.73,\,4.22]}} & \makecell{4.37\\{\scriptsize[4.14,\,4.61]}} & \makecell{4.33\\{\scriptsize[4.08,\,4.57]}} & \makecell{3.90\\{\scriptsize[3.66,\,4.14]}} & \makecell{4.42\\{\scriptsize[4.19,\,4.65]}} & \makecell{4.36\\{\scriptsize[4.12,\,4.60]}} \\
Building rapport & \makecell{3.72\\{\scriptsize[3.49,\,3.95]}} & \makecell{3.69\\{\scriptsize[3.46,\,3.92]}} & \makecell{4.32\\{\scriptsize[4.08,\,4.55]}} & \makecell{4.36\\{\scriptsize[4.11,\,4.60]}} & \makecell{4.34\\{\scriptsize[4.09,\,4.57]}} & \makecell{4.74\\{\scriptsize[4.50,\,4.97]}} & \makecell{4.33\\{\scriptsize[4.08,\,4.57]}} \\
Listening & \makecell{4.45\\{\scriptsize[4.22,\,4.69]}} & \makecell{5.36\\{\scriptsize[5.09,\,5.62]}} & \makecell{4.08\\{\scriptsize[3.85,\,4.31]}} & \makecell{4.23\\{\scriptsize[4.00,\,4.46]}} & \makecell{4.34\\{\scriptsize[4.11,\,4.56]}} & \makecell{3.95\\{\scriptsize[3.71,\,4.19]}} & \makecell{4.28\\{\scriptsize[4.05,\,4.51]}} \\
Communicating & \makecell{3.56\\{\scriptsize[3.34,\,3.78]}} & \makecell{3.98\\{\scriptsize[3.76,\,4.21]}} & \makecell{4.30\\{\scriptsize[4.08,\,4.52]}} & \makecell{4.31\\{\scriptsize[4.08,\,4.54]}} & \makecell{4.28\\{\scriptsize[4.05,\,4.51]}} & \makecell{4.29\\{\scriptsize[4.05,\,4.51]}} & \makecell{4.27\\{\scriptsize[4.04,\,4.50]}} \\
Checking & \makecell{4.12\\{\scriptsize[3.88,\,4.35]}} & \makecell{4.14\\{\scriptsize[3.90,\,4.37]}} & \makecell{4.29\\{\scriptsize[4.05,\,4.52]}} & \makecell{4.23\\{\scriptsize[4.00,\,4.46]}} & \makecell{4.02\\{\scriptsize[3.79,\,4.24]}} & \makecell{4.22\\{\scriptsize[3.98,\,4.46]}} & \makecell{4.18\\{\scriptsize[3.94,\,4.41]}} \\
Computer use & \makecell{3.72\\{\scriptsize[3.48,\,3.95]}} & \makecell{3.31\\{\scriptsize[3.08,\,3.55]}} & \makecell{4.01\\{\scriptsize[3.78,\,4.24]}} & \makecell{3.91\\{\scriptsize[3.67,\,4.15]}} & \makecell{4.11\\{\scriptsize[3.86,\,4.34]}} & \makecell{4.17\\{\scriptsize[3.93,\,4.41]}} & \makecell{4.07\\{\scriptsize[3.83,\,4.31]}} \\
Data manipulation & \makecell{4.19\\{\scriptsize[3.96,\,4.44]}} & \makecell{4.85\\{\scriptsize[4.60,\,5.10]}} & \makecell{3.97\\{\scriptsize[3.73,\,4.21]}} & \makecell{3.95\\{\scriptsize[3.71,\,4.18]}} & \makecell{3.80\\{\scriptsize[3.57,\,4.03]}} & \makecell{4.05\\{\scriptsize[3.81,\,4.28]}} & \makecell{3.97\\{\scriptsize[3.73,\,4.21]}} \\
Decision making & \makecell{4.22\\{\scriptsize[4.00,\,4.45]}} & \makecell{3.65\\{\scriptsize[3.42,\,3.88]}} & \makecell{3.95\\{\scriptsize[3.73,\,4.18]}} & \makecell{4.07\\{\scriptsize[3.84,\,4.31]}} & \makecell{3.70\\{\scriptsize[3.47,\,3.92]}} & \makecell{3.92\\{\scriptsize[3.69,\,4.15]}} & \makecell{3.88\\{\scriptsize[3.64,\,4.11]}} \\
Problem solving & \makecell{4.05\\{\scriptsize[3.81,\,4.28]}} & \makecell{3.52\\{\scriptsize[3.30,\,3.75]}} & \makecell{3.87\\{\scriptsize[3.64,\,4.11]}} & \makecell{3.97\\{\scriptsize[3.74,\,4.21]}} & \makecell{3.93\\{\scriptsize[3.70,\,4.16]}} & \makecell{4.06\\{\scriptsize[3.82,\,4.28]}} & \makecell{3.85\\{\scriptsize[3.62,\,4.08]}} \\
Managing people & \makecell{4.46\\{\scriptsize[4.24,\,4.69]}} & \makecell{3.66\\{\scriptsize[3.43,\,3.88]}} & \makecell{3.87\\{\scriptsize[3.64,\,4.11]}} & \makecell{4.20\\{\scriptsize[3.96,\,4.42]}} & \makecell{3.97\\{\scriptsize[3.75,\,4.20]}} & \makecell{4.21\\{\scriptsize[3.97,\,4.44]}} & \makecell{3.84\\{\scriptsize[3.62,\,4.07]}} \\
Coding & \textemdash & \textemdash & \makecell{4.04\\{\scriptsize[3.81,\,4.27]}} & \makecell{3.67\\{\scriptsize[3.45,\,3.90]}} & \textemdash & \textemdash & \makecell{3.80\\{\scriptsize[3.57,\,4.03]}} \\
ST planning & \makecell{4.04\\{\scriptsize[3.81,\,4.27]}} & \makecell{3.13\\{\scriptsize[2.89,\,3.36]}} & \makecell{3.88\\{\scriptsize[3.66,\,4.11]}} & \makecell{3.77\\{\scriptsize[3.54,\,4.00]}} & \makecell{3.44\\{\scriptsize[3.20,\,3.67]}} & \makecell{3.68\\{\scriptsize[3.45,\,3.91]}} & \makecell{3.71\\{\scriptsize[3.48,\,3.94]}} \\
Creative thinking & \makecell{3.69\\{\scriptsize[3.46,\,3.93]}} & \makecell{3.60\\{\scriptsize[3.36,\,3.84]}} & \makecell{3.62\\{\scriptsize[3.38,\,3.86]}} & \makecell{3.56\\{\scriptsize[3.33,\,3.79]}} & \makecell{3.95\\{\scriptsize[3.72,\,4.18]}} & \makecell{3.87\\{\scriptsize[3.63,\,4.10]}} & \makecell{3.61\\{\scriptsize[3.37,\,3.84]}} \\
Managing resources & \makecell{3.97\\{\scriptsize[3.75,\,4.20]}} & \makecell{3.74\\{\scriptsize[3.52,\,3.97]}} & \makecell{2.52\\{\scriptsize[2.29,\,2.75]}} & \makecell{4.06\\{\scriptsize[3.83,\,4.30]}} & \makecell{3.37\\{\scriptsize[3.15,\,3.59]}} & \makecell{3.55\\{\scriptsize[3.32,\,3.78]}} & \makecell{3.61\\{\scriptsize[3.38,\,3.84]}} \\
Tool use & \makecell{3.66\\{\scriptsize[3.43,\,3.89]}} & \makecell{3.82\\{\scriptsize[3.59,\,4.05]}} & \makecell{3.92\\{\scriptsize[3.68,\,4.16]}} & \makecell{3.23\\{\scriptsize[3.00,\,3.45]}} & \makecell{4.63\\{\scriptsize[4.40,\,4.87]}} & \makecell{3.68\\{\scriptsize[3.44,\,3.91]}} & \makecell{3.59\\{\scriptsize[3.36,\,3.83]}} \\
LT planning & \makecell{3.61\\{\scriptsize[3.39,\,3.84]}} & \makecell{2.61\\{\scriptsize[2.38,\,2.84]}} & \makecell{3.41\\{\scriptsize[3.18,\,3.63]}} & \makecell{3.57\\{\scriptsize[3.34,\,3.80]}} & \makecell{3.72\\{\scriptsize[3.50,\,3.96]}} & \makecell{3.28\\{\scriptsize[3.06,\,3.51]}} & \makecell{3.37\\{\scriptsize[3.15,\,3.60]}} \\
\bottomrule
\end{tabular}

\vspace{0.5em}
{\raggedright \footnotesize
\textit{Note.} Each cell reports the log deployment-priority score, $\log P_t$, where $P_t = I_tS_t$ is the product of the task-importance score ($I_t$) and mean AI suitability ($S_t$). Suitability is computed as the equally weighted geometric mean across the six profiled AI systems (equivalently, the arithmetic mean on the log-capability scale). The 95\% credible interval is obtained by shifting the suitability posterior by the fixed $\log I_t$ offset. Rows are ordered by overall deployment priority. Column keys: WL = Warehouse or logistics; MMR = Manufacture, maintenance, or repair; NDP = Numerical, data, or programming; AOP = Administration, organisational, or planning; CMH = Customer service, marketing, or HR; HSC = Hospitality, sales, or client care. A dash (\textemdash{}) indicates that a task was absent from that domain's data.
\par}
\end{table*}

\end{document}